\PassOptionsToPackage{libertine,vvarbb}{newtxmath}
\documentclass[a4paper]{article}
\usepackage[final]{colm2025_conference}

\setcitestyle{numbers,square,comma,sort,compress}

\usepackage{xspace}
\usepackage{url}
\usepackage{comment}
\usepackage{tabularray}
\UseTblrLibrary{booktabs}
\usepackage{amsmath}
\usepackage{amsfonts}
\usepackage{bm}
\usepackage{bbm}
\usepackage{algorithm}
\usepackage{textcomp}
\usepackage{algpseudocode}
\usepackage{booktabs}
\usepackage{longtable}
\usepackage{graphicx}
\usepackage{tikz}
\usetikzlibrary{arrows.meta,calc,fit,positioning}
\usepackage{subcaption}
\usepackage{xcolor}
\usepackage{colortbl}
\usepackage{multirow}
\usepackage{adjustbox}
\usepackage{geometry}
\usepackage{pdflscape}
\usepackage{lineno}
\usepackage{placeins}
\usepackage{setspace}
\usepackage{threeparttable}
\usepackage{hyperref}
\usepackage{cleveref}
\usepackage{etoolbox}
\usepackage[scheme=plain]{ctex}

\graphicspath{{./fig/}{./figures/generated/}{./figures/manual/}}

\definecolor{openweight}{RGB}{255, 245, 230}  %
\definecolor{fullyopen}{RGB}{245, 235, 255}   %
\definecolor{ours}{RGB}{235, 255, 235}        %

\definecolor{openweight}{RGB}{255, 245, 230} %
\definecolor{fullyopen}{RGB}{245, 235, 255}  %
\definecolor{ours}{RGB}{235, 255, 235}       %

\usepackage{amsmath,amsfonts,bm}

\def\eqref#1{\Cref{#1}}

\def\1{\bm{1}}

\DeclareMathAlphabet{\mathsfit}{\encodingdefault}{\sfdefault}{m}{sl}
\SetMathAlphabet{\mathsfit}{bold}{\encodingdefault}{\sfdefault}{bx}{n}

\newcommand{\sys}{\textsc{Puro-2B}\xspace}
\newcommand{\fullsys}{\textsc{Puro-2B}\xspace}
\newcommand{\dataframework}{\textsc{Kai\-yuan-Spark}\xspace}

\newcommand{\papertitle}{\fullsys: \emph{Poor} Lab's Qwen2-1.5B Trained on RTX~5090 within \$5090}

\let\tsup\textsuperscript

\hypersetup{
  pdftitle={\papertitle},
  pdfauthor={Kairong Luo, Jiarui Cui, Yaorui Yin, Shengqi Chen, Yanmohan Wang, Yiming Yang, Chengxia Li, Linxiang Gao, Mingzhe Zhang, Kaifeng Lyu, Wenguang Chen},
}

\fancypagestyle{firstpage}{%
  \fancyhf{}
  \fancyfoot[C]{\thepage}
  \renewcommand{\headrulewidth}{0pt}
}

\crefname{section}{\S}{\S\S}
\Crefname{section}{Section}{Sections}

\NewTblrEnviron{narrowtblr}
\SetTblrInner[narrowtblr]{
  hline{1,Z} = {1pt},
  hline{2}   = {0.5pt},
  rows = {rowsep=0.8pt},
  rowhead = 1,
  rows = {valign=m},
  row{1} = {font=\bfseries, bg=gray!10},
}

\NewTblrEnviron{narrowtalltblr}
\SetTblrOuter[narrowtalltblr]{tall}
\SetTblrInner[narrowtalltblr]{
  hline{1,Z} = {1pt},
  hline{2}   = {0.5pt},
  rows = {rowsep=0.8pt},
  rowhead = 1,
  rows = {valign=m},
  row{1} = {font=\bfseries, bg=gray!10},
}

\providecommand{\MuonHComputeMultiplier}{\ensuremath{1.19\times}\xspace}
\providecommand{\MuonHComputeSaving}{16.1\%\xspace}
\providecommand{\MuonHComputeMultiplierLOO}{\ensuremath{1.17\text{--}1.28\times}\xspace}
\providecommand{\TargetTrainingCompute}{\ensuremath{1.68\times10^{22}}\xspace}
\providecommand{\TargetMuonHCompute}{\ensuremath{1.41\times10^{22}}\xspace}
\providecommand{\FPEightPrecisionRetention}{98.0\%\xspace}
\providecommand{\FPEightPrecisionRetentionLOO}{97.6--98.1\%\xspace}
\providecommand{\FPEightPrecisionPenalty}{2.0\%\xspace}
\providecommand{\FPEightThroughputSpeedup}{\ensuremath{1.36\times}\xspace}
\providecommand{\FPEightNetSpeedup}{\ensuremath{1.34\times}\xspace}
\providecommand{\FPEightGPUHourSaving}{25.2\%\xspace}
\providecommand{\TargetBFSixteenGPUHours}{30,286\xspace}
\providecommand{\TargetFPEightGPUHours}{22,654\xspace}

\providecommand{\HardwareCostEfficiencyMultiplier}{\ensuremath{2.77\times}\xspace}

\providecommand{\PhaseOneTokens}{438.8B\xspace}

\providecommand{\PhaseOneEnglishShare}{73.2\%\xspace}

\providecommand{\PhaseOneWandBPhaseTokens}{438.8B\xspace}

\providecommand{\PhaseOneWandBFinalValidationLoss}{2.730\xspace}

\providecommand{\PhaseOneWandBMedianThroughput}{238\xspace}

\providecommand{\PhaseOneWandBWorldSize}{24\xspace}
\providecommand{\PhaseOneWandBGlobalBatchSize}{1536\xspace}

\providecommand{\PhaseTwoWandBPhaseTokens}{960.0B\xspace}
\providecommand{\PhaseTwoWandBCumulativeTokens}{1.4T\xspace}

\providecommand{\PhaseTwoWandBFinalValidationLoss}{2.488\xspace}

\providecommand{\PhaseTwoWandBMedianThroughput}{192\xspace}

\providecommand{\PhaseTwoWandBWorldSize}{96\xspace}

\title{\papertitle}

\usepackage[tt=false]{libertine} %
\makeatletter
\let\@maketitleold\@maketitle
\def\@maketitle{%
\fancyheadoffset{0.5cm}%
\@maketitleold%
\lhead{\papertitle}%
\rhead{\leftmark}%
}
\pretocmd{\@maketitle}{\def\@makefnmark{\hbox{$^{\@thefnmark}$}}}{}{}
\makeatother

\author{
Kairong Luo\tsup{1}\thanks{Equal core contribution.},
Jiarui Cui\tsup{1}\footnotemark[1],
Yaorui Yin\tsup{1}\footnotemark[1],
Shengqi Chen\tsup{1}\footnotemark[1],
Yiming Yang\tsup{1},
Linxiang Gao\tsup{1},\\
\bf\ 
Yanmohan Wang\tsup{1},
Chengxia Li\tsup{1},
Mingzhe Zhang\tsup{1},
Kaifeng Lyu\tsup{1}\thanks{Corresponding authors.},
Wenguang Chen\tsup{1,2}\setcounter{footnote}{2}\footnotemark[2]\\
\tsup{1}Tsinghua University\quad \tsup{2\textbf{}}Pengcheng Laboratory\\
\texttt{luokr24@mails.tsinghua.edu.cn\quad \{klyu,cwg\}@tsinghua.edu.cn}
}

\begin{document}

\maketitle
\thispagestyle{firstpage}

\begin{abstract}

Language model pretraining has become almost synonymous with prohibitive cost, placing it out of reach for much of the academic and open-source communities. Although strong open-source efforts already exist, including open-weight models and open-source training recipes, a \emph{cost-efficient, hardware-accessible, and open-source} pretraining recipe has long been missing. Even at a small scale, training Llama-3.2-3B costs over \$1.5M, and reproducing SmolLM3-3B needs over \$700K. In this report, we present an open pretraining recipe designed to lower this barrier.
Using this recipe, we train a collection of \fullsys (普罗-2B) models from scratch on up to 1.4 trillion tokens with FP8 precision on consumer-grade RTX~5090 GPUs\footnote{We sincerely thank Yanfu Investments for generously providing GPU computing resources to support this research.}. The models in the collection differ in token budgets and selected recipe variants. Our best model is trained at a compute cost of less than \$6.9K and approaches Qwen2.5-1.5B performance under our evaluation protocol. This cost efficiency is enabled by a combination of approaches, including hardware selection, low-precision training, hyperball optimization, curriculum model averaging, and the data recipe. Beyond the recipe itself, we provide two additional results. First, across the \fullsys collection, we derive a \emph{Puro Cost Scaling Law} that relates training cost to average model performance; the fitted law suggests that about \$4.4K, \emph{less than \$5,090}, is sufficient to reach the performance of Qwen2-1.5B. Second, as an end-to-end case study, we examine how pretraining data curricula shape downstream performance after post-training.
Such controlled studies are enabled by having access to the full pretraining pipeline rather than model weights alone. 
We release the full training recipe for \fullsys, including data, code, and model weights under Apache 2.0 at     \url{https://huggingface.co/collections/thu-pacman/puro-2b}. 
\end{abstract}

\begin{figure*}[h!]
    \vspace{-0.20in}
    \centering
    \includegraphics[width=0.92\textwidth]{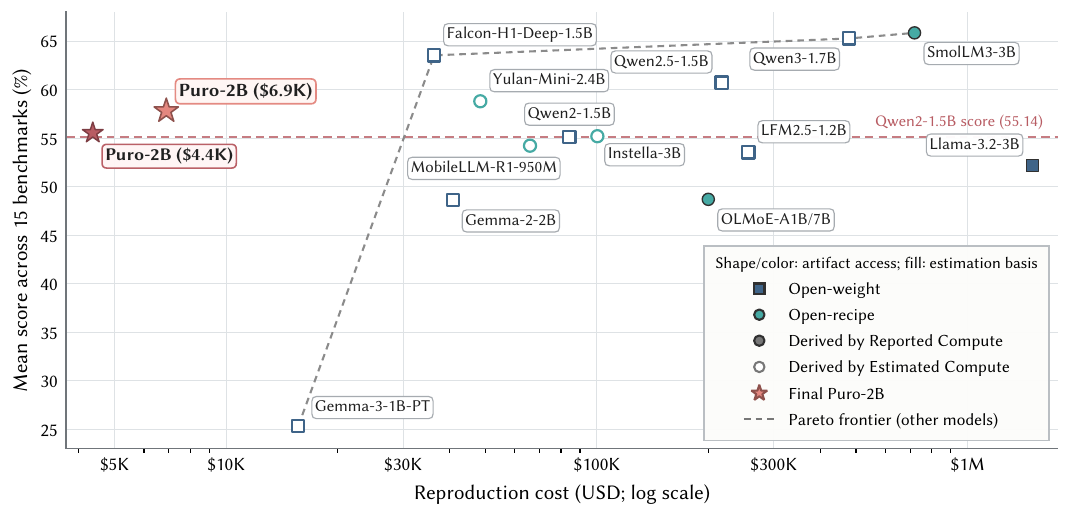}
    \caption{Model performance versus reproduction cost under
    the accounting protocol in~\Cref{sec:cost-account,sec:recipe:cost:reproduction}. Performance is measured by the
    average scores over the 15 mathematics, code, reasoning, and knowledge
    benchmarks in~\Cref{tab:math_code_capabilities,tab:reasoning_knowledge}. The cost summary of two \sys runs can be found in \Cref{tab:pipeline-cost-summary}.
    }
    \label{fig:pareto-optimal}
\end{figure*}

\tableofcontents
\clearpage

\section{Introduction}

Scaling model parameters and training data has substantially improved the capabilities of large language models (LLMs)~\citep{hoffmann2022chinchilla,kaplan2025scaling}. However, the cost of pretraining limits the ability of academic and open-source researchers to study training behavior, reproduce complete pipelines, and test alternatives at meaningful scales. Lowering this barrier requires not only open model weights, but also reproducible data, software, infrastructure, training details, and transparent cost accounting.

Existing model releases span three levels of openness. \emph{Closed models}, including proprietary families such as GPT, Claude, and Gemini~\citep{openai2023gpt4,gemini2025gemini25}, expose capabilities through hosted services and technical reports but do not release model weights or the artifacts required to inspect pretraining. \emph{Open-weight models}, including the Qwen, Gemma, and Llama families~\citep{qwen2,qwen2025qwen25technicalreport,qwen3,morgane2024gemma2,gemma3,llama3.2}, release checkpoints that enable local evaluation, deployment, and post-training, but generally withhold the exact pretraining data, sample order, and complete training state. \emph{Open-recipe models} go further by releasing reconstructible data mixtures, training code, configurations, and checkpoints. Projects such as OLMo and OLMoE, SmolLM, YuLan, Marin, and Instella therefore make the pretraining process itself available for scientific study~\citep{olmo20252olmo2furious,muennighoff2025olmoe,smol2,smollm3,yiwen-etal-2025-yulan,liu2025instella,hall2025marin}.

Greater openness, however, does not by itself make pretraining affordable to reproduce. Even at a small scale, the compute cost of pretraining can be prohibitive.
Under the rental-equivalent accounting in~\Cref{fig:pareto-optimal} and~\Cref{app:figure-one-calculation}, training Llama3.2-3B costs over \$1.5M in our estimation. For open-recipe models, reproducing OLMoE-1B-7B would cost \$200K, and for SmolLM3-3B, the estimated reproduction cost rises to \$719K. These cost budgets are derived from reported or estimated compute of these models. Therefore, even transparent open-recipe releases can remain \emph{open to the world but out of reach for many small and resource-constrained research labs --- the poor labs}, leaving a practical gap between reproducibility and accessibility.

To bridge this gap, we present a systematic reproducibility recipe covering system, algorithm, and data design for training a collection of \textbf{\fullsys (普罗-2B)} checkpoints\footnote{Unless otherwise noted, we will use \fullsys to denote the best one in this collection.}. The recipe targets low-cost, accessible infrastructure for dense billion-parameter/trillion-token pretraining and produces a 2B model with performance matched against Qwen2-1.5B and Qwen2.5-1.5B under our protocol.
These checkpoints share the same 2B-parameter architecture but differ in training budget and recipe variants. Training uses predominantly open-source datasets and runs on a cluster of consumer-grade RTX~5090 GPUs. %
\Cref{fig:pareto-optimal} compares counterpart models under a common cost-accounting protocol. \sys attains competitive performance at the lowest reproduction cost among the plotted models under the stated accounting protocol, placing it beyond the baseline cost--performance frontier.
Our best checkpoint surpasses Qwen2-1.5B overall and approaches Qwen2.5-1.5B. Detailed accounting is discussed in~\Cref{sec:cost-estimation} and~\Cref{app:cost-assumptions}
As shown in \Cref{tab:pipeline-cost-summary}, the production schedule uses 438.8B Phase~1 tokens and 960.0B Phase~2 tokens. These stages total 1.4T scheduled tokens and 22,514 active-training GPU-hours, corresponding to a compute cost of about \$6.9K and 17.6 elapsed days.
The pipeline covers the data recipe, hardware infrastructure, supporting software, and training strategy.

\begin{samepage}
\noindent
\end{samepage}

We have released the following artifacts under the Apache 2.0 license unless otherwise noted separately; upstream data terms remain component-specific.

\begin{itemize}
    \item Model weights: \url{https://huggingface.co/thu-pacman/Puro-2B-Base} and variants (10 different versions of checkpoints in total for better transparency and reproducibility)
    \item Data manifests and materialized components\footnote{Different licenses may apply to different dataset components, some of which may not be permissive.}: \url{https://huggingface.co/datasets/thu-pacman/Puro-2B}
    \item Training code: \url{https://github.com/thu-pacman/Puro-Megatron}
    \item Data-processing code (\dataframework): \url{https://github.com/thu-pacman/Kaiyuan-Spark}
\end{itemize}

The recipe is built around five efficiency-oriented components spanning hardware, numerical precision, optimization, data ordering, and data selection.
\begin{enumerate}
    \item We use \textbf{consumer-grade RTX~5090 GPUs} as an accessible training platform that provides a favorable cost-efficiency trade-off in our setting~(\Cref{sec:recipe:infra}).
    \item Blockwise \textbf{FP8 training} reduces per-token execution time while
    maintaining comparable model quality to bfloat16 (BF16) training~(\Cref{sec:recipe:fp8}).
    \item We apply the \textbf{MuonH optimizer}, which extends the Muon optimizer with hyperball constraints on parameter weights and updates, together with a carefully designed learning rate schedule~(\Cref{sec:recipe:hyperball}). 
    \item \textbf{Curriculum Model Averaging (CMA)} organizes training over coarse-grained data chunks according to configured source-local preferences and averages selected checkpoints~\citep{luo2025learningratedecaywastes}~(\Cref{sec:recipe:curriculum}).
    \item Proxy experiments provide empirical signals for dataset selection and mixture design under a large candidate data pool~(\Cref{sec:data-recipe}).
\end{enumerate}

\begin{figure}[t!]
    \centering
    \begin{subfigure}{0.50\textwidth}
        \centering
        \includegraphics[width=\linewidth]{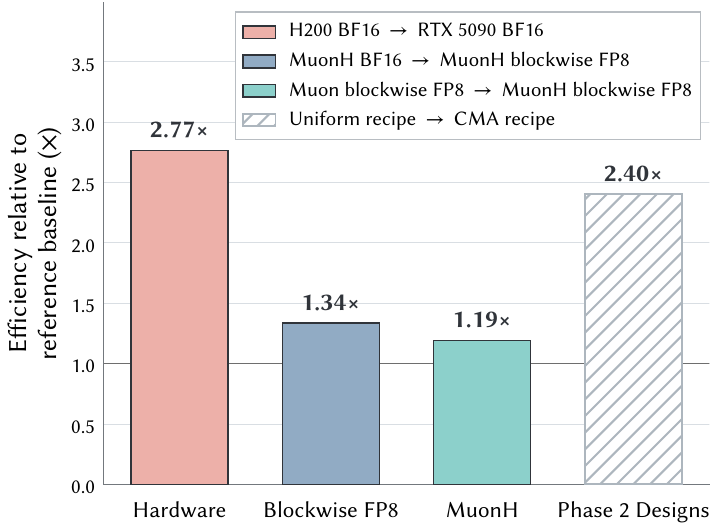}
        \caption{Relative Cost-Efficiency Improvement.}
    \end{subfigure}\hfill
    \begin{subfigure}{0.49\textwidth}
        \centering
        \includegraphics[width=\linewidth]{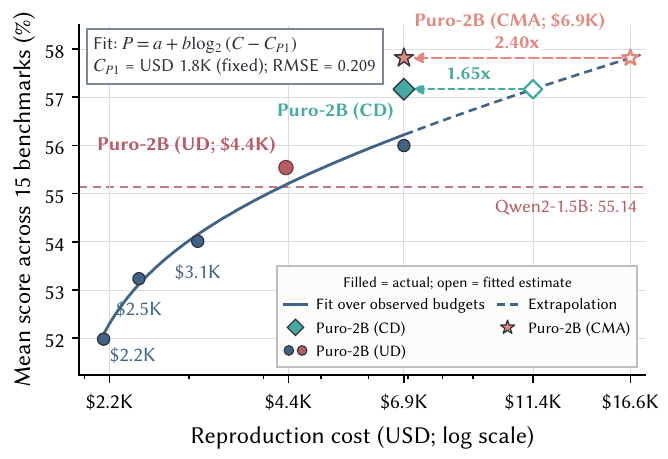}
        \caption{Puro Cost Scaling Law (scale-down)}
        \label{fig:renminbi-scaling-law}
    \end{subfigure}
    \caption{\textbf{(a)} Relative cost-efficiency improvement for hardware, FP8 precision, MuonH, and various Phase~2 designs. The gains come from different sources but are all converted into equivalent savings in USD. Each bar uses its own baseline, so the numbers should not be directly multiplied together. \textbf{(b)} Puro Cost Scaling Law is fitted from Phase~2 training results with uniform data at different token budgets. UD denotes it using \textbf{u}niform data ordering with LR \textbf{d}ecay. The \$4.4K UD checkpoint exceeds Qwen2-1.5B in average performance over the 15 benchmarks reported in~\Cref{tab:math_code_capabilities,tab:reasoning_knowledge}. At the \$6.9K reproduction cost, using a data \textbf{c}urriculum with LR \textbf{d}ecay (CD) in Phase~2 training achieves 1.65$\times$ cost-efficiency gains over the uniform data recipe. Incorporating more joint designs (\textbf{c}urriculum \textbf{m}odel \textbf{a}verage, CMA) in Phase~2 yields our canonical Puro-2B, achieving cost-efficiency gains of 2.40$\times$ relative to the uniform scaling curve~(\Cref{sec:recipe:curriculum}). The details are given in~\Cref{sec:recipe:curriculum,fig:phase2-uniform-budget-schedule}. Filled markers denote actual reproduction costs, while hollow markers denote fitted uniform-equivalent costs.
    See~\Cref{sec:recipe:curriculum} and~\Cref{tab:puro-scaling-checkpoints,tab:puro-scaling-evaluations} for details.}
    \label{fig:cost-saving-factors}
\end{figure}

Furthermore, we conduct targeted ablations over the first four design choices and report cost-efficiency estimates~(\Cref{fig:cost-saving-factors} and~\Cref{sec:cost-estimation}). Our systematic recipe covers the full efficiency stack of pretraining: data selection reduces the required token budget, MuonH and the CMA recipe improve token efficiency by extracting more capability from training tokens, FP8 improves hardware utilization by increasing computational throughput, and hardware selection reduces the cost per unit of computation.
Together, these choices optimize different bottlenecks across the training pipeline rather than targeting a single part.

Importantly, these components are not independent optimizations combined in isolation; they form a co-designed training system with interactions across different layers of the pipeline. For example, the Blackwell architecture of RTX~5090 provides the hardware foundation for efficient FP8 training. Both MuonH and CMA involve careful learning rate schedule design, and data curriculum relies on preprocessing decisions that determine the ordering of source-local data chunks. Therefore, the effectiveness of our recipe comes from the coordinated design of data, optimization, algorithmic, and system-level choices as an integrated approach to affordable pretraining.

In addition, we formulate the \textit{Puro Cost Scaling Law}, a recipe-specific scale-down relationship between rental-equivalent training budget and model capability. As illustrated in \Cref{fig:renminbi-scaling-law}, the law is designed for scale-down scenarios, serving users with limited compute and time budgets. There are three Phase~2 recipes used: \textbf{U}niform ordering with LR \textbf{D}ecay (UD), data \textbf{C}urriculum with LR \textbf{D}ecay (CD), and CMA. Their detailed ordering, learning rate schedule, and averaging choices are defined in~\Cref{sec:recipe:curriculum}. The scale-down recipe uses the UD recipe. We resume training from the same Phase~1 checkpoints and adopt different Phase~2 token budgets. We then evaluate the resulting checkpoints and fit the trend of model performance as a function of total training cost. This scaling curve provides a reference for estimating the model capability achievable under a limited budget with our recipe. Interestingly, the Puro Cost Scaling Law suggests that our recipe can cross the Qwen2-1.5B performance aggregate at a cost of USD~4.4K. Its accounting assumptions and recipe-level interpretation are detailed in~\Cref{sec:cost-estimation}. Moreover, the CD and CMA adoption can provide a further speedup compared to the UD scaling trend.

Beyond pretraining, the following post-training reveals that the difference between the CMA and UD Phase~2 recipes persists after matched supervised adaptation. In GSM8K-based SFT and Math\&Code SFT with replay, CMA-based SFT yields higher GSM8K accuracy in repeated runs.
In the Tulu-3 mixed-domain SFT setting, CMA-based SFT also improves the comprehensive capabilities and instruction following. Our Puro-2B recipes make these kinds of end-to-end comparisons possible at a meaningful scale and within an affordable budget.

In summary, our contributions are:
\begin{enumerate}
    \setlength{\itemsep}{0.2em}
    \setlength{\topsep}{0.2em}
    \item We provide a cost-effective, from-scratch pretraining recipe and train a Puro-2B model collection. The model in this collection exceeds Qwen2-1.5B at about \$4.4K, and its best model checkpoint approaches Qwen2.5-1.5B with \$6.9K. More broadly, \emph{our main contribution is to show that affordable, open pretraining is practical today}; we expect future efforts to push this frontier toward even lower cost and higher efficiency.
    \item We evaluate the main recipe choices: RTX~5090 infrastructure, blockwise FP8, MuonH, and curriculum model averaging with supporting ablation experiments. Moreover, we formulate the Puro Cost Scaling Law for the recipe's scale-down budget capability trend.
    \item We release the datasets, model weights, intermediate checkpoints, training configurations, and implementation provided to reproduce and inspect this pretraining recipe, along with a post-training case study to show how this recipe helps end-to-end comparison for one pretraining design.
\end{enumerate}

\section{Overview}

\begin{figure}[ht]
    \centering
    \resizebox{\linewidth}{!}{\input{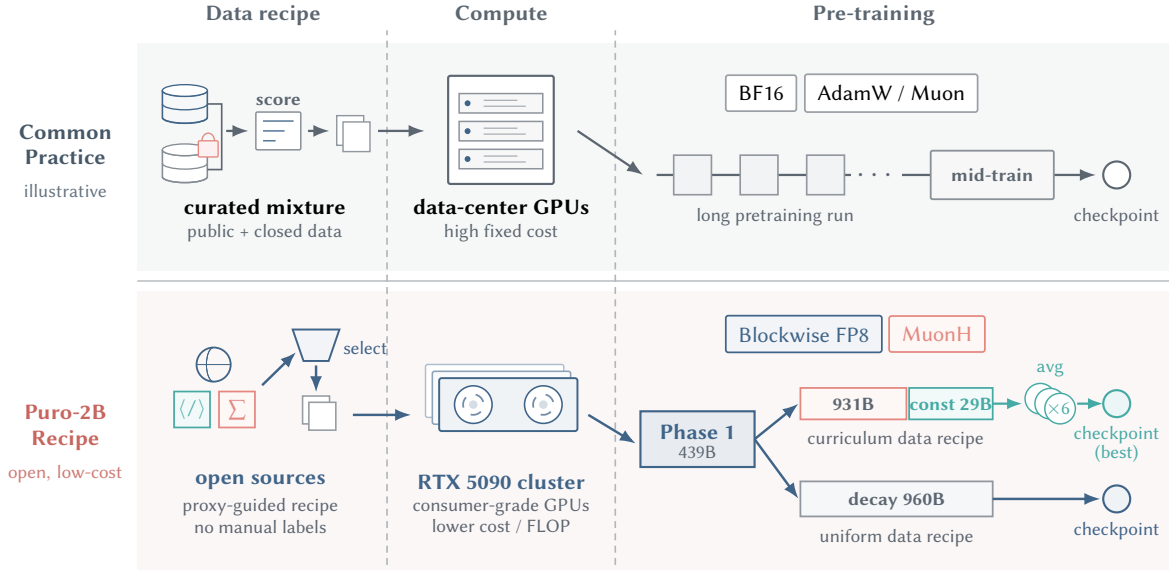}}
    \caption{
    \textbf{Puro-2B pipeline for \emph{poor}-lab design.} The upper row presents the work flow in common practice, and the lower row shows \emph{our open, low-cost Puro-2B recipe}. We collect and select publicly accessible datasets to build a training dataset~(\Cref{sec:data-recipe}), without elaborative data scoring and curation. We use an RTX~5090 cluster, which consists of consumer-grade GPUs and features higher cost-efficiency~(\Cref{sec:recipe:infra}), in contrast to costly data-center GPUs. Our pretraining process consists of two phases, and Phase~2 has two variants: uniform data recipe with LR decay (UD) and curriculum model averaging (CMA) with late constant-LR continuation and six-checkpoint averaging. Both variants use 960B tokens in Phase~2~(\Cref{sec:recipe:curriculum}). As a comparison, common practice goes through a long pretraining phase and a shorter mid-training phase with LR annealing to get the final checkpoint. To further improve cost efficiency, we adopt blockwise FP8 training~(\Cref{sec:recipe:fp8}) and Muon with Hyperball optimizer~(\Cref{sec:recipe:hyperball}) throughout Phase~1 and Phase~2.
    }
    \label{fig:pipeline-demo}
\end{figure}

Before discussing individual design choices in detail, we first provide an overview of the complete training recipe. The pipeline starts with constructing the training corpus from publicly accessible sources and selecting cost-efficient, widely accessible hardware. It then proceeds through two phases of pretraining, including our optimization and data-curriculum designs, followed by model averaging, post-training, and evaluation. We also define the cost-accounting protocol and its boundary to illustrate the reported training costs. The following sections then discuss the motivation, implementation, and evidence for each major design choice. The overall pipeline is illustrated in \Cref{fig:pipeline-demo}.

\subsection{Pipeline at a Glance}

\begin{table}[ht]
\centering
\small
\caption{
We summarize two Puro-2B run cost accounts in \Cref{fig:pareto-optimal}. Both runs share Phase~1 and use different Phase~2 recipe variants. The \$4.4K run uses uniform ordering with LR decay (UD), while the canonical \$6.9K run uses CMA~(\Cref{sec:recipe:curriculum}). We use 24~GPUs for Phase~1 and extend our compute resources to 96~GPUs in Phase~2. We compute the cost from GPU-hours and the unit compute cost specified in \Cref{sec:cost-account}.
}
\label{tab:pipeline-cost-summary}
\resizebox{\linewidth}{!}{%
\begin{tabular}{@{}lrrrrrrrrrr@{}}
\toprule
\multirow{2}{*}{Checkpoint} & \multicolumn{4}{c}{Phase~1 (shared)} & \multicolumn{4}{c}{Phase~2} & \multicolumn{2}{c}{Total} \\
\cmidrule(lr){2-5}\cmidrule(lr){6-9}\cmidrule(lr){10-11}
& GPUs & Tokens & Wall time & Cost & GPUs & Tokens & Wall time & Cost & Active GPU-h & Cost \\
\midrule
Puro-2B (UD; \$4.4K) & 24 & 438.84B & 10.43d & \$1.84K & 96 & 480.00B & 3.58d & \$2.53K & 14,262 & \$4.37K \\
Puro-2B (CMA; \$6.9K canonical) & 24 & 438.84B & 10.43d & \$1.84K & 96 & 959.99B & 7.16d & \$5.05K & 22,514 & \$6.89K \\
\bottomrule
\end{tabular}%
}
\end{table}

\paragraph{Model architecture and pretraining system~(\Cref{sec:recipe:infra,sec:recipe:fp8}).}
We train a dense decoder-only Transformer from scratch through two consecutive pretraining phases. The model adopts the Qwen3-1.7B architectural configuration, but unties the input embedding matrix from the output language-model head. This yields approximately 2B parameters in total. The Qwen3 architecture is familiar to the community and broadly supported. The model scale is also compatible with our replication budget. Our training is implemented in Megatron Core and adapted to an RTX~5090 cluster. To exploit the GPUs' Blackwell FP8 support, the main Transformer linear-layer GEMMs use blockwise FP8, while numerically sensitive operations and persistent training states, including master weights and optimizer states, remain in BF16 or FP32~(\Cref{sec:recipe:fp8}). Phase~1 runs on 24 GPUs, and Phase~2 continues from the Phase~1 endpoint on 96 GPUs~\footnote{Yanfu investments help extend our compute resources in Phase~2.}. 
The distributed parallel configuration is adjusted for the larger GPU count. The detailed low-precision compute flow, RTX~5090 system adaptations, and phase-specific parallel configurations are provided in~\Cref{sec:recipe:infra:rtx5090,sec:infra:parallel}.

\paragraph{Two-phase pretraining, curriculum, and optimization~(\Cref{sec:recipe:hyperball,sec:recipe:curriculum}).}
Using the architecture and training system above, we pretrain \sys in two phases. Phase~1 processes \PhaseOneWandBPhaseTokens tokens using the Phase~1 mixture, while Phase~2 consumes \PhaseTwoWandBPhaseTokens tokens in total. The canonical run applies a data curriculum over the Phase~2 data pool~(\Cref{sec:recipe:curriculum}), to present the more preferred portion of each scored source later in training. Rather than defining a global quality score across datasets, we primarily use the quality labels provided by source datasets. Within each dataset with usable score labels, examples are ordered from lower to higher score and partitioned into coarse chunks. Datasets without usable scores are instead randomly ordered and partitioned in the same way. Chunks from different sources are then aligned by their normalized within-source ranks, so that Phase~2 progresses from lower to higher configured ranks while approximately preserving the cross-component mixture. An illustration of this curriculum design is provided in~\Cref{fig:curriculum-model-average}.

To preserve the influence of data presented late in the curriculum, we jointly design the late-stage learning rate schedule and checkpoint averaging, following the curriculum-aware averaging principle of \citet{luo2025learningratedecaywastes}, and use the averaged weights as the final model. Phase~2 therefore has recipe variants: \emph{UD} uses a uniform data ordering with LR decay, \emph{CD} uses the component-local data curriculum with LR decay, and \emph{CMA} adds the late constant-LR continuation and six-checkpoint average. For comparison, the UD variant uses the same Phase~1 checkpoint and Phase~2 pool but globally reshuffles the Phase~2 data instead of following the curriculum ordering.

Across both phases, we use Hyperball optimization for selected approximately
scale-invariant matrices.  MuonH updates these matrices and projects each one
back to its initial Frobenius radius after every step, while AdamW updates the
remaining parameters~(\Cref{sec:recipe:hyperball}).  Both parameter groups share a base LR schedule. MuonH group uses a 10 times base LR schedule as Hyperball weight LR. The base schedule follows a power schedule~\citep{shen2024power} in Phase~1 and
continues from its terminal Phase~1 value with linear decay along the main
Phase~2 trajectory.  The late CMA continuation instead holds the base LR fixed
at its resume-point value before checkpoint
averaging~(\Cref{sec:recipe:curriculum}).  Both phases use a sequence length of 4,096
tokens and a global batch size of 1,536 sequences.  The optimizer
configuration, learning rate multipliers, curriculum construction, transition
schedule, checkpoint-averaging rule, and exact token accounting
are reported
    in~\Cref{sec:recipe:hyperball,sec:recipe:curriculum}, \Cref{app:training-details}, and~\Cref{tab:prom-2b-training-setup}.

\paragraph{Data acquisition, filtering, and mixture~(\Cref{sec:data-recipe}).}
We construct the training corpus from publicly accessible datasets with licenses or terms that permit their use, enabling the corpus to be reproducibly reconstructed from the original sources. The corpus spans web text, code, and mathematics, and also includes synthetic examples. We first use our \dataframework framework~(\Cref{sec:recipe:data:puro-spark}) to deduplicate the large-scale web portion. Because the available data far exceeds our training budget, we must select source datasets, filter samples within each source, and combine the retained portions to improve overall data utility and match the desired capability profile. However, it is difficult to design a single sample-level quality score that is comparable across domains, apply it to every sample, and use it to determine which samples to retain. Instead, we run proxy experiments on candidate sources and representative data slices~(\Cref{sec:recipe:data:proxy-feature}). Specifically, we train controlled small-scale models on these sources or slices and evaluate the resulting models on downstream tasks. The resulting vector of benchmark scores defines the proxy profile of each source or slice, which guides source selection and mixture design~(\Cref{sec:recipe:data:selection}). For example, when stronger mathematical capability is desired, we can assign higher weights to datasets whose proxy profiles show stronger performance on mathematics benchmarks. Some open-source datasets also provide sample-level quality scores. Although these scores are not directly comparable across sources, they can support within-source filtering. For each such dataset, we sample proxy slices from different score ranges, compare their proxy profiles, and use the results to select a dataset-specific threshold for retaining high-scoring samples. The final domain-level recipe for the two pretraining phases is summarized in~\Cref{tab:data-recipe-summary} and~\Cref{fig:data-domain-composition}, while~\Cref{tab:data-components-detail} reports family-level retained token counts, selection modes, assigned phases, source identifiers, and license terms rather than a complete component-level manifest.

\paragraph{Evaluation and post-training~(\Cref{sec:recipe:posttrain,sec:evaluation}).}
Our evaluation combines a broad pretrained-model comparison with controlled tests of downstream adaptation.  We first evaluate \sys against open-weight and reproducible open-recipe models of comparable scale on general reasoning, mathematics, and coding.  Under the report's common budget accounting, \sys surpasses Qwen2-1.5B and Gemma-2-2B while using less than one sixth of the stated training cost of comparable open recipes.  We then use SFT as a probe of how the Phase~2 recipes transfer through post-training.  Within each paired experiment, the CMA and UD initializations receive identical supervised data and optimization; mathematics checkpoints use one frozen generation and extraction protocol, while broad tasks use benchmark-appropriate prompts and scorers.  The three settings are GSM8K-based SFT, Math\&Code SFT with replay, and Tulu-3 mixed-domain SFT, with complete definitions and results in~\Cref{sec:recipe:posttrain} and~\Cref{app:posttrain-details}.

\subsection{Reproduction-Cost Boundary}
\label{sec:cost-account}

Throughout this report, reproduction cost refers specifically to the compute cost of rerunning the finalized two-phase pretraining recipe once. The accounting boundary begins after the phase-specific training shards have been materialized and includes only the compute time of the Phase~1 and Phase~2 production runs.  We report measured GPU-hours separately and convert them using the normalized RTX~5090 rental-equivalent rate specified in~\Cref{sec:cost-estimation} and~\Cref{app:cost-assumptions}.  The ledger retains CNY(RMB) for traceability, while the main cost figures use the corresponding USD values.  Both are reproducible cost proxies under the stated rate, not the authors' total cash expenditure.

We do not estimate costs outside these production pretraining runs.  The headline therefore excludes data acquisition and preprocessing, proxy-model experiments, scaling and ablation studies, failed or exploratory runs, post-training, evaluation, checkpoint averaging, and research labor. It also excludes non-accelerator resources such as CPU processing, storage, and networking, as well as taxes, depreciation, and other ownership costs not represented by the rental-equivalent rate.  These exclusions do not imply that the activities are free; they mean that the reported number should be interpreted as a narrowly defined marginal accelerator cost for reproducing the final pretraining run, rather than as the total cost of developing the model or reconstructing every artifact from scratch.

\section{Training Recipe}
\label{sec:training-recipe}
We organize the training recipe as follows. We first describe the infrastructure design that supports our training setup. We then present the pretraining recipe, focusing on optimization and data curriculum. Next, we explain how the training dataset is constructed. Finally, we present a post-training case study to demonstrate how our recipe enables end-to-end approach comparison.

\subsection{Training Infrastructure}
\label{sec:recipe:infra}

\subsubsection{RTX~5090 as a Cost-Effective GPU Choice}
\label{sec:recipe:infra:rtx5090}

To make our recipe cost-efficient and the hardware accessible to the broader community, we made a non-standard infrastructure choice: we run all pretraining and post-training on a cluster equipped with consumer-grade GPUs (RTX~5090), instead of renting conventional data-center-oriented accelerators. Although data-center GPUs are generally expected to be more stable and compute-efficient, our cost analysis showed that RTX~5090 is a better fit for this workload.

\begin{table}[ht]
\centering
\renewcommand{\tablename}{Table}
\begin{adjustbox}{max width=\textwidth}
\begin{narrowtalltblr}[
  caption = {Peak hardware capability and cost efficiency of candidate NVIDIA GPUs.},
  label = {tab:gpu-cost-efficiency},
  note{a} = {All performance numbers are from official NVIDIA documentation / datasheets / blog posts~\cite{nvidia2022hopper,nvidia2024h200, nvidia2025blackwell,nvidia2025blackwellpro} and refer to dense matrix operations with FP32 accumulation.},
  note{b} = {The prices do not include CPU / storage / network, as they are marginal and often free on GPU rental platforms. A100, H200, and RTX PRO 6000 prices were collected from a public GPU rental platform on 16 Aug 2026. RTX~5090 has no widely available public rental channel due to NVIDIA EULA restrictions; we estimate its effective price by amortizing our hardware and electricity costs over five years. An 8-GPU node costs approximately 12,000 CNY (\$1763) per month for our setup. Some private compute providers confirmed similar rental pricing for RTX~5090.},
  note{c} = {Hopper FP8 Tensor Cores use a reduced-precision path for nominal FP32 accumulation, with 22 effective precision bits~\cite{zhang2024sageattention2}, which may affect numerical accuracy.},
  note{d} = {PCIe Gen5 $\times$ 16 provides approximately 64 GB/s of theoretical bandwidth per direction, or approximately 128 GB/s of aggregate bidirectional bandwidth.
 Driver tweaks needed to achieve this bandwidth are discussed in~\Cref{sec:recipe:infra:hardware}.},
]{
  colspec = {l r r r c c c c c r r r},
  rowhead = 2,
  row{1-2} = {font=\bfseries, bg=gray!10},
  cell{1}{1} = {r=2}{l},
  cell{1}{2} = {c=3}{c},
  cell{1}{5} = {r=2}{c},
  cell{1}{6} = {r=2}{c},
  cell{1}{7} = {r=2}{c},
  cell{1}{8} = {r=2}{c},
  cell{1}{9} = {r=2}{c},
  cell{1}{10} = {c=3}{c},
  hline{3} = {0.5pt},
}
GPU Model & \shortstack {Peak tensor performance \\ (TFLOPS)\TblrNote{a}} & & & \shortstack{Memory\\capacity} & \shortstack{Memory\\bandwidth} & TDP & \shortstack{Intra-node\\P2P BW\\(Bi-Di)} & \shortstack{Single-GPU\\price\TblrNote{b}} & \shortstack{Peak compute/\$\\(EFLOP/USD)} & & \\
& BF16 & FP8 & NVFP4 & & & & & & BF16 & FP8 & FP4 \\
A100 SXM4 & 312 & N/A & N/A & 80 GB & 2 TB/s & 400 W & 600 GB/s & \$1.79/h & 0.63 & N/A & N/A \\
H200 SXM5 & 989.5 & 1979 \TblrNote{c} & N/A & 141 GB & 4.8 TB/s & 700 W & 900 GB/s & \$4.00/h & 0.89 & 1.78 & N/A \\
RTX PRO 6000 & 503.8 & 1007.6 & 2015.2 & 96 GB & 1.8 TB/s & 600 W & 128 GB/s \TblrNote{d} & \$1.89/h & 0.96 & 1.92 & 3.85 \\
RTX 5090 & 209.5 & 419 & 1676 & 32 GB & 1.8 TB/s & 575 W & 128 GB/s \TblrNote{d} & \textbf{\$0.31/h} & \textbf{2.43} & \textbf{4.87} & \textbf{19.46} \\
\end{narrowtalltblr}
\end{adjustbox}
\end{table}

\Cref{tab:gpu-cost-efficiency} shows the features and prices of different generations of NVIDIA GPUs. RTX~5090 has a clear gap in absolute peak throughput compared with data-center GPUs, especially because its Tensor Cores are artificially limited to half of actual peak performance when using FP32 accumulation. However, its much lower effective price gives substantially higher compute per dollar: its BF16 and FP8 cost efficiency is about 2.7$\times$ that of H200. Moreover, because policy and cost constraints made Blackwell data-center accelerators such as GB200 unavailable to us, RTX~5090 is also the most practical option with FP4 support.

RTX~5090 still has significant hardware limitations, including smaller memory capacity and the lack of NVLink for high-bandwidth GPU-to-GPU scaling. These drawbacks cannot be ignored. Nevertheless, our target model is small enough that, with the hardware configuration and parallel training setup described below, the RTX~5090 cluster reaches a mixed-precision effective MFU of approximately 73\% under the stated precision-weighted peak convention. \Cref{sec:infra:parallel} defines the convention and measurement boundary; this is not a BF16-only full-scale result. Its cost advantage therefore remains valid for our setting.

\subsubsection{Hardware Setup and Tweaks}
\label{sec:recipe:infra:hardware}

Our training cluster consists of multiple x86 GPU servers. Each server is equipped with dual CPUs, eight RTX 5090 GPUs (four in each CPU socket), and sufficient host memory. The performance numbers in~\Cref{sec:recipe:infra:rtx5090} are single-GPU metrics; for multi-GPU training, both intra-node P2P bandwidth and inter-node network bandwidth must be considered to sustain training throughput.

\paragraph{Intra-node bandwidth.}

Since RTX~5090 does not provide NVLink, the only intra-node interconnect is PCIe. NVIDIA disables PCIe P2P on non-data-center GPUs, which forces GPU communication traffic to be staged through host memory, also known as ``ping-pong'' transfer, which can substantially reduce communication efficiency. However, this restriction appears to be enforced by the NVIDIA GPU driver rather than the limitation of GPU hardware. We used a modified version of the open-source NVIDIA driver~\cite{aikitoria2026driver}, along with the necessary platform configuration changes~\cite{2026harry5090p2p}, including disabling IOMMU and PCIe ACS as well as adjusting NPS (NUMA Per Socket, also known as Sub-NUMA Clustering on some platforms) to enable GPU P2P on our servers.

On our hardware, enabling P2P improved one-way bandwidth from 31.5 GB/s to 56 GB/s. The former is bounded by half of the PCIe 5.0 $\times$16 bandwidth, i.e., 32 GB/s, because ping-pong transfer requires two host-memory copies; the latter approaches the 64 GB/s PCIe link limit. Bidirectional bandwidth increased from 32 GB/s to 111 GB/s, compared with theoretical limits of 128 GB/s. Communication latency decreased drastically from 14.3 $\mu$s to 0.4 $\mu$s.
The eight-GPU \texttt{AllReduce} bandwidth over PCIe, measured as \texttt{busbw} reported by \texttt{nccl-tests}, increased from 14.75 GB/s to 27.34 GB/s. When the test was restricted to the four GPUs attached to a single CPU socket, the gain was larger, from 16.33 GB/s to 46.31 GB/s, about 2.8$\times$. Adding a PLX PCIe switch between the GPUs and the root complex might further improve P2P bandwidth toward the link limit; this is a prospective topology change, not a configuration used in the reported run.

\paragraph{Inter-node bandwidth.}

To scale training beyond a single server, the inter-node network must provide bandwidth comparable to the intra-node communication path. We set up a 400 Gbps InfiniBand network among our servers, i.e., 100 GB/s bidirectional bandwidth: one dual-port Mellanox ConnectX-7 host card adapter (HCA) is installed in each server, with both ports connected to a 200 Gbps InfiniBand HDR switch. Half of the GPUs in the server could communicate with the HCA directly, while the other half must route through the inter-socket link to reach the HCA.

Even after PCIe P2P is enabled, non-data-center NVIDIA GPUs still do not support GPUDirect RDMA (shortened as GDR), another performance-critical feature for inter-node communication. GDR builds on P2P and allows a GPU to access remote GPU memory over RDMA NICs without staging through host memory. Through testing, we found that this restriction is also software-enforced: a minimal binary modification to the CUDA user-space driver can enable GDR on RTX~5090~\cite{2026harry5090gdr}. Due to EULA constraints, we are not able to disclose the details in this report.
When this configuration is unavailable, the same training recipe can be run with the stock NVIDIA driver, falling back to the standard communication path at lower inter-node bandwidth. In end-to-end tests, enabling GDR improved the 24-GPU (3 servers) \texttt{AllReduce} bandwidth, measured as \texttt{busbw}, from approximately 8.87 GB/s to 19.93 GB/s. We believe that if a more balanced network topology is used (e.g., one InfiniBand HCA on each socket), the performance gain could be even higher.

\paragraph{Caveats.}

These modifications should be used with caution. First, modifying drivers is unsupported by the vendor and must be done at the user's own risk. Second, these changes cannot exceed the physical PCIe bandwidth limit. Third, P2P and GDR should not be enabled blindly; the optimal setting depends on the actual hardware topology and platform capability. For example, if the PCIe root complex of a CPU has limited peer-switching capability (which is often the case on low-end models), heavy P2P traffic among many devices may become congested, and host-memory ping-pong transfer can be faster.

\subsubsection{Efficient Training System\label{sec:infra:parallel}}

We build our training system on Megatron Core~\cite{shoeybi2019megatronlm}, specifically release v0.16.0~\cite{nvidia2026megatroncore0160}, together with its Transformer Engine dependency. This software stack provides native support for blockwise FP8 training and the Muon optimizer, making it a suitable basis for a model at our scale. Nevertheless, the combination of consumer GPUs, a mixed-precision training recipe, and our model architecture still requires careful system-level tuning.

\paragraph{Communication-aware parallelism.}

The limited communication bandwidth of RTX~5090 motivates us to use only parallel dimensions with modest communication demands. We combine data parallelism (DP), whose dominant collectives are gradient \texttt{ReduceScatter} and parameter \texttt{AllGather} around optimization, with pipeline parallelism (PP), which exchanges hidden-state tensors only at stage boundaries~\cite{narayanan2021efficientmegatron}. We do not use tensor parallelism (TP), which introduces frequent collectives within every Transformer layer. Compared with data-center-class systems, our communication disadvantage is substantially more pronounced within a node. We therefore depart from Megatron Core's default rank ordering and use a \texttt{pp-dp} ordering that places each PP group on topologically closer GPUs within a node, thereby reducing sensitivity to communication latency.

\paragraph{Appropriate micro-batch size.}

FP8 kernels complete their arithmetic more quickly while introducing additional quantization operations, so they become memory-bound more readily than BF16 kernels. This effect is amplified by the relatively small hidden dimension of our model. Increasing the micro-batch size (MBS) improves the arithmetic intensity and Tensor Core utilization, but an excessive MBS exhausts device memory: DP does not partition activations; and 1F1B pipeline schedule~\cite{narayanan2021efficientmegatron} helps control the number of simultaneously resident microbatches, but it does not eliminate the per-microbatch activation-memory growth caused by increasing MBS.
We therefore extend Megatron-LM's analytical FLOP estimator to report the invocation shapes of GEMM and FlashAttention kernels. For each shape induced by our model configuration, we benchmark the corresponding Transformer Engine kernel and select the smallest MBS beyond which achieved FLOP/s no longer increases materially, analogous to locating the knee of a roofline curve.

\paragraph{Overcoming load imbalance.}

Although our model is relatively small, its vocabulary is large; consequently, the embedding and LM head account for a non-negligible fraction of both computation and memory. FP8 GEMMs accelerate the internal Transformer layers, further increasing their performance gap with the LM head. As a result, the LM head alone entails computation comparable to several Transformer layers, creating substantial pipeline imbalance. Guided by our extended FLOP estimator, which attributes theoretical work by layer, submodule, operator, and precision, we assign fewer Transformer layers to the stage containing the LM head. This balances pipeline computation at the cost of greater memory pressure on the first stage. Tensors assigned to Muon are preferably kept intact rather than sharded. We replace the zigzag placement policy of the layer-wise distributed optimizer in this version of Megatron Core, which over-concentrates the embedding and LM-head tensors, with a memory-aware allocation~\cite{xu2026deepseek}: Muon tensors are greedily assigned by memory footprint like bin packing, after which flattened Adam tensors fill the remaining per-device capacity as evenly as possible. This hybrid placement balances optimizer memory across GPUs and is important for fitting the training state into the 32~GB memory available on each RTX~5090.

These principles yield the same fastest configuration as an exhaustive enumeration of the candidate parallel strategies. Together with the hardware optimizations described above and standard Megatron Core options such as the \texttt{[T,H,D]} layout for packed-sequence attention, the Phase~1 production run sustains a median \PhaseOneWandBMedianThroughput~TFLOP/s per GPU with global batch size \PhaseOneWandBGlobalBatchSize on \PhaseOneWandBWorldSize GPUs across three nodes. The best configuration is MBS${}=2$, PP${}=2$ with layout \texttt{(18|10)}\footnote{The PP layout lists the numbers of Transformer layers assigned to successive stages. In \texttt{(18|10)}, the first stage contains the embedding and the first 18 Transformer layers, while the second contains the remaining 10 layers and the LM head.}, and DP${}=12$. The measured throughput corresponds to an equivalent MFU\footnote{Model FLOPs utilization (MFU) is conventionally defined as achieved model FLOP/s divided by the accelerator's peak FLOP/s~\cite{korthikanti2023reducing}. For mixed-precision training, we weight the peak Tensor Core performance by the fraction of theoretical work at each precision. In our recipe, FP8 accounts for 72\% of the theoretical Tensor Core operations and BF16 for the remaining 28\%, giving an effective peak of $P_{\mathrm{eff}}=(0.72/419+0.28/209.5)^{-1}\approx327$~TFLOP/s per GPU.} of approximately 73\%. For larger-scale Phase~2 training, strong scaling is limited by Amdahl's law, and gradient synchronization accounts for a larger fraction of the step time. We consequently use PP${}=4$ with layout \texttt{(9|9|9|1)} and DP${}=24$ on \PhaseTwoWandBWorldSize GPUs, while still achieving median throughput \PhaseTwoWandBMedianThroughput~TFLOP/s per GPU. The run-level setup and final losses are reported in~\Cref{tab:prom-2b-training-setup}.

\subsection{FP8 Mixed-Precision Training}
\label{sec:recipe:fp8}

We use FP8 mixed precision from random initialization onward, without a BF16 warm-up stage or a later precision switch, using our training stack~\cite{nvidia2026megatroncore0160} and its blockwise FP8 support~\cite{nvidia2026transformerenginefp8}.  The training state and numerically sensitive operations retain the standard BF16/FP32 mixed-precision path, while Transformer linear layers use E4M3 operands.  FP8 is therefore an online compute and activation-storage format rather than the persistent dtype of the whole model.

E4M3~\cite{micikevicius2022fp8formats} provides more mantissa bits than E5M2 but has a narrower dynamic range.  With one scale for an entire tensor, a few outliers can enlarge the quantization interval and reduce the effective precision of ordinary values~\cite{deepseekai2025deepseekv3technicalreport}.  We instead compute scales online for local groups: activations and activation gradients use one-dimensional groups of 128 consecutive values along the GEMM reduction dimension, while weights use two-dimensional $128\times128$ blocks~\cite{nvidia2026megatronmoe}.  This fine-grained numerical design follows the recipe introduced and validated at scale by DeepSeek-V3~\cite{deepseekai2025deepseekv3technicalreport}.

On our RTX~5090 (SM~120)~\cite{nvidia2025blackwell}, each logical block scale is constrained to a power of two and therefore has the exponent-only expressiveness of E8M0.  Transformer Engine~\cite{nvidia2026transformerenginefp8} implements these blocks through Blackwell's native MXFP8 path.
Thus, our logical block sizes follow DeepSeek-V3~\cite{deepseekai2025deepseekv3technicalreport}, while the scaling-factor representation and execution path follow MXFP8; the fact that DeepSeek-V3 was trained on Hopper GPUs may account for this implementation difference.

The complete precision flow is shown in~\Cref{fig:fp8-precision-flow}. Here, Transformer linear layers include the QKV projections, attention output projection, and MLP projections; their Fprop, Dgrad, and Wgrad GEMMs all use the FP8 path~\cite{deepseekai2025deepseekv3technicalreport}.  Core attention refers only to the FlashAttention/SDPA computation between these projections and remains in BF16~\cite{nvidia2026megatronmoe}.  A linear layer quantizes its BF16 inputs and weights immediately before GEMM and materializes a BF16 output for the surrounding model.  Its quantized input is retained for Wgrad, reducing the dominant saved-activation footprint~\cite{nvidia2026megatronmoe}.  FP8-accelerated operations account for 72\% of the overall computation; the cost impact and net benefit of FP8 are analyzed in \Cref{fig:cost-saving-factors}.

The model-weight portion of each training checkpoint consequently remains BF16; the desired release or deployment representation is produced by a separate post-training conversion.  In the matched 20-token-per-parameter scaling ladder, blockwise FP8 increases validation loss by 0.0031--0.0039 relative to BF16 across the five tested model sizes.  The shared-shape fit maps this difference to \FPEightPrecisionRetention BF16-equivalent compute retention.  At the 1.7B scale, however, blockwise FP8 improves median training throughput by \FPEightThroughputSpeedup, yielding a quality-adjusted net gain of \FPEightNetSpeedup.  The complete setup, fitting procedure, and size-wise results are reported in~\Cref{sec:recipe:cost:fp8-save}.

\subsection{Hyperball Optimization}
\label{sec:recipe:hyperball}

\begin{figure}[ht]
    \centering
    \resizebox{0.96\linewidth}{!}{\definecolor{hbblue}{HTML}{3F6488}
\definecolor{hborange}{HTML}{E4877F}
\definecolor{hbgreen}{HTML}{46AAA4}
\definecolor{hbgray}{HTML}{5C6670}

\begin{tikzpicture}[
  x=1cm,
  y=1cm,
  font=\sffamily\scriptsize,
  vector/.style={-{Latex[length=1.7mm]}, line width=1.05pt},
  guide/.style={line width=0.65pt, dash pattern=on 2pt off 1.5pt},
  panel/.style={draw=hbgray!60, line width=0.72pt, rounded corners=1.5pt},
  paneltitle/.style={font=\sffamily\small, text=hbgray, fill=white, inner sep=1.5pt},
  note/.style={font=\sffamily\scriptsize, text=hbgray!80, align=center},
]

\draw[panel] (0.15,0.15) rectangle (6.65,3.55);
\node[paneltitle, anchor=west] at (0.38,3.55) {Muon + weight decay: radius evolves};
\begin{scope}[yshift=0.15cm]
\coordinate (OL) at (2.05,1.52);
\coordinate (WL) at (2.36,2.42);
\coordinate (SL) at (2.28,2.19);
\coordinate (NL) at (3.35,2.03);
\draw[guide, draw=hbblue] (OL) circle (0.95cm);
\draw[guide, draw=hbgreen] (OL) circle (1.40cm);
\fill[hbgray] (OL) circle (1.15pt) node[below left, text=hbgray] {$0$};
\draw[vector, draw=hbblue] (OL) -- (WL)
  node[pos=0.55, left, text=hbblue] {$W_t$};
\fill[hbblue] (WL) circle (1.4pt);
\draw[vector, draw=hbgray] (WL) -- (SL);
\node[text=hbgray, anchor=east, fill=white, inner sep=0.7pt] at (2.14,2.57) {WD};
\draw[vector, draw=hborange] (SL) -- (NL);
\node[text=hborange, fill=white, inner sep=0.7pt] at (3.15,2.54) {$-\ell_tu_t$};
\draw[vector, draw=hbgreen] (OL) -- (NL)
  node[pos=0.73, below, text=hbgreen] {$W_{t+1}$};
\fill[hbgreen] (NL) circle (1.4pt);
\node[note, anchor=west, text width=2.20cm] at (4.05,2.10)
  {soft radial control\\$R_{t+1}\neq R_t$};
\node[text=hbblue, fill=white, inner sep=0.6pt] at (1.28,0.91) {$R_t$};
\node[text=hbgreen, fill=white, inner sep=0.6pt] at (1.13,0.54) {$R_{t+1}$};
\end{scope}

\draw[panel] (7.05,0.15) rectangle (13.55,3.55);
\node[paneltitle, anchor=west] at (7.28,3.55) {MuonH: fixed-radius projection};
\coordinate (OR) at (8.95,1.52);
\coordinate (WR) at (9.26,2.42);
\coordinate (TR) at (10.40,2.15);
\coordinate (NR) at (9.822,1.899);
\draw[draw=hbblue, line width=0.9pt] (OR) circle (0.95cm);
\fill[hbgray] (OR) circle (1.15pt) node[below left, text=hbgray] {$0$};
\draw[vector, draw=hbblue] (OR) -- (WR)
  node[pos=0.55, left, text=hbblue] {$W_t$};
\fill[hbblue] (WR) circle (1.4pt);
\draw[vector, draw=hborange] (WR) -- (TR);
\node[text=hborange, fill=white, inner sep=0.7pt] at (9.76,2.62) {$-\rho_tR\widehat u_t$};
\fill[hborange] (TR) circle (1.4pt);
\draw[vector, draw=hbgray] (TR) -- (NR);
\node[text=hbgray, anchor=west, fill=white, inner sep=0.7pt] at (10.12,1.86) {projection};
\draw[vector, draw=hbgreen] (OR) -- (NR);
\node[text=hbgreen, fill=white, inner sep=0.7pt] at (9.55,1.18) {$W_{t+1}$};
\fill[hbgreen] (NR) circle (1.4pt);
\node[note, anchor=west, text width=2.25cm] at (11.05,2.02)
  {fixed radius\\$\lVert W_{t+1}\rVert_F=R$};
\node[text=hbblue, fill=white, inner sep=0.6pt] at (8.18,0.91) {$R$};

\end{tikzpicture}}
    \\[1em]
    {\small
    \setlength{\tabcolsep}{4pt}
    \begin{tabular}{@{}p{0.16\linewidth}p{0.43\linewidth}p{0.30\linewidth}@{}}
        \toprule
        & \textbf{Muon + weight decay} & \textbf{MuonH} \\
        \midrule
        Weight radius & Evolves under weight decay and optimizer updates & Fixed at $R=\lVert W_0\rVert_F$ after each step \\
        Displacement norm & $\eta_t\lVert u_t\rVert_F$ (state dependent) & $\rho_tR$ (scheduled) \\
        Effective LR & $\eta_t\lVert u_t\rVert_F/\lVert W_t\rVert_F$ (indirect) & $\rho_t$ (explicit) \\
        \bottomrule
    \end{tabular}}
    \caption{Effective-LR control for one matrix. Ordinary Muon (left) applies a scalar learning rate to a state-dependent update norm, so both its effective LR and matrix radius emerge indirectly. MuonH (right) normalizes the update, prescribes a displacement of length $\rho_tR$, and projects the result back to the fixed-radius sphere.}
    \label{fig:hyperball-principle}
\end{figure}
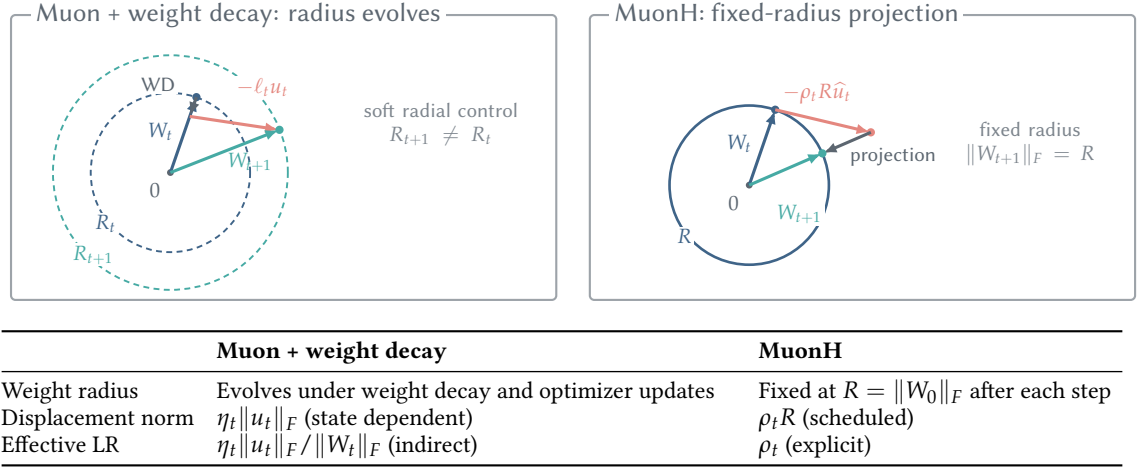

In our recipe, we adopt Muon with the Hyperball wrapper (MuonH), introduced by~\citet{wen2026fantastic2}, as the training optimizer for selected weight matrices. We apply MuonH to approximately scale-invariant weights, including attention and MLP matrices, while embeddings, normalization layers, the language-model head, and the remaining parameters use AdamW. Following~\citet{wen2026fantastic2}, MuonH normalizes the Muon update and constrains each wrapped matrix to a fixed-radius sphere.
More specifically, for each wrapped matrix $W_t$, MuonH fixes the radius at
$R=\lVert W_0\rVert_F$ and normalizes the Muon update $u_t$ as
$\widehat u_t=u_t/\lVert u_t\rVert_F$. It then applies
\begin{equation}
\widetilde W_{t+1}
=
W_t-\eta_tR\widehat u_t,
\qquad
W_{t+1}
=
R\,\operatorname{Normalize}(\widetilde W_{t+1}),
\label{eq:hyperball-update}
\end{equation}
where $\operatorname{Normalize}(A)=A/\lVert A\rVert_F$. The first operation normalizes the update scale, while the second projects the updated weight matrix back to its initial Frobenius radius. The illustration of MuonH optimizer is shown in \Cref{fig:hyperball-principle}.

We vary the learning rate $\eta_t$ according to a two-phase learning rate schedule. The design of this schedule is motivated by an analysis of how the loss gap between MuonH and Muon arises from the perspective of effective learning rate, as we elaborate below.

\subsubsection{Effective Learning Rate }

First, we briefly revisit the rationale behind Hyperball optimization and explain how it makes the effective LR schedule explicit~\citep{wen2026fantastic2}. We then examine the loss gap between ordinary Muon and MuonH runs and find that the effective LR schedule strongly affects the loss convergence rate. This observation leads us to focus on learning rate schedule design for the MuonH optimizer. 

\paragraph{Scale invariance motivates the effective learning rate.}
Hyperball is motivated by approximately scale-invariant parameter matrices. For a selected weight matrix $W$, we call the parameter approximately \emph{scale invariant} when positively rescaling it has little effect on the model function or loss,
\begin{equation}
\mathcal{L}(cW) \approx \mathcal{L}(W),
\qquad c>0,
\end{equation}
with the remaining parameters held fixed~\citep{wen2026fantastic2}. Normalization in modern Transformers makes this approximation relevant for several attention and MLP weight matrices. In this regime, the update magnitude is naturally interpreted relative to the current weight scale. We therefore consider the matrix-wise effective LR, also called the \emph{effective learning rate} (ELR). For ordinary Muon, with Muon update $u_t$, scalar learning rate $\eta_t$, and decoupled weight-decay coefficient $\lambda$,
\begin{equation}
W_{t+1}
=
\bigl(1-\lambda\eta_t\bigr)W_t-\eta_tu_t,
\label{eq:muon-weight-decay}
\end{equation}
and, ignoring the radial shrinkage from weight decay, the effective LR is
\begin{equation}
\rho_t(W)
=
\frac{\eta_t\lVert u_t\rVert_F}
{\lVert W_t\rVert_F}.
\label{eq:relative-update}
\end{equation}
Thus, ordinary Muon's effective LR is not determined by the scalar learning rate alone, because both the weight norm and the Muon-update norm evolve during training.

\paragraph{MuonH makes the effective LR schedule explicit.}
Under the Hyperball update in~\Cref{eq:hyperball-update}, the pre-projection displacement $\eta_t R\widehat u_t$ has norm $\eta_tR$, while the wrapped matrix $W_t$ has norm $R$. Its effective LR is therefore $\rho_t(W)=\eta_t\lVert u_t\rVert_F/\lVert W_t\rVert_F=\eta_tR/R=\eta_t$, numerically equal to the Hyperball weight learning rate, and the subsequent projection restores the fixed radius. Thus, ordinary Muon induces its effective LR indirectly through the scalar learning rate and evolving norms, whereas \emph{MuonH directly controls the effective LR schedule through its learning rate schedule.} This explicit control is the property of Hyperball that is most relevant to our subsequent schedule design.

In our experiments, we use different learning rates for the MuonH-wrapped and AdamW-optimized parameter groups. We define a base learning rate schedule $\eta_t^{\mathrm{base}}$ for the AdamW parameter groups and set the Hyperball weight learning rate to a fixed multiple, $\eta_t^{H}=m\,\eta_t^{\mathrm{base}}.$
For MuonH-wrapped matrices, $\eta_t^{H}$ is numerically equal to the prescribed effective LR. In the production run, we use $m=10$, while the remaining parameter groups follow $\eta_t^{\mathrm{base}}$ directly with AdamW.

\begin{figure}[ht]
\centering
\includegraphics[width=0.94\linewidth]{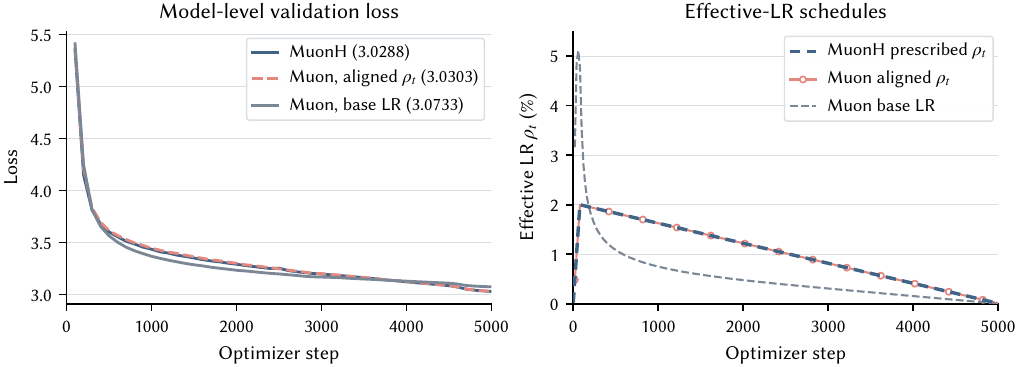}
\caption{170M BF16 comparison of MuonH, effective-LR aligned Muon, and ordinary-LR Muon. \textbf{Left:} validation loss. \textbf{Right:} effective LR traces for the selected MLP down-projection matrices. The training run with aligned-effective-LR and the normal Muon optimizer follows the predefined MuonH effective LR schedule. This run reaches a similar final loss, whereas ordinary-LR Muon produces a faster early decay of its induced effective LR and finishes with higher loss.}
\label{fig:hyperball-trace-alignment}
\end{figure}

\paragraph{Matching the effective LR schedule also aligns validation loss.} 
The MuonH optimizer can make training more efficient than the ordinary Muon optimizer, and we find that the efficiency gap can be attributed to the difference in their effective LR schedule.
We compare three 170M-parameter BF16 runs with the same architecture, batch size, sequence length, seed, and evaluation protocol. The MuonH run uses a predefined effective LR schedule $\rho_t^H$. The ordinary-LR run uses the base LR schedule of the MuonH run as the LR schedule and does not compensate for the evolving weight and update norms.
Then, as shown in \Cref{fig:hyperball-trace-alignment}, the ordinary-LR Muon run produces an effective LR trace that decays rapidly early in training and finishes close to zero.
The MuonH run instead follows the predefined linear-decay effective-LR schedule. Although the ordinary-LR Muon run achieves lower validation loss early in training, MuonH overtakes it near the end. The MuonH run finishes at a validation loss of $3.029$, compared with $3.073$ for ordinary-LR Muon.

For comparison, we conduct an effective-LR aligned run, which uses the ordinary Muon optimizer but adjusts its scalar learning rate online to follow the same effective LR curve,
\begin{equation}
\eta_t
=
\rho_t^H
\frac{\lVert W_t\rVert_F}
{\lVert u_t\rVert_F},
\end{equation}
so that the effective LR $\rho_t(W)=\rho_t^H$ without projecting $W_{t+1}$ back to a fixed-radius sphere. 
As shown in~\Cref{fig:hyperball-trace-alignment}, effective-LR aligned Muon follows the prescribed  linear post-warmup LR decay and reaches a final validation loss of $3.030$, close to MuonH at $3.029$. Thus, in this selected diagnostic, \emph{ordinary Muon largely recovers the MuonH loss curve when its effective LR schedule is aligned}. 
This comparison supports the effective-LR schedule as an important factor in MuonH's training behavior. It also suggests that better control of the effective-LR schedule can improve loss convergence.
Recent work also discusses related behavior from the perspective of angular update size~\citep{zhou-zhou-gu-2026-elr,xiao2026hyperball}. 
We provide an additional ordinary Muon control in~\Cref{app:hill-like} that can even use a counterintuitive hill-like scalar learning rate schedule to achieve loss convergence while maintaining the prescribed effective LR trajectory.

\paragraph{Effective LR provides a more informative descriptor of the loss trend.}
The experiments above show a close relationship between effective LR and loss dynamics. We further test whether effective LR provides a more informative descriptor than the raw optimizer learning rate. 
We apply the Multi-Power Law (MPL), a LR-schedule scaling law that predicts loss-curve shape from the LR schedule, to both the effective-LR schedule and the raw learning rate schedule~\citep{luo2025multipowerlawlosscurve}.
For the two ordinary-Muon runs above, we fit the post-warmup validation curves using either the scalar learning rate or the induced effective LR, while holding out the final $20\%$ of validation targets. As shown in \Cref{fig:effective-lr-mpl-fc2}, using the effective LR reduces the mean held-out RMSE across the two runs from $0.0265$ to $0.0210$. The improvement is driven mainly by the base-Muon run; for effective-LR aligned Muon, the two LR signals are comparably predictive. 
This small diagnostic therefore supports effective LR as a more informative descriptor for schedule-dependent training dynamics. The MPL formulation and fitting results are provided in~\Cref{app:muonh:mpl}.

\paragraph{Effective LR MPL helps interpret the late-stage crossover.}
In the selected 170M diagnostic, MuonH and effective-LR aligned Muon
converge more slowly at the beginning but outperform ordinary Muon near the
end of training. This pattern is consistent with the learning rate analysis
in~\Cref{app:lr-schedule-analysis}. In particular, the MPL
fit~\citep{luo2025multipowerlawlosscurve} provides a simple diagnostic relationship: \emph{a decrease in learning rate can induce an approximately proportional reduction in loss, with the loss decrease taking effect over subsequent training steps.} Applying this view to the effective LR, ordinary Muon decays its effective LR more aggressively early in training, which is consistent with its faster early loss reduction but leaves less effective LR decay for the later stage. MuonH and aligned Muon instead follow a more gradual, approximately linear effective LR decay, giving up some early convergence speed while retaining more decay toward the end of training. This view provides additional motivation for using a linear-decay schedule in our production design.

Together, these diagnostics motivate treating effective LR as the primary hyperparameter for MuonH-wrapped weights. They also suggest that the placement of its decay can affect where loss reduction occurs during training. 
We therefore next study the practical design of the LR schedule for production run.

\subsubsection{Learning Rate Schedule Design}

\paragraph{Continual training motivates an open-ended schedule followed by a controlled terminal decay.}
In practice, the total token budget is commonly undetermined in the beginning~\citep{hu2024minicpmunveilingpotentialsmall,shen2024power,smol2}. Hence, continual pretraining is a practical setting and may change the data mixture after the initial horizon. We therefore want to retain an open-ended schedule before applying a controlled terminal decay. Warmup-Stable-Decay (WSD) schedule provides a testbed for choosing the decay duration~\citep{hu2024minicpmunveilingpotentialsmall}. Motivated by the previous section's experiments, we use a linear decay function in WSD. All sweeps use a $0.6$B model with the Qwen3-0.6B architecture and global batch size 512.  At 20 tokens per parameter (TPP), or $11.3$B tokens, we vary the effective peak from $0.008$ to $0.024$ in increments of $0.004$ and the WSD decay ratio from $0.2$ to $1.0$ in increments of $0.2$.  At peaks $0.008$ and $0.012$, we additionally test TPP 50 and 100.

\begin{figure}[ht]
    \centering
    \includegraphics[width=\linewidth]{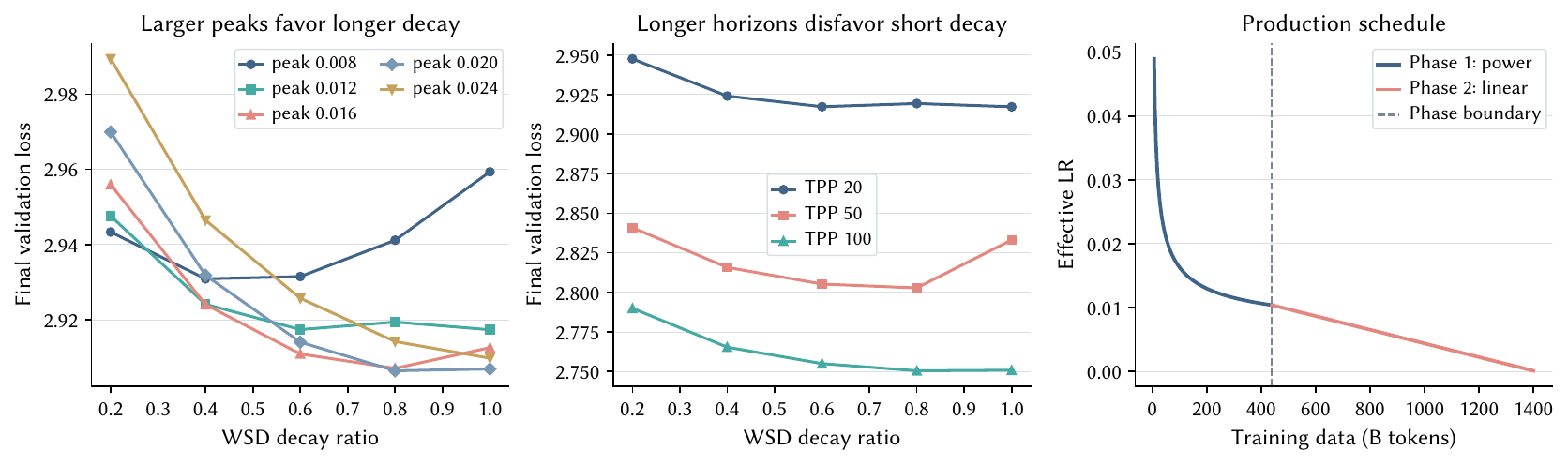}
    \caption{Experiments that support our observations.  \textbf{Left:} At TPP 20, larger effective peaks increasingly penalize short WSD decay.  \textbf{Center:} At fixed effective peak $0.012$, longer horizons make short decay less competitive.  Each sweep curve reports final validation loss from one run per configuration.  \textbf{Right:} The production effective LR schedule uses an open-ended power-decay phase followed by a long linear-decay phase.}
    \label{fig:lr-wsd-decay-ratio}
\end{figure}

\paragraph{Observation 1: a larger peak LR calls for a longer decay.}
In the left panel of~\Cref{fig:lr-wsd-decay-ratio}, the best grid points at
effective peaks $0.008$, $0.012$, $0.016$, $0.020$, and $0.024$ are decay ratios
$0.4$, $1.0$, $0.8$, $0.8$, and $1.0$.
These point optima fluctuate because the long-decay region is broad.  A more stable summary is the lower edge of the region within $0.01$ validation loss of the minimum: it moves nondecreasingly as
$0.4$, $0.4$, $0.6$, $0.6$, and $0.8$, as shown
in~\Cref{fig:app-wsd-limited-compute}.
At peaks $0.020$ and $0.024$, ratios $0.8$
and $1.0$ differ by only $0.0005$ and $0.0044$ loss.  Thus an almost fully linear
decay is already competitive near the high end of the tested peak range.  The
two-anchor interpolation in~\Cref{fig:app-wsd-two-anchor-example} provides a
complementary check: using only ratios $0.2$ and $1.0$ at peak $0.024$, it places
the estimated minimum at ratio $0.85$ within the same long-decay region.

\paragraph{Observation 2: a longer training horizon also calls for a longer decay.}
At effective peak $0.012$, the center panel of~\Cref{fig:lr-wsd-decay-ratio}
shows the competitive region shifting toward
longer decay.  Its lower edge moves from $0.4$ at TPP 20 to $0.6$ at TPP 50 and
100, while the point optima are $1.0$, $0.8$, and $0.8$.  At peak $0.008$, the
lower edge remains $0.4$, although the point optimum moves from $0.4$ to $0.6$.
The latter TPP100 grid lacks ratio $0.2$, so we treat it only as supporting
evidence.  The complete peak-$0.012$ lower-edge trace appears
in~\Cref{fig:app-wsd-limited-compute}; the conclusion is a broad shift toward
longer decay, not a precise monotone point optimum.

\paragraph{The production schedule combines these three practical features.}

The right panel of~\Cref{fig:lr-wsd-decay-ratio} shows the complete production effective-LR schedule. Here, we present the Hyperball weight LR, which is numerically the effective LR for wrapped matrices, instead of the base LR. Phase~1 uses the open-ended power schedule
in~\Cref{eq:production-phase1-lr}, decreasing from approximately
$5\times10^{-2}$ to $1.04\times10^{-2}$ over \PhaseOneWandBPhaseTokens.  Its
tail preserves a continuation path as the data distribution
evolves~\citep{shen2024power}.  Phase~2 joins near $10^{-2}$ and uses a long linear
decay over \PhaseTwoWandBPhaseTokens rather than spending most of training in a
stable segment followed by a short terminal drop.  The design therefore
combines continual trainability, a high initial effective LR, and a long
final decay.

\paragraph{Multi-power law provides a back-of-the-envelope, cross-schedule estimator under limited compute.}
Before a long run, exhaustively sweeping every decay ratio is impractical. Therefore, we fit a multi-power law (MPL) \cite{luo2025multipowerlawlosscurve} as a diagnostic, using only the two endpoint schedules. The runs are at TPP 20 and with decay ratios $0.2$ and $1.0$. Two curves per effective peak recover the increasing peak-to-decay trend: the estimated ratio rises from $0.33$ at peak $0.008$ to $0.85$ at peak $0.024$.  Although it is not a certificate of optimality, we use this result as a back-of-the-envelope trend diagnostic. The fitting protocol, unseen schedules, and errors are reported in~\Cref{app:lr-schedule-analysis}.

\subsection{Curriculum Model Averaging}
\label{sec:recipe:curriculum}

\begin{figure*}[t]
    \centering
    \resizebox{0.98\textwidth}{!}{\definecolor{cmablue}{HTML}{3F6488}
\definecolor{cmaorange}{HTML}{E4877F}
\definecolor{cmagreen}{HTML}{46AAA4}
\definecolor{cmagray}{HTML}{5C6670}

\begin{subfigure}[t]{0.49\textwidth}
\centering
\begin{tikzpicture}[
  x=1cm,
  y=1cm,
  font=\sffamily\footnotesize,
  label/.style={font=\sffamily\footnotesize, text=cmagray},
  note/.style={font=\sffamily\scriptsize, text=cmagray},
]

\draw[-{Latex[length=1.7mm]}, draw=cmablue, line width=0.65pt]
  (1.55,4.05) -- (7.45,4.05);
\node[label, anchor=west] at (1.55,4.27) {lower score};
\node[label, anchor=east] at (7.45,4.27) {higher score};

\node[label, font=\sffamily\bfseries\footnotesize, anchor=east]
  at (1.72,3.21) {Component A};
\node[note, anchor=east] at (1.72,2.96) {scored};
\foreach \xl/\xr/\tone in {2.05/2.50/18,2.50/2.95/30,
                              2.95/3.40/42,3.40/3.85/54,
                              5.25/5.70/66,5.70/6.15/78,
                              6.15/6.60/89,6.60/7.05/100} {
  \path[fill=cmablue!\tone, draw=white, line width=0.22pt]
    (\xl,2.95) rectangle (\xr,3.35);
}
\draw[cmablue!85, line width=0.45pt] (2.05,2.95) rectangle (3.85,3.35);
\draw[cmablue!85, line width=0.45pt] (5.25,2.95) rectangle (7.05,3.35);
\node[note] at (2.275,3.15) {$A_1$};
\node[note] at (2.725,3.15) {$A_2$};
\node[note] at (3.175,3.15) {$\cdots$};
\node[note, font=\sffamily\bfseries\scriptsize] at (3.625,3.15) {$A_k$};
\node[label] at (4.55,3.15) {$\cdots$};
\node[note, text=cmablue!90!black] at (5.65,3.56)
  {376 score-ranked slices};

\node[label, font=\sffamily\bfseries\footnotesize, anchor=east]
  at (1.72,2.05) {Component B};
\node[note, anchor=east] at (1.72,1.80) {unscored};
\foreach \xl/\xr in {2.05/2.50,2.50/2.95,2.95/3.40,3.40/3.85,
                       5.25/5.70,5.70/6.15,6.15/6.60,6.60/7.05} {
  \path[fill=cmaorange!68, draw=white, line width=0.22pt]
    (\xl,1.79) rectangle (\xr,2.19);
}
\draw[cmaorange!88, line width=0.45pt] (2.05,1.79) rectangle (3.85,2.19);
\draw[cmaorange!88, line width=0.45pt] (5.25,1.79) rectangle (7.05,2.19);
\node[note] at (2.275,1.99) {$B_1$};
\node[note] at (2.725,1.99) {$B_2$};
\node[note] at (3.175,1.99) {$\cdots$};
\node[note, font=\sffamily\bfseries\scriptsize] at (3.625,1.99) {$B_k$};
\node[label] at (4.55,1.99) {$\cdots$};
\node[note, text=cmaorange!90!black] at (5.65,2.44)
  {376 randomly ordered slices};

\node[label] at (2.95,1.48) {$\vdots$};
\node[label] at (4.55,1.48) {$\ddots$};
\node[label] at (6.15,1.48) {$\vdots$};
\node[label, font=\sffamily\bfseries\footnotesize, anchor=east]
  at (1.72,0.88) {Component N};
\node[note, anchor=east] at (1.72,0.63) {scored};
\foreach \xl/\xr/\tone in {2.05/2.50/12,2.50/2.95/24,
                              2.95/3.40/36,3.40/3.85/48,
                              5.25/5.70/60,5.70/6.15/70,
                              6.15/6.60/80,6.60/7.05/90} {
  \path[fill=cmagray!\tone, draw=white, line width=0.22pt]
    (\xl,0.62) rectangle (\xr,1.02);
}
\draw[cmagray!85, line width=0.45pt] (2.05,0.62) rectangle (3.85,1.02);
\draw[cmagray!85, line width=0.45pt] (5.25,0.62) rectangle (7.05,1.02);
\node[note] at (2.275,0.82) {$N_1$};
\node[note] at (2.725,0.82) {$N_2$};
\node[note] at (3.175,0.82) {$\cdots$};
\node[note, font=\sffamily\bfseries\scriptsize] at (3.625,0.82) {$N_k$};
\node[label] at (4.55,0.82) {$\cdots$};

\draw[cmagreen, rounded corners=0.14cm, line width=0.75pt]
  (3.335,0.56) rectangle (3.915,3.40);
\draw[-{Latex[length=1.5mm]}, draw=cmagreen, line width=0.65pt]
  (3.625,0.54) -- (3.625,0.19);
\node[label, text=cmagreen!85!black] at (3.05,0.03)
  {Phase 2 slice $k$};
\draw[-{Latex[length=1.5mm]}, draw=cmagreen, line width=0.65pt]
  (4.18,0.03) -- (5.00,0.03);

\begin{scope}
  \clip (5.30,-0.29) rectangle (6.24,0.35);
  \path[fill=cmagray!48, draw=white, line width=0.25pt]
    (5.30,-0.29) rectangle (6.24,-0.21);
  \path[fill=white, draw=white, line width=0.25pt]
    (5.30,-0.21) rectangle (6.24,-0.13);
  \path[fill=cmaorange!68, draw=white, line width=0.25pt]
    (5.30,-0.13) rectangle (6.24,-0.05);
  \path[fill=cmablue!54, draw=white, line width=0.25pt]
    (5.30,-0.05) rectangle (6.24,0.03);
  \path[fill=cmagray!48, draw=white, line width=0.25pt]
    (5.30,0.03) rectangle (6.24,0.11);
  \path[fill=white, draw=white, line width=0.25pt]
    (5.30,0.11) rectangle (6.24,0.19);
  \path[fill=cmaorange!68, draw=white, line width=0.25pt]
    (5.30,0.19) rectangle (6.24,0.27);
  \path[fill=cmablue!54, draw=white, line width=0.25pt]
    (5.30,0.27) rectangle (6.24,0.35);
\end{scope}
\draw[cmagreen, line width=0.65pt]
  (5.30,-0.29) rectangle (6.24,0.35);
\node[note] at (5.77,-0.17)
  {$\cdotp\!\cdotp\!\cdotp$};
\node[note] at (5.77,0.15)
  {$\cdotp\!\cdotp\!\cdotp$};

\end{tikzpicture}
\caption{Rank within each component and merged by progress}
\end{subfigure}\hfill%
\begin{subfigure}[t]{0.49\textwidth}
\centering
\begin{tikzpicture}[
  x=1cm,
  y=1cm,
  font=\sffamily\scriptsize,
  label/.style={font=\sffamily\scriptsize, text=cmagray},
  axis/.style={-{Latex[length=1.7mm]}, draw=cmagray, line width=0.55pt},
  checkpoint/.style={circle, draw=cmagray, fill=cmagreen!78,
    line width=0.65pt, minimum size=0.25cm, inner sep=0pt},
  legendcheckpoint/.style={circle, draw=cmagray, fill=cmagreen!78,
    line width=0.50pt, minimum size=0.18cm, inner sep=0pt},
]

\path (8.50,-0.18) (15.80,4.35);
\draw[axis] (8.90,0.02) -- (15.80,0.02);
\draw[axis] (8.90,0.02) -- (8.90,4.00);
\node[label, font=\sffamily\footnotesize, rotate=90]
  at (8.55,2.10) {Learning rate};
\node[label, font=\sffamily\footnotesize, anchor=north east]
  at (15.80,-0.12) {Iteration};

\draw[cmablue, line width=1.15pt] (9.38,2.98) -- (12.55,1.82);
\node[label, text=cmablue!90!black, anchor=south west]
  at (10.35,2.74) {linear decay};
\draw[cmablue, line width=1.0pt, dash pattern=on 2.4pt off 1.5pt]
  (12.55,1.82) -- (15.55,0.72);

\path[fill=cmagreen]
  (12.55,1.802) rectangle (15.45,1.838);
\node[label, text=cmagreen!85!black, anchor=west]
  at (12.78,2.47) {resume at 218k};
\node[label, text=cmagreen!85!black, anchor=west]
  at (12.78,2.21) {constant-LR continuation};

\foreach \xx in {12.90,13.41,13.92,14.43,14.94,15.45}
  \node[checkpoint] at (\xx,1.82) {};
\path[fill=white, draw=cmagray!75, line width=0.50pt]
  (9.20,0.30) rectangle (11.10,0.85);
\node[legendcheckpoint] at (9.44,0.575) {};
\node[label, anchor=west, inner sep=0pt] at (9.58,0.575) {checkpoints};
\foreach \xx in {12.90,13.41,13.92,14.43,14.94,15.45}
  \draw[cmagreen!78, line width=0.45pt] (\xx,1.67) -- (\xx,1.22);
\draw[cmagreen!78, line width=0.65pt] (12.90,1.22) -- (15.45,1.22);
\draw[-{Latex[length=1.6mm]}, draw=cmagreen, line width=0.7pt]
  (14.18,1.22) -- (14.18,0.86);
\node[draw=cmagreen, rounded corners=0.12cm, fill=white,
  line width=0.65pt, inner xsep=5pt, inner ysep=3pt,
  text=cmagreen!85!black] at (14.18,0.55)
  {equal-weight average $\bar{w}$};

\end{tikzpicture}
\caption{Constant-LR continuation \& Model Average}
\end{subfigure}}
    \caption{\textbf{CMA-inspired data and optimization flow in the production
    Phase~2 recipe.} Component-local ordering, transition, late constant-LR
    continuation, and six-checkpoint averaging are shown together. Exact bucket
    construction, transition, and averaging details are given in \Cref{app:curriculum:buckets,app:curriculum:transition,app:curriculum:averaging}.
    }

    \label{fig:curriculum-model-average}
\end{figure*}
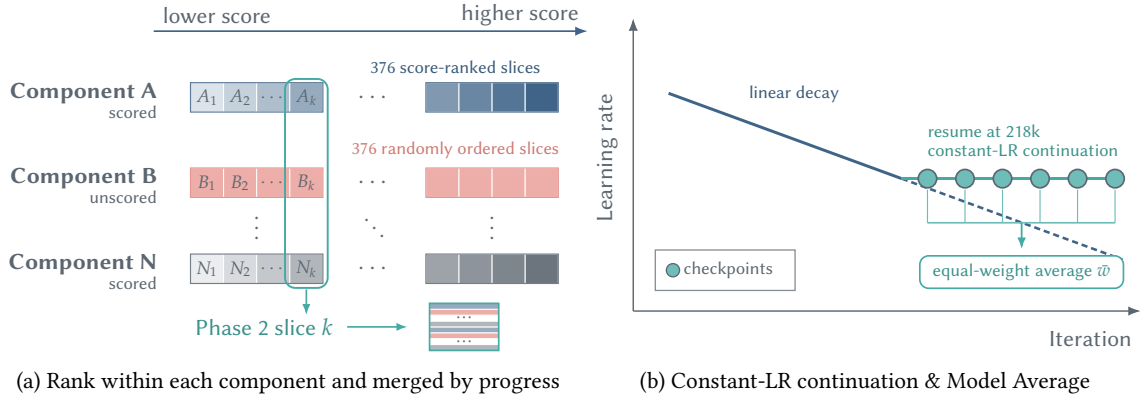

When open-source datasets provide usable sample-level scores, we rank samples within each source and construct a mixture-preserving curriculum that places more preferred samples later in training. This ordering is designed to improve the data efficiency of higher-quality samples. However, a conventional terminal learning-rate decay can work against this objective: \emph{examples presented near the end of the curriculum are also processed when parameter updates have become small}~\citep{luo2025learningratedecaywastes}. Curriculum Model Averaging (CMA)~\citep{luo2025learningratedecaywastes} addresses this conflict by combining an ascending data curriculum with constant-LR training and averaging late checkpoints. Our production recipe adopts this principle in Phase~2. We first follow the scheduled learning-rate trajectory, then resume from a selected late checkpoint, hold the base learning rate fixed, and average six checkpoints from the continuation. The complete data and optimization flow is summarized in~\Cref{fig:curriculum-model-average}, with implementation details provided in~\Cref{app:curriculum:buckets,app:curriculum:averaging}. 
We next describe the curriculum construction, the selected constant-LR continuation with checkpoint averaging, and the corresponding ablation results.

For convenience, we label the Phase~2 variants: \textbf{UD} (\textbf{U}niform data with LR \textbf{D}ecay) globally reshuffles the Phase~2 pool and follows the scheduled learning-rate decay. \textbf{CD} (\textbf{C}urriculum data with LR \textbf{D}ecay) keeps that decay but presents the data pool in the component-local curriculum order. \textbf{CDC} (\textbf{C}urriculum data, with LR \textbf{D}ecay then \textbf{C}onstant LR) uses the curriculum order, follows the decay, then continues at a constant learning rate, and reports the final checkpoint. \textbf{CMA} (\textbf{C}urriculum \textbf{M}odel \textbf{A}veraging) is the production variant: it averages the last several checkpoints from the CDC recipe.

\paragraph{The data curriculum arranges the order within each component while keeping the target mixture weight.} We first establish a reproducible order separately for every retained
component. If a component provides a usable sample-level quality signal, we
sort its examples according to the source-specific interpretation of its score,
placing less preferred score ranges earlier and more preferred ranges later.
Components without a usable score instead follow a fixed random order. We then assign each example a normalized within-component rank according to
its position in this ordered sequence, measured by cumulative token mass.
Intuitively, this rank indicates which source-local score range an example
falls into. For example, a rank of \(0.25\) means that approximately one
quarter of the component's tokens appear earlier in its ordered sequence.

We divide the normalized rank range into 376 intervals, with each curriculum bucket formed by taking the corresponding interval from every component. Each bucket contains approximately \(2.5\)B tokens, drawing roughly \(1/376\) of tokens from every component and preserving the selected cross-component mixture.
As training advances through the buckets, scored components move toward their
more preferred score ranges, while unscored components advance through their
fixed random orders. 
Since scores are interpreted only within their own sources and are not compared
across components, this construction is not a global quality ranking. The
scalable approximation used to construct the 376 buckets and the exact
transition schedule are given
in~\Cref{app:curriculum:buckets,app:curriculum:transition}
and~\Cref{tab:transition-token-ledger}.

\paragraph{The canonical CMA run uses the best-performing late-stage continuation with model averaging.}
We compare three late-stage training strategies: averaging checkpoints directly from the terminal-decay trajectory, and two constant-LR continuations resumed from steps 215{,}000 and 218{,}000. Among these configurations, the continuation from step 218{,}000 gives the strongest observed performance and is therefore selected for the production model. For this continuation, training resumes from step 218{,}000 with the base learning rate fixed at $4.08\times10^{-5}$. With the production Hyperball multiplier $m=10$, the Hyperball weight LR is therefore $4.08\times10^{-4}$, which is also the effective LR for the MuonH-wrapped matrices. 
During the late stage of this continuation, we save checkpoints at approximately regular intervals and form the released model by equally averaging the parameters of six such checkpoints. Neighboring checkpoints are typically separated by 100 optimizer steps, or approximately $0.63$B tokens under the production batch configuration. Full resume, checkpoint-spacing, and averaging details are provided in~\Cref{app:curriculum:averaging}. 

\paragraph{Ablation study on curriculum, model average, and const-LR continuation.} The endpoint comparisons in~\Cref{fig:curriculum-endpoint-scores} support a structured
comparison of the jointly selected recipe ingredients, as illustrated in \Cref{app:puro-scaling-ledger,fig:curriculum-endpoint-scores}.  First, \emph{curriculum
ordering improves over uniform ordering} by $1.18$ points without model
averaging (CD: 57.17 versus UD: 55.99) and by $1.61$ points when comparing the
model averages of the corresponding decay trajectories (CD model average: 57.18 versus UD model average: 55.57).
Second, \emph{model averaging is not independently beneficial in both ordering
conditions}: it changes the UD endpoint by $-0.42$ points and the CD endpoint by
$+0.01$ points without constant-LR continuation.  Finally, the two CDC controls score 55.64 and 57.12 for the 215k and
218k continuations, respectively, when evaluated at their unaveraged final
checkpoints. With six-checkpoint averaging, the corresponding branches reach
56.80 and 57.81. We therefore select the 218k averaged branch as the
production CMA endpoint. It combines curriculum ordering, late constant-LR
continuation and six-checkpoint averaging.
Part of the observed advantage of CD over UD ordering could be attributed to benchmark
contamination in the later-stage data, as discussed in~\Cref{app:limitations}. But the remaining advantage is observed after post-training in~\Cref{sec:posttrain-ablation}, and the contamination alone is unlikely to explain the persistent improvement. We report the exact base LR and Hyperball effective LR, checkpoint IDs, and the model-export procedure in~\Cref{app:curriculum:averaging}.

\subsection{Post-Training Recipe}\label{sec:recipe:posttrain}

\paragraph{Motivation and question.}

In the above, we present a complete, low-cost pretraining recipe. The resulting pretrained checkpoints are later used for post-training. Here we focus on supervised fine-tuning (SFT). This raises a natural but underexplored question: \emph{does the choice of pretraining recipe affect model performance after SFT?} 
A released checkpoint alone is usually not enough to answer this question. Even when a checkpoint is publicly available, its full training recipe and data may be unavailable. This makes it difficult and expensive to rerun the training and ablate specific pretraining choices. Small-scale ablations are easier, but the undertrained base models may be too weak to show a stable post-training performance~\citep{qi2025evollm}. Therefore, our low-cost, fully open recipe allows us to study this question at the scale of the final model.

In particular, we compare the SFT results of checkpoints from two Phase~2 recipes: uniform data ordering with learning rate decay (UD) and Curriculum Model Averaging (CMA). The UD checkpoint performs worse on the base-model benchmark suite than the CMA checkpoint. Although both recipes use the same model architecture and Phase~2 training corpus, they differ in data ordering and checkpoint construction. This comparison therefore requires two separate Phase~2 training runs. In the CMA recipe, the curriculum places more high-quality data near the end of training. This changes the data distribution over the course of Phase~2. The base-model improvement may therefore be temporary and tied mainly to the final stage of pretraining. We test whether this improvement remains after both checkpoints undergo the same SFT procedure.

\paragraph{Data construction and experimental setup.}

To study this question, we apply the same SFT procedure to the UD and CMA checkpoints for each data setup. All other training conditions are matched within the same setup. We continue to use the MuonH optimizer, as in pretraining~\citep{wen2026fantastic2}. The base learning rate follows a cosine schedule that decays from $1\times 10^{-5}$ to $1\times 10^{-7}$. The MuonH-managed matrix parameters use a $10\times$ multiplier to obtain their Hyperball weight learning rate. We set the global batch size to 160.

To examine the results under different SFT data distributions, we construct three data setups. (1) \emph{GSM8K-based SFT}. It combines the original GSM8K training split with GSM8K-related synthetic data from MetaMathQA and OpenMathInstruct-2. It also includes selected Tulu-3 components, which cover mathematics and instruction following.
(2) \emph{Math\&Code SFT with replay}. It contains a larger collection of mathematics data from GSM8K, MetaMathQA, OpenMathInstruct-1, and OpenMathInstruct-2. It also includes code data from OpenCodeInstruct and Magicoder. We add a small, shuffled replay set from the final part of the Phase~2 curriculum, which contains the highest-quality Phase~2 data. (3) \emph{Tulu-3 mixed-domain SFT}. It uses the English portion of \texttt{Tulu-3-sft-mixture-0225}~\citep{lambert2024tulu3}. Because this mixture covers several domains, it provides a broader test of general capability and instruction following.

For all three setups, we deduplicate the training examples and discard empty sequences. For decontamination, we remove any training example that shares a 13-gram with an evaluation test example. Detailed data composition and training settings are provided in~\Cref{app:posttrain-details}.

\paragraph{Evaluation.}

We use the same evaluation protocol for the UD and CMA models within each setup. For GSM8K-based SFT and Math\&Code SFT with replay, we evaluate the resulting models on GSM8K with the \texttt{lm-evaluation-harness} framework. Because generated answers can differ in format, we use flexible extraction to identify the final answer. We report the average result over three repeated seeds. For Tulu-3 mixed-domain SFT, we use OpenCompass to evaluate the resulting models on 15 tasks for base models and IFEval. The evaluation protocol of the 15 tasks is adapted to the fine-tuned models. This suite covers knowledge, reasoning, code, and instruction following. We report both the overall average and the individual task results, since the effect may differ across tasks. Further evaluation details are provided in~\Cref{app:posttrain-details}.

\subsection{Data Recipe}\label{sec:data-recipe}

\begin{figure}[ht]
\centering
\begin{minipage}[c]{0.48\textwidth}
\refstepcounter{table}
\label{tab:data-recipe-summary}
\centering
\small
\begin{adjustbox}{max width=\linewidth}
\begin{tabular}{lrrrr}
\toprule
Domain & \multicolumn{2}{c}{Phase 1} & \multicolumn{2}{c}{Phase 2} \\
\cmidrule(lr){2-3}\cmidrule(lr){4-5}
 & Tokens & Share & Tokens & Share \\
\midrule
English & 321.31B & 73.2\% & 557.52B & 59.4\% \\
Math & 31.56B & 7.2\% & 171.29B & 18.3\% \\
Chinese & 51.34B & 11.7\% & 88.48B & 9.4\% \\
Code & 34.62B & 7.9\% & 108.11B & 11.5\% \\
SFT / instruction & -- & 0.0\% & 12.66B & 1.3\% \\
\midrule
Total & 438.84B & 100.0\% & 938.06B & 100.0\% \\
\bottomrule
\end{tabular}
\end{adjustbox}

\end{minipage}\hfill
\begin{minipage}[c]{0.50\textwidth}
\centering
\includegraphics[width=\linewidth]{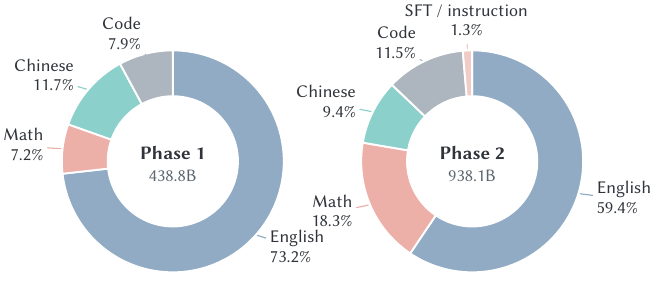}
\end{minipage}
\caption{Domain composition of the two pretraining phases. The left table reports materialized token counts and shares for Phase~1 and Phase~2, and the right panel visualizes the same composition. Phase~1 emphasizes broad coverage, while Phase~2 allocates a larger share to mathematics and introduces instruction-like data. The Phase~2 token budget excludes Phase~1 replay. See \Cref{sec:recipe:data:puro-spark} for details.}
\label{fig:data-domain-composition}
\end{figure}

\begin{figure}[ht]
\centering
\includegraphics[width=\linewidth]{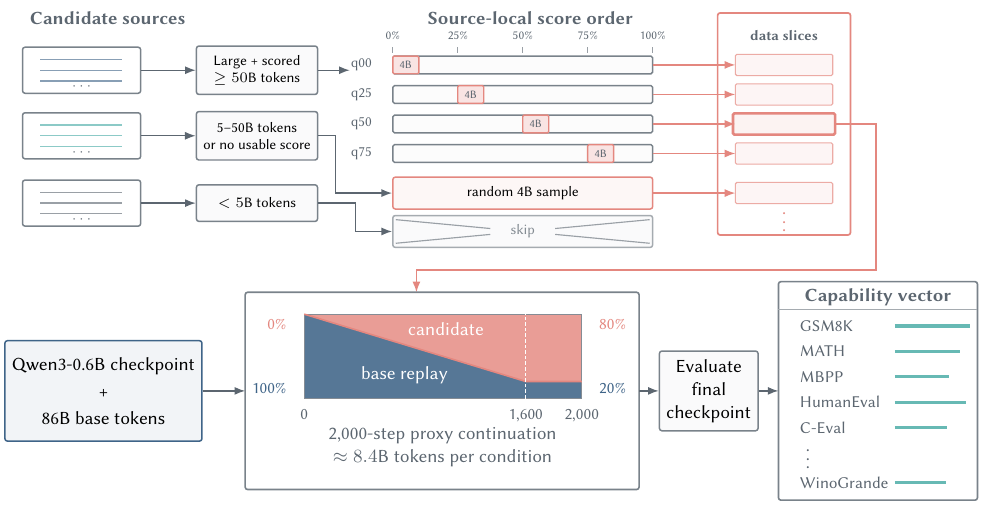}
\caption{
Proxy benchmarking protocol. Candidate slices are first constructed according to corpus size and the availability of sample-level quality scores. Large scored sources contribute four within-source quantile slices, while other eligible sources contribute one random 4B-token slice. Each retained slice is used for a 2{,}000-step continuation run from the same Qwen3-0.6B checkpoint. During continuation training, the candidate-data ratio gradually rises to 80\%, while replay from the base mixture decreases accordingly. The final checkpoint is evaluated on the same 15-benchmark suite to obtain a capability vector. These vectors guide the fine-grained slice selection, source allocation, and capability balancing described below.
}
\label{fig:proxy-benchmark-protocol}
\end{figure}

Our goal is to select a compact set of high-quality and diverse data sources while preserving broad domain coverage. Candidate datasets differ in both quality and capability profile, so source-level metadata alone is not sufficient for data selection. We therefore use \emph{proxy benchmarking} to compare candidate data slices under a shared protocol. As illustrated in~\Cref{fig:proxy-benchmark-protocol}, each run starts from the same Qwen3-0.6B checkpoint and follows the same continuation-training schedule. After continuation training, we evaluate the resulting checkpoint on a fixed 15-benchmark suite. The score vector from this evaluation serves as the capability profile of the candidate slice. These score vectors can guide our data recipe in heuristic. The following three paragraphs describe how we construct fine-grained slices, select and allocate data sources, and balance different capabilities in the final mixture. Detailed settings are provided in~\Cref{app:data-recipe:proxy-protocol}.

\subsubsection{How Do We Select a Good Data Recipe?}
\label{sec:recipe:data:selection}\label{sec:recipe:data:insight}

\paragraph{Fine-grained data slice selection.}
We construct candidate data slices according to source size and the availability of sample-level quality scores. We apply three sampling rules. First, if a dataset contains more than 50B tokens and provides usable score labels, we sort its examples by score and sample four local slices around the 0th, 25th, 50th, and 75th percentile positions. Second, for datasets between 5B and 50B tokens, as well as larger datasets without usable score labels, we randomly sample one 4B-token slice. Third, we do not evaluate the remaining smaller datasets. For each large scored source, the four slices form a within-source quantile benchmark. Because these slices come from different positions in the same score ordering, they reveal quality differences that a single source-level result would hide. For example, the \texttt{q00} slice of DCLM-Dedup ranks third in average proxy performance, while its \texttt{q25} slice ranks eleventh. This result shows that different regions of the same source can provide substantially different training value.

\paragraph{Feature-guided data selection.}\label{sec:recipe:data:proxy-feature}
We use the measured capability profiles to guide both source selection and data allocation. MegaMath-Web-Pro performs strongly on Math and General, while MegaMath-Code provides the strongest Code signal among the measured candidates. We therefore assign both sources larger shares in Phase~2 to strengthen general and target capabilities. The quantile results also guide within-source filtering. Because the top-scoring region of DCLM-Dedup performs much better than its lower-ranked regions, we retain 34B tokens from this region in Phase~2. This selection corresponds to approximately the top 5\% of DCLM-Dedup. FineWeb-Edu-CN ranks highest on Chinese capability, so we also increase its share to preserve Chinese performance. The proxy results provide useful evidence for increasing, reducing, or filtering different data sources. Here, we take heuristics from these capability profiles and still choose the concrete mixing ratios manually.

\paragraph{Capability balance across domains.}
The proxy results show that different datasets have substantially different capability profiles. A source that performs well on one group of benchmarks may perform less well on others. For example, FineWeb-Edu-CN ranks first on Chinese capability but is weaker on Math, Code, and General~(\Cref{app:data-recipe}). We therefore need to balance broad general knowledge with more specialized reasoning capabilities rather than maximize a single benchmark dimension. To make these trade-offs more explicit, we apply PCA to the capability vectors. As shown in \Cref{fig:proxy-feature-pca}, General aligns mainly with PC1, while Code points in the opposite direction. Math aligns mainly with PC2 and remains slightly positive on PC1. In this projection, Code shows a stronger trade-off with General than Math does~(\Cref{app:data-recipe:proxy-pca}). This observation informs the roles of the two pretraining phases. Phase~1 preserves broad coverage, with English accounting for \PhaseOneEnglishShare of \PhaseOneTokens~(\Cref{tab:data-recipe-summary,fig:data-domain-composition}). Phase~2 then adds more specialist data, but allocates a slightly larger share to mathematics than to code. In this way, the two-phase recipe strengthens the target reasoning capabilities while retaining broad general coverage.

\subsubsection{How Do We Preprocess Data and Reproduce the Shards?}
\label{sec:recipe:data:puro-spark}

\paragraph{Data preprocessing pipeline.}
The pipeline turns the candidate collection into fixed training shards in four stages. \emph{Within-source deduplication} removes repeated documents from web components such as DCLM-Dedup and FineWeb-Edu-EN. We do not deduplicate across sources because our analysis found little additional reduction in data volume. \emph{Proxy benchmarking} evaluates each eligible candidate source or within-source slice under the shared protocol described above and produces a multidimensional capability profile. \emph{Heuristic filtering} combines these profiles with domain coverage, language balance, and token-budget constraints to determine the retained portion and mixture weight of each source. \emph{Final materialization} freezes the source revisions, retention rules, mixture weights, and random seeds. It then merges the retained components into fixed shards that are read sequentially during training, without online mixing or sampling. In addition, we also use PreSelect fastText~\citep{shum2025predictivedataselectiondata} to evaluate the Nemotron-HQ~\citep{mahabadi2025nemotron} and Nemotron-HQ-Synthetic datasets. Apart from this step, we do not perform additional compute-intensive data scoring and primarily rely on existing quality scores.

\paragraph{\dataframework preprocessing framework.}
We implement the recipe with \dataframework, a Spark-based data preprocessing framework~\citep{luo2025pcmind}. The framework is designed for efficient large-scale data processing. For compute-intensive operations such as MinHash deduplication, it uses native C++ kernels through the Chukonu integration. This implementation is inherited from Chukonu, whose original Spark benchmark reports an approximately $2.5\times$ speedup over the corresponding JVM implementation~\citep{chukonu}. The framework also supports the curriculum bucket materialization required by the Phase~2 design. Given a fixed component pool and mixture, it assigns normalized progress within each component, aligns the resulting slices into token-sized buckets, and materializes their order using recorded random seeds. The same component pool can therefore produce either an ordered curriculum stream or a uniformly ordered stream by changing only the materialization order. These two data streams support the UD, CD, and CMA recipes in Phase~2~(\Cref{sec:recipe:curriculum}).

\paragraph{Two-phase data and transition schedule.}
All training data are materialized as static shards and manifests before training. Production runs therefore do not rely on online scoring or dynamic mixture sampling. The Phase~1-to-Phase~2 transition spans approximately $43.9$B training tokens. During this transition, the sampling weight of the Phase~1 replay mixture decreases \emph{linearly}, while that of the Phase~2 mixture increases \emph{linearly}. The transition therefore consumes approximately $21.9$B tokens from the Phase~1 replay data and $21.9$B tokens from the beginning of the Phase~2 data pool. The full Phase~2 component pool contains approximately $938.1$B tokens, including the $21.9$B tokens consumed during the transition. After the transition, the remaining $916.1$B Phase~2 tokens are consumed without Phase~1 replay. Counting the transition, the Phase~2 training stage consumes approximately $960.0$B tokens in total: $21.9$B tokens of Phase~1 replay and $938.1$B tokens from the Phase~2 pool. Together with Phase~1 training, the complete pretraining run consumes approximately $1.4$T tokens.

\section{Evaluation}\label{sec:evaluation}
\subsection{Evaluation Setup}

\subsubsection{Model Selection}

We compare \sys with recent language models of comparable size, primarily in
the 1B--3B parameter range.
We group these baselines according to the scope of their public releases.
\textbf{Open-weight models} make their weights publicly available but do not
release the complete training data or recipe, whereas \textbf{open-recipe
models} additionally release the data and training artifacts needed to study
or reproduce training.

\noindent\textbf{Open-weight models.}
\begin{itemize}
\setlength{\itemsep}{2pt}
\setlength{\parskip}{0pt}
\setlength{\parsep}{0pt}
    \item \textbf{Qwen series}~\citep{qwen2,qwen2025qwen25technicalreport,
    qwen3,qwen35}:
    We include Qwen2-1.5B, Qwen2.5-1.5B, Qwen3-1.7B-Base, and
    Qwen3.5-2B-Base, which provide successive generations of strong compact
    foundation models in the 1.5B--2B range.

    \item \textbf{Gemma series}~\citep{morgane2024gemma2,gemma3,gemma4}:
    We select Gemma-2-2B, Gemma-3-1B-PT, and Gemma-4-E2B-Base to cover three
    generations of Google's lightweight open models.

    \item \textbf{Llama 3.2}~\citep{llama3.2}: We include Llama-3.2-3B, the latest
    pretrained Llama checkpoint available in the compact 1B--3B range.

    \item \textbf{LFM2.5}~\citep{liquidai2025lfm2}: We include
    LFM2.5-1.2B-Base as a recent compact base model designed for efficient
    on-device deployment.

    \item \textbf{Falcon-H1}~\citep{zuo2025falconh1}: We evaluate the standard
    and deeper 1.5B base variants as efficiency-oriented hybrid-model baselines.
\end{itemize}

\noindent\textbf{Open-recipe models.}
\begin{itemize}
\setlength{\itemsep}{2pt}
\setlength{\parskip}{0pt}
\setlength{\parsep}{0pt}
    \item \textbf{Instella}~\citep{liu2025instella}:
    Instella-3B was trained entirely on openly available data, with model
    weights, training data, configurations, and code released to support
    reproducibility. Despite using fewer pretraining tokens than many
    contemporary models, it is reported to remain competitive with leading
    open-weight models at a similar scale.

    \item \textbf{OLMoE}~\citep{muennighoff2025olmoe}:
    OLMoE-1B-7B-0125 is a sparse model with 1B active and 7B total parameters.
    Its release includes intermediate checkpoints, training data, code, logs,
    and evaluation resources. The model is positioned as a strong
    performance--efficiency baseline among fully-open mixture-of-experts models.

    \item \textbf{Yulan-Mini-2.4B}~\citep{yiwen-etal-2025-yulan}:
    Yulan-Mini-2.4B is a 2.4B model trained on 1.1T tokens, with its training recipe
    and phase-wise data composition publicly documented. It emphasizes
    data-efficient pretraining and reports competitive performance against
    similarly sized models trained with substantially more data.

    \item \textbf{SmolLM3}~\citep{smollm3}:
    SmolLM3-3B-Base was pretrained on approximately 11T tokens, with the public
    data mixture and training configurations released alongside the model. It
    is presented as a strong fully-open model at the 3B scale, with performance
    competitive with several larger compact models.

    \item \textbf{MiniCPM5}~\citep{openbmb2026minicpm5}:
    MiniCPM5-1B-Base is a compact dense model developed for local and
    resource-constrained deployment. Its associated web and mathematical
    training corpora are publicly released, while the MiniCPM5 family targets
    strong reasoning, coding, and mathematical capabilities within a small
    deployment footprint.
\end{itemize}

For a fair comparison, we consistently evaluate the pretrained/base versions
of all models.

\subsubsection{Benchmarks}

Our evaluation covers 15 benchmarks, organized into two groups: mathematics
and code, and reasoning and knowledge. For mathematical reasoning, we use
GSM8K~\citep{gsm8k} and the original MATH benchmark~\citep{MATH}, which measure
grade-school arithmetic and competition-level mathematics, respectively. For
code generation, we use sanitized-MBPP~\citep{mbpp} and
HumanEval~\citep{humaneval}, where generated Python programs are verified by
unit tests. For reasoning and knowledge, we use MMLU~\citep{mmlu},
MMLU-Pro~\citep{NEURIPS2024_ad236edc}, ARC-Challenge and ARC-Easy~\citep{arc},
BoolQ~\citep{clark-etal-2019-boolq}, CommonsenseQA
(CSQA)~\citep{commonsenseqa}, HellaSwag~\citep{hellaswag}, PIQA~\citep{piqa},
SocialIQA (SIQA)~\citep{socialiqa}, WinoGrande~\citep{winogrande}, and
Big-Bench Hard (BBH)~\citep{suzgun-etal-2023-challenging}. Together, these
benchmarks assess mathematical reasoning, code generation, academic knowledge,
reading comprehension, commonsense understanding, and general multi-step
reasoning.

\subsubsection{Implementation Details}

We evaluate every model with OpenCompass~\citep{2023opencompass} using a
fixed checkpoint and tokenizer revision, dataset snapshot, benchmark-specific
prompt, shot examples, generation configuration, and answer postprocessor.
GSM8K, MATH, sanitized-MBPP, HumanEval, MMLU-Pro, and BBH use
generation-based (GEN) evaluation: the model produces a free-form response from
which the final answer or executable code is parsed. GEN decoding is greedy
with fixed maximum output and stopping rules. The
remaining nine benchmarks use perplexity-based (PPL)
evaluation, which ranks the provided candidate answers by their fixed token
log-likelihoods and therefore introduces no generation randomness.
For GSM8K, a shared deterministic postprocessor prioritizes explicit final-answer
markers and boxed answers, preserves signs and fractions, and uses the final
numeric expression only when no explicit answer is present.

Following the OLMES~\citep{gu2025olmes} convention, for tasks supporting both cloze formulation (CF) and multiple-choice formulation (MCF), we evaluate both formulations for each model and use the better-performing formulation when computing the reported aggregate. This selection is performed independently for each model. All scores are percentages. In each table, Avg is the unweighted arithmetic mean
over the benchmarks displayed in that table. Throughout the cost--performance
and scaling analyses, overall performance $P$ denotes the unweighted arithmetic
mean over all 15 benchmark scores reported
in~\Cref{tab:math_code_capabilities,tab:reasoning_knowledge}. Every score in our
tables is produced by this common, deterministic pipeline.

\subsection{Resulting Performance}

\begin{table}[ht]
\centering
\caption{Core capabilities in mathematics and code generation. All scores are
produced by the same evaluation pipeline and reported as percentages. Avg is
the unweighted arithmetic mean over the four benchmarks.}
\label{tab:math_code_capabilities}
\setlength{\tabcolsep}{7.0pt}
\renewcommand{\arraystretch}{1.12}
\resizebox{0.9\textwidth}{!}{%
\begin{tabular}{l c cc cc c}
\toprule
\textbf{Model Name} & \textbf{Params}
& \multicolumn{2}{c}{\textbf{Math}}
& \multicolumn{2}{c}{\textbf{Code}}
& \textbf{Avg} \\
\cmidrule(lr){3-4}\cmidrule(lr){5-6}
& & \shortstack{GSM8K\\4-shot}
& \shortstack{MATH\\4-shot}
& \shortstack{sanitized-MBPP\\3-shot}
& \shortstack{HumanEval\\3-shot}
& \\
\midrule

\multicolumn{7}{l}{\textit{\textbf{Open-Weight Models}}} \\
\rowcolor{openweight} Qwen2-1.5B & 1.5B & 59.82 & 23.70 & 50.19 & 27.44 & 40.29 \\
\rowcolor{openweight} Qwen2.5-1.5B & 1.5B & 67.70 & 32.28 & 58.37 & 31.71 & 47.52 \\
\rowcolor{openweight} Qwen3-1.7B-Base & 1.7B & 76.04 & 42.44 & 64.20 & 51.22 & 58.48 \\
\rowcolor{openweight} Qwen3.5-2B-Base & 2B & 69.83 & 35.82 & 43.58 & 32.32 & 45.39 \\
\rowcolor{openweight} Gemma-2-2B & 2B & 33.81 & 14.84 & 38.52 & 20.12 & 26.82 \\
\rowcolor{openweight} Gemma-3-1B-PT & 1B & 2.58 & 0.94 & 9.34 & 8.54 & 5.35 \\
\rowcolor{openweight} Gemma-4-E2B-Base & E2B & 30.71 & 9.68 & 47.86 & 24.39 & 28.16 \\
\rowcolor{openweight} Llama-3.2-3B & 3B & 28.20 & 8.14 & 50.19 & 31.10 & 29.41 \\
\rowcolor{openweight} LFM2.5-1.2B-Base & 1.2B & 53.22 & 28.00 & 38.13 & 25.61 & 36.24 \\
\rowcolor{openweight} Falcon-H1-1.5B-Base & 1.5B & 74.68 & 42.60 & 61.09 & 28.05 & 51.61 \\
\rowcolor{openweight} Falcon-H1-Deep-1.5B-Base & 1.5B & 53.53 & 41.28 & 63.04 & 32.93 & 47.70 \\

\midrule
\multicolumn{7}{l}{\textit{\textbf{Open-Recipe Models}}} \\
\rowcolor{fullyopen} Instella-3B & 3B & 64.44 & 13.30 & 41.25 & 14.63 & 33.41 \\
\rowcolor{fullyopen} OLMoE-1B-7B-0125 & A1B/7B & 54.59 & 15.48 & 33.46 & 12.80 & 29.08 \\
\rowcolor{fullyopen} Yulan-Mini-2.4B & 2.4B & 68.99 & 27.18 & 62.26 & 50.00 & 52.11 \\
\rowcolor{fullyopen} SmolLM3-3B-Base & 3B & 81.73 & 56.12 & 60.31 & 37.80 & 58.99 \\
\rowcolor{fullyopen} MiniCPM5-1B-Base & 1B & 39.88 & 23.34 & 46.30 & 29.27 & 34.70 \\
\rowcolor{fullyopen} MobileLLM-R1-950M-base & 950M & 68.31 & 27.24 & 50.58 & 43.29 & 47.36 \\

\midrule
\multicolumn{7}{l}{\textit{\textbf{Ours}}} \\
\rowcolor{ours} Puro-2B (\$4.4K) & 2B & 56.03 & 27.86 & 49.81 & 20.73 & 38.61 \\
\rowcolor{ours} Puro-2B & 2B & 59.67 & 30.30 & 52.92 & 31.10 & 43.50 \\

\bottomrule
\end{tabular}%
}
\end{table}

\paragraph{Mathematics and code.}
As shown in~\Cref{tab:math_code_capabilities}, \sys obtains an average score of
43.50 across the four generative benchmarks. It exceeds Qwen2-1.5B by 3.21
points and is within 4.02 points of Qwen2.5-1.5B. Among the open-recipe
baselines, \sys outperforms Instella-3B, OLMoE-A1B/7B, and MiniCPM5-1B-Base,
while Yulan-Mini-2.4B, SmolLM3-3B-Base, and the newly included
MobileLLM-R1-950M-base achieve higher averages. MobileLLM-R1 is a strong
math/code-oriented compact baseline in this panel, exceeding \sys by 3.86
points on the four-task average. Given these differences in model size and
capability emphasis, \sys remains a competitive open-recipe model for
mathematics and code generation at a comparatively small scale.

\begin{table}[ht]
\centering
\caption{Reasoning and knowledge capabilities. All scores are produced by the
same evaluation pipeline and reported as percentages. Avg is the unweighted
arithmetic mean over all eleven benchmarks.}
\label{tab:reasoning_knowledge}
\setlength{\tabcolsep}{3.2pt}
\renewcommand{\arraystretch}{1.10}
\resizebox{\textwidth}{!}{%
\begin{tabular}{l c *{11}{c} c}
\toprule
\textbf{Model Name} & \textbf{Params}
& \multicolumn{11}{c}{\textbf{Reasoning \& Knowledge}}
& \textbf{Avg} \\
\cmidrule(lr){3-13}
&
& \shortstack{MMLU\\5-shot}
& \shortstack{MMLU-Pro\\5-shot CoT}
& \shortstack{ARC-C\\5-shot}
& \shortstack{ARC-E\\5-shot}
& \shortstack{BoolQ\\5-shot}
& \shortstack{CSQA\\5-shot}
& \shortstack{HSwag\\5-shot}
& \shortstack{PIQA\\5-shot}
& \shortstack{SIQA\\5-shot}
& \shortstack{Wino\\5-shot}
& \shortstack{BBH\\4-shot}
& \\
\midrule

\multicolumn{14}{l}{\textit{\textbf{Open-Weight Models}}} \\
\rowcolor{openweight} Qwen2-1.5B & 1.5B & 56.39 & 23.62 & 69.83 & 83.77 & 71.90 & 70.52 & 59.91 & 75.19 & 63.36 & 59.67 & 31.73 & 60.54 \\
\rowcolor{openweight} Qwen2.5-1.5B & 1.5B & 61.56 & 30.43 & 79.32 & 90.48 & 76.39 & 75.10 & 64.18 & 76.17 & 64.94 & 59.67 & 42.64 & 65.53 \\
\rowcolor{openweight} Qwen3-1.7B-Base & 1.7B & 65.47 & 37.60 & 80.34 & 91.89 & 79.63 & 74.69 & 60.47 & 76.06 & 68.68 & 59.19 & 51.18 & 67.75 \\
\rowcolor{openweight} Qwen3.5-2B-Base & 2B & 63.95 & 37.34 & 83.73 & 93.12 & 85.63 & 73.05 & 63.52 & 76.50 & 69.14 & 58.80 & 60.37 & 69.56 \\
\rowcolor{openweight} Gemma-2-2B & 2B & 55.22 & 24.40 & 66.44 & 82.54 & 71.83 & 66.18 & 35.22 & 66.10 & 65.86 & 52.01 & 35.88 & 56.52 \\
\rowcolor{openweight} Gemma-3-1B-PT & 1B & 26.17 & 11.77 & 26.78 & 26.46 & 62.91 & 20.56 & 25.65 & 49.40 & 32.55 & 49.57 & 26.89 & 32.61 \\
\rowcolor{openweight} Gemma-4-E2B-Base & E2B & 59.72 & 25.55 & 73.56 & 86.24 & 77.28 & 65.77 & 44.38 & 67.79 & 65.66 & 52.64 & 39.69 & 59.84 \\
\rowcolor{openweight} Llama-3.2-3B & 3B & 57.55 & 27.71 & 72.88 & 84.83 & 76.02 & 66.91 & 48.12 & 76.06 & 64.28 & 52.57 & 38.01 & 60.45 \\
\rowcolor{openweight} LFM2.5-1.2B-Base & 1.2B & 57.50 & 17.74 & 82.03 & 87.83 & 74.40 & 59.13 & 54.20 & 74.97 & 59.26 & 57.54 & 33.54 & 59.83 \\
\rowcolor{openweight} Falcon-H1-1.5B-Base & 1.5B & 65.32 & 36.44 & 81.02 & 90.30 & 78.84 & 71.25 & 62.24 & 75.73 & 67.30 & 57.77 & 45.35 & 66.51 \\
\rowcolor{openweight} Falcon-H1-Deep-1.5B-Base & 1.5B & 68.74 & 42.99 & 85.08 & 93.12 & 85.23 & 72.40 & 65.24 & 76.61 & 70.83 & 60.14 & 41.86 & 69.29 \\

\midrule
\multicolumn{14}{l}{\textit{\textbf{Open-Recipe Models}}} \\
\rowcolor{fullyopen} Instella-3B & 3B & 58.55 & 25.32 & 74.92 & 87.30 & 76.76 & 75.43 & 59.92 & 73.23 & 69.91 & 53.99 & 39.11 & 63.13 \\
\rowcolor{fullyopen} OLMoE-1B-7B-0125 & A1B/7B & 54.11 & 19.67 & 66.44 & 84.30 & 72.97 & 62.82 & 36.54 & 71.16 & 65.15 & 50.28 & 30.74 & 55.83 \\
\rowcolor{fullyopen} Yulan-Mini-2.4B & 2.4B & 51.74 & 23.68 & 65.08 & 82.54 & 78.47 & 66.18 & 59.92 & 76.22 & 63.31 & 62.35 & 44.12 & 61.24 \\
\rowcolor{fullyopen} SmolLM3-3B-Base & 3B & 62.24 & 39.02 & 79.32 & 89.95 & 84.80 & 72.73 & 70.23 & 81.12 & 70.62 & 64.25 & 37.52 & 68.35 \\
\rowcolor{fullyopen} MiniCPM5-1B-Base & 1B & 45.01 & 21.98 & 46.44 & 70.90 & 73.12 & 56.92 & 51.51 & 71.87 & 50.15 & 54.06 & 36.51 & 52.59 \\
\rowcolor{fullyopen} MobileLLM-R1-950M-base & 950M & 48.24 & 21.95 & 58.31 & 78.13 & 78.87 & 61.59 & 54.16 & 73.50 & 59.77 & 55.33 & 34.15 & 56.73 \\

\midrule
\multicolumn{14}{l}{\textit{\textbf{Ours}}} \\
\rowcolor{ours} Puro-2B (\$4.4K) & 2B & 54.82 & 24.02 & 66.78 & 84.66 & 77.34 & 67.65 & 63.62 & 75.84 & 62.90 & 59.91 & 41.14 & 61.70 \\
\rowcolor{ours} Puro-2B & 2B & 57.44 & 25.81 & 74.24 & 86.07 & 76.85 & 67.81 & 63.21 & 75.52 & 63.61 & 60.62 & 42.05 & 63.02 \\

\bottomrule
\end{tabular}%
}
\end{table}

\paragraph{Reasoning and knowledge.}
\Cref{tab:reasoning_knowledge} shows that \sys achieves a mean score of 63.02
over the eleven reasoning and knowledge benchmarks. It outperforms Qwen2-1.5B
by 2.48 points and is within 2.51 points of Qwen2.5-1.5B. Within the open-recipe
group, \sys surpasses OLMoE-A1B/7B, Yulan-Mini-2.4B, MiniCPM5-1B-Base, and
MobileLLM-R1-950M-base, while trailing the larger Instella-3B by only 0.11
points. SmolLM3-3B-Base achieves a higher average, but it is also a larger
model. These aggregate results show that \sys remains competitive with larger
open-recipe models and retains a 6.29-point reasoning/knowledge advantage over
MobileLLM-R1. Together with the reduced training cost enabled by low-precision
training, this performance indicates a favorable balance between broad
capability and replication cost.

\FloatBarrier

\subsection{Cost Estimation}
\label{sec:cost-estimation}

\subsubsection{Cost-Saving Factors}
We estimate four cost factors separately: hardware price efficiency, FP8 throughput, MuonH quality-equivalent compute, and the joint Phase~2 recipe. Each factor uses its own reference comparison, so the reported values are not measurements of one sequential end-to-end speedup.
The product of these components can only be treated as illustrative transfers.  Hardware
uses a specification-and-price accounting proxy, blockwise FP8 uses quality-adjusted throughput with MuonH fixed, and the complete MuonH recipe uses quality-equivalent compute with blockwise FP8 fixed. Phase~2 first fits a scaling law of compute cost using the UD recipe~(\Cref{fig:renminbi-scaling-law}), then measures the equivalent costs of the UD recipe for the CD and canonical CMA endpoints to quantify the corresponding gains. 
These comparisons do not form a joint factorial ablation due to different references, so their products are accounting estimates rather than directly measured end-to-end speedups\footnote{For example, MuonH ablation and FP8 ablation use token-per-parameter of 20 (TPP=20) while the production training is an overtrained setup (TPP=700). We run the alternative compute-optimal setup due to limited compute.}.

For the ladder evidence below, $C=6ND$ denotes theoretical training compute, where $N$ is the scaling parameter count and $D$ is the number of training tokens.  A compute-equivalent multiplier $\kappa$ means that a variant trained with $C/\kappa$ reaches the fitted loss of its baseline at $C$.  For precision, we define BF16-equivalent compute retention as $\rho_C=C_{\mathrm{BF16}}/C_{\mathrm{FP8}}$ at matched fitted loss; this converts the vertical precision gap into a horizontal compute adjustment.

\paragraph{RTX~5090 exhibits a high cost-performance ratio and utilization.}
Under the specifications and prices in~\Cref{tab:gpu-cost-efficiency}, the RTX~5090 proxy provides \HardwareCostEfficiencyMultiplier the peak BF16 compute per unit price of H200 and \ensuremath{2.74\times} the peak FP8 compute per unit price of H200. Moreover, with training configuration tuning, as shown in \Cref{sec:infra:parallel}, our recipe can achieve 73\% MFU in mixed-precision training, making good use of the RTX~5090's capability.

\begin{figure}[H]
    \centering
    \begin{subfigure}[t]{0.49\linewidth}
        \centering
        \includegraphics[width=\linewidth]{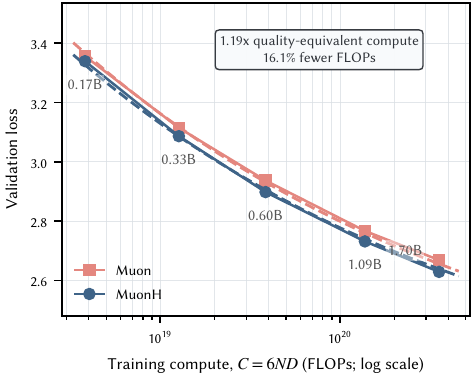}
        \caption{Complete MuonH versus the tuned Muon baseline. }
        \label{fig:muonh-compute-scaling}
    \end{subfigure}\hfill
    \begin{subfigure}[t]{0.49\linewidth}
        \centering
        \includegraphics[width=\linewidth]{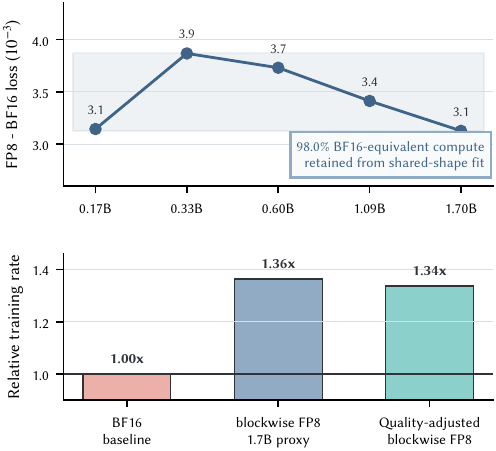}
        \caption{Blockwise FP8 quality and throughput. The upper figure shows the FP8 validation loss penalty across 5 scales; the lower figure shows wall-time speedup and net benefit of FP8 after considering the loss penalty.
        }
        \label{fig:fp8-ladder-summary}
    \end{subfigure}
    \caption{Matched scaling-ladder evidence behind two cost factors.
    \textbf{Left:} Validation loss versus theoretical compute for tuned Muon
    and MuonH recipes with blockwise FP8 fixed; dashed curves are
    the shared-floor, shared-exponent fit. The boxed note reports a fitted
    quality-equivalent compute multiplier. \textbf{Right:} With MuonH fixed, the FP8-minus-BF16 validation-loss gap remains within 0.0031--0.0039 across five scales. Only
    the 1.7B throughput ratio is used as the 2B proxy; its
    \FPEightThroughputSpeedup measured gain is multiplied by
    \FPEightPrecisionRetention fitted BF16-equivalent compute retention to
    obtain a \FPEightNetSpeedup matched-quality speedup, where
    $\rho_C=C_{\mathrm{BF16}}/C_{\mathrm{FP8}}$ is derived from the shared-shape
    loss fit.  Complete setups and
    fit diagnostics remain in~\Cref{app:muonh-scaling-ladder}.}
    \label{fig:cost-factor-ladders}
\end{figure}

\paragraph{MuonH provides a quality-equivalent compute shift in the matched ladder.}

Under the definition above, the quality-equivalent compute reduction is
$1-1/\kappa$.
The matched scaling ladder in~\Cref{fig:muonh-compute-scaling} estimates a
\MuonHComputeMultiplier compute-equivalent multiplier for the complete MuonH
recipe relative to Muon, corresponding to \MuonHComputeSaving less theoretical
compute at matched validation loss. The scaling estimation is performed over TPP=20 ladder with a shared-shape fit to transfer to production run.
At the production run of 2B
parameters and 1.4T tokens, using this efficiency estimation, the MuonH optimizer can use \TargetMuonHCompute FLOPs to match \TargetTrainingCompute FLOPs with the Muon baseline.

\paragraph{Blockwise FP8 retains a stable quality fraction while accelerating the 2B proxy.\label{sec:recipe:cost:fp8-save}}

A precision format may change both execution rate and model quality, so its net
efficiency multiplier is the product of throughput speedup and effective-compute
retention.  As summarized in~\Cref{fig:fp8-ladder-summary}, blockwise FP8 has a
small validation-loss gap at all five ladder scales.  The shared-shape fit gives
$\rho_C=\FPEightPrecisionRetention$.  Matching BF16 quality therefore requires
approximately \FPEightPrecisionPenalty additional
nominal compute. 
At 1.7B, the closest throughput proxy to the 2B production
model, median throughput improves by \FPEightThroughputSpeedup; charging the
quality penalty yields a \FPEightNetSpeedup net speedup, or
\FPEightGPUHourSaving fewer GPU-hours at matched quality. Then, by this estimation, the 2B-parameter, 1.4T-token canonical run would consequently use \TargetFPEightGPUHours FP8 GPU-hours to achieve an equivalent performance using approximately \TargetBFSixteenGPUHours BF16 GPU-hours.

\paragraph{The Puro Cost Scaling Law summarizes recipe-specific scale-down behavior.}

Starting from the same Phase~1 checkpoint, we continue Phase~2 training with
UD at several budgets and use these runs to characterize the
empirical cost--performance relationship, which we refer to as the Puro Cost
Scaling Law. It is a scale-down relationship for this recipe, not a universal
law across model families. %
Here, we focus on the scale-down trend to help the community with limited compute and time budgets decide a feasible performance target.
We fit $P=a+b\log_2(C-C_{\mathrm{P1}})$, where $P$ is the unweighted
arithmetic mean over the 15 benchmarks reported
in~\Cref{tab:math_code_capabilities,tab:reasoning_knowledge}, $C$ is total
reproduction cost, and $a$ and $b$ are the fitted intercept and slope. The
checkpoint-level values of $P$ used in the fit are reported
in~\Cref{tab:puro-scaling-evaluations}. Costs shown in the figure use the USD conversion
in~\Cref{tab:cost-reference-prices}. We set
$C_{\mathrm{P1}}=\text{USD }1.84$K from the measured Phase~1 active-training
GPU-hours and hold it fixed while fitting performance against the incremental
Phase~2 cost.
The observed UD checkpoint at approximately \$4.4K already exceeds
Qwen2-1.5B without CMA.
This shifted fit reduces in-sample RMSE from 0.452 for the unshifted log-cost
fit to 0.209. Inverting it at the CD endpoint score places its
UD cost equivalent at approximately \$11.36K, or 1.65$\times$ the
production recipe's \$6.89K rental-equivalent estimate from measured
active-training GPU-hours. Inverting the same fit at the reported Puro-2B
aggregate, which additionally includes a constant-LR continuation and averages
the last six checkpoints, gives approximately \$16.55K, or 2.40$\times$ the production cost.
The canonical CMA endpoint optimizes ordering, continuation, and checkpoint averaging together. We therefore report its fitted cost equivalent as a recipe-level point estimate rather than as a curriculum-only gain or a universal cost law.

\subsubsection{Cross-Model Reproduction Cost\label{sec:recipe:cost:reproduction}}
\paragraph{Estimation protocol.}

For~\Cref{fig:pareto-optimal}, we prioritize reported monetary cost or accelerator
usage. We use reported GPU-hours, or GPU-hours precisely derived from other
reported statistics, and convert them with the reference rental rates listed
in~\Cref{tab:cost-reference-prices} in the appendix. When no such GPU-hour basis is
available, we use the token-based route and estimate compute as $C=6ND$, using
the reported accelerator and MFU when available and H100-equivalent GPU-hours
at 70\% MFU otherwise. Models that disclose neither source are omitted. For \sys, we use measured active-training GPU-hours: 6,009.46 GPU-hours
for Phase~1 and 16,504.95 GPU-hours for Phase~2. We convert their sum using
the RTX~5090 rate in~\Cref{tab:cost-reference-prices}.
Across all models, this hierarchy distinguishes reproduction-cost estimates based on reported monetary costs or accelerator usage (\emph{reported compute}) from token-based estimates (\emph{estimated compute}).
 The complete coordinate construction,
including the treatment of special cases, and the Pareto-frontier procedure are
given in~\Cref{app:figure-one-calculation}.

\paragraph{Pareto-frontier result.}
Under the stated accelerator-cost assumptions, \sys appears above and to the
left of the comparison-model frontier in~\Cref{fig:pareto-optimal}, combining
higher average performance with a lower estimated replication cost. This
advantage is clearest relative to the open-recipe models.

\subsection[Post-Training]{Post-Training Results and Analysis}
\label{sec:posttrain-ablation}

We conduct an ablation study of the two Phase~2 designs to examine how the pretraining recipe affects the results after SFT. The uniform data ordering with learning-rate decay (UD) recipe trains the model on globally reshuffled Phase~2 tokens, while the curriculum model averaging (CMA) recipe follows a data curriculum, keeps the learning rate constant near the end of training, and averages the final checkpoints (\Cref{sec:recipe:curriculum}). We apply the same SFT procedure to the two pretraining checkpoints and compare the resulting UD-based and CMA-based models. The experimental setup is detailed in~\Cref{sec:recipe:posttrain} and we analyze the results in this section.

\paragraph{GSM8K-based SFT.}

\begin{figure}[ht]
    \centering
    \begin{subfigure}[t]{0.48\textwidth}
        \centering
        \includegraphics[width=\linewidth]{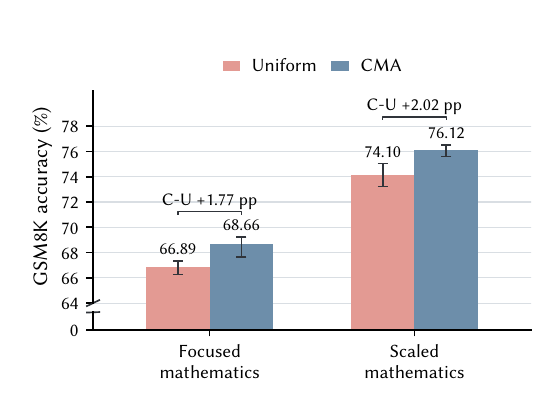}
        \caption{
        GSM8K-based SFT and Math\&Code SFT with replay.
        }
        \label{fig:posttrain-scaled-gsm8k}
    \end{subfigure}%
    \hspace{0.03\textwidth}%
    \begin{subfigure}[t]{0.48\textwidth}
        \centering
        \includegraphics[width=\linewidth]{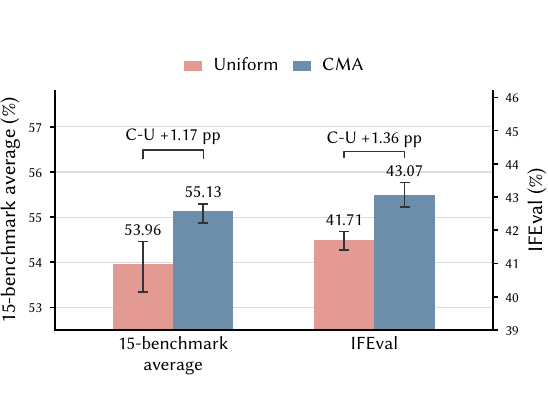}
        \caption{
        Tulu-3 mixed-domain SFT.
        }
        \label{fig:posttrain-general-tulu}
    \end{subfigure}
    \caption{Post-training results. \textbf{(a)} Final GSM8K results under GSM8K-based SFT and Math\&Code SFT with replay. Bars report the average GSM8K accuracy over three repeated runs, with error bars spanning the minimum and maximum. UD (uniform data with LR decay) and CMA denote SFT initialized from the corresponding pretraining checkpoints.  \textbf{(b)} Tulu-3 mixed-domain SFT summary. The two groups show the average over 15 benchmarks in \Cref{tab:math_code_capabilities,tab:reasoning_knowledge} and the additional IFEval score. Each group contains paired results. Detailed task-level results in~\Cref{tab:posttrain-broad-details}.}
\end{figure}

The GSM8K-based SFT experiment compares the UD and CMA checkpoints without pretraining replay. After SFT, the UD-based model reaches 66.89\% mean GSM8K accuracy, while the CMA-based model reaches 68.66\%. On average over three random seeds, the CMA-based model is ahead by 1.77 percentage points
(\Cref{fig:posttrain-scaled-gsm8k} and~\Cref{tab:focused-gsm-per-run}).
We use flexible answer extraction to limit score differences caused by output formatting. We then compare the sets of questions answered correctly by the two models. Under this protocol, the CMA-based model still solves more questions that the UD-based model misses. The remaining advantage is therefore better explained by stronger mathematical capability than by answer formatting.

\paragraph{Math\&Code SFT with replay.}

The Math\&Code SFT with replay experiment extends the comparison to a larger mathematics mixture, together with code data and a small replay component. It also uses a longer SFT process. One possible explanation for the base-model difference is a temporary advantage from placing higher-quality data near the end of pretraining. A longer SFT process on a different data distribution tests whether this advantage persists.
After this longer SFT process, the UD-based model reaches 74.10\% mean GSM8K accuracy, while the CMA-based model reaches 76.12\%. On average over three random seeds, the CMA-based model is ahead by 2.02 percentage points
(\Cref{tab:math_code_capabilities,tab:posttrain-scaled-data}). The CMA advantage therefore remains after more SFT updates on a broader data distribution. The gap is also slightly larger than in the GSM8K-based SFT setting.

\paragraph{Tulu-3 mixed-domain SFT.}

The first two setups focus primarily on mathematics. To examine whether the difference extends beyond this domain, we use Tulu-3 mixed-domain SFT and evaluate the resulting models on 15 benchmarks
(\Cref{tab:math_code_capabilities,tab:reasoning_knowledge}). As shown in \Cref{fig:posttrain-general-tulu}, the CMA-based model scores 1.17 points higher on average and performs better on 10 of the 15 benchmarks. The aggregate advantage therefore extends beyond mathematics, although the gains are not uniform across tasks.
The task-level results show a different balance across capabilities rather than a uniform improvement. We separately evaluate instruction following on IFEval, where the CMA-based model scores 1.36 points higher than the UD-based model. The results are averaged over three repeated runs, which show the same overall trend. Further details are provided in~\Cref{app:posttrain-details}.

Taken together, the three experiments show that the difference between the two pretraining recipes remains after increasingly broad SFT. The CMA-based models retain stronger mathematics performance in both the GSM8K-based SFT and Math\&Code SFT with replay settings. They also maintain a higher aggregate score after Tulu-3 mixed-domain SFT and perform better on instruction following. Because the task-level gains are uneven, these results do not imply that CMA improves every capability uniformly. This variation makes more controlled curriculum design an important direction for future work.

\section{Related Works}

\subsection{Open-Recipe Language Models}

The core science and engineering of large-scale pretraining remain difficult to
study when leading systems are either closed or release weights without the data
and training recipe needed to reconstruct the run~\citep{luo2025pcmind}. The
term ``open model'' therefore covers substantially different release practices.
Widely used model families such as Qwen~\citep{qwen2,qwen2025qwen25technicalreport,
qwen3,qwen35}, Gemma~\citep{morgane2024gemma2,gemma3,gemma4},
Llama~\citep{llama3.2}, LFM~\citep{liquidai2025lfm2}, and
Falcon~\citep{zuo2025falconh1} publish model weights and enough
implementation information for evaluation and deployment. However, their exact
pretraining corpora, sample order, and complete training state are not released,
so an independent group cannot reconstruct the original training run. We refer
to these releases as \emph{open-weight}. In contrast, a \emph{open-recipe}
release aims to expose the model weights, training data or a reconstructible
data recipe, training and evaluation code, and sufficiently detailed
configurations and intermediate artifacts to support scientific inspection and
reproduction~\citep{groeneveld-etal-2024-olmo,olmo20252olmo2furious}. Such
releases turn a final checkpoint into a reproducible account of how the model
was built, making controlled study possible for resource-limited research
communities.

Open-recipe projects provide the broader experimental stack needed to study
pretraining itself. Pythia releases ordered data and dense intermediate
checkpoints; OLMo and OLMoE release data, training and evaluation code,
checkpoints, and logs; and projects such as SmolLM3, Yulan-Mini-2.4B, and Instella
provide public data mixtures and reproducible training
configurations~\citep{biderman2023pythia,groeneveld-etal-2024-olmo,muennighoff2025olmoe,
smollm3,yiwen-etal-2025-yulan,liu2025instella}. These resources allow other
researchers to investigate learning dynamics, data selection, scaling, and
optimization without relying on undocumented industrial pipelines. Open-recipe
models can also be competitive: as shown
in~\Cref{tab:math_code_capabilities,tab:reasoning_knowledge}, models such as
SmolLM3-3B, Yulan-Mini-2.4B, and Instella-3B match or surpass several open-weight
models of comparable scale on parts of our evaluation, making this direction
important beyond reproducibility alone.

This openness and capability have nevertheless often required substantial
resources. Yulan-Mini-2.4B was trained on 48 A800 GPUs, SmolLM3 on 384 H100 GPUs,
and the first stage of Instella-3B on 128 MI300X
GPUs~\citep{yiwen-etal-2025-yulan,smollm3,liu2025instella}. Kaiyuan-2B, our previous
work, provides a more resource-conscious example by improving data and training
efficiency on Ascend 910A hardware~\citep{luo2025pcmind}. These examples expose
a remaining gap between openness and practical accessibility: a training
pipeline may be fully reproducible yet remain too costly for small research
groups to rerun. Addressing this gap requires considering training workload,
numerical precision, hardware accessibility, and explicit cost accounting
together.

\subsection{Low-Cost Language Model Pretraining}

Training cost is meaningful only under an explicit accounting boundary. Token
count and FLOPs alone capture only part of pretraining cost.
End-to-end replication cost also depends on the workload required to reach a
target quality, realized hardware throughput, and hardware price. We therefore
consider cost jointly with model quality rather than using any single measure
of compute or hardware efficiency as a proxy for low-cost training.

Low-precision training primarily reduces the cost of executing a fixed
workload. FP8 mixed precision accelerates the matrix multiplications that
dominate Transformer training while retaining higher precision where needed.
DeepSeek-V3 shows that this design can remain stable at large scale~\citep{
micikevicius2022fp8formats,deepseekai2025deepseekv3technicalreport}. Quartet II
extends this direction to NVFP4 and improves its numerical behavior~\citep{
panferov2026quartetii}. These methods improve execution efficiency for a given
training workload, but do not by themselves reduce the token or nominal FLOP
budget required to reach a target quality.

A complementary line of work reduces the training workload through model
architecture, data selection, staged training, and systems optimization.
TinyLlama, JetMoE, Yulan-Mini-2.4B, and Kaiyuan-2B explore different combinations of
these techniques~\citep{zhang2024tinyllama,shen2024jetmoe,yiwen-etal-2025-yulan,
luo2025pcmind}. Collectively, these projects show that useful small-model
capability can be obtained under substantially smaller compute budgets than
those of large-scale foundation models. The reported runs, however, generally
rely on specialized data-center accelerators, so the hardware barrier for a
small research group remains.

Hardware accessibility introduces a separate constraint. QLoRA reduces the
memory required to fine-tune a pretrained model, leaving the original
pretraining cost unchanged~\citep{dettmers2023qlora}. LLMQ studies full-model
pretraining on RTX~4090 GPUs using FP8, recomputation, sharding, and CPU
offloading~\citep{schultheis2025llmq}, while Quartet II measures the throughput
benefit of NVFP4 on a single RTX~5090~\citep{panferov2026quartetii}. The studies
considered here establish the feasibility of relevant training operations on
consumer GPUs, but focus primarily on systems feasibility or throughput rather
than on releasing a trillion-token-scale base model with an explicit
replication-cost boundary.

\sys lies at the intersection of these directions. It is a from-scratch 2B
base model trained on more than 1.4T tokens using FP8 on a multi-node RTX~5090
cluster, with measured accelerator usage and an explicit replication-cost
boundary. We therefore evaluate the training recipe and hardware platform end
to end, measuring both the resulting model quality and replication cost.
Together, these components provide researchers with a reproducible and
cost-accessible end-to-end pipeline for studying pretraining behavior and
testing alternative methods at meaningful scales.

\section{Conclusion and Future Direction}

In this report, we present a cost-efficient, hardware-accessible reproducibility recipe for language model pretraining, and use it to train the \fullsys model collection from scratch on consumer-grade RTX~5090 GPUs. Across different training budgets and recipe variants, the collection provides a direct view of the cost--performance tradeoff of our pipeline. A model from this collection reaches the performance of Qwen2-1.5B at about \$4.4K, while our best checkpoint approaches Qwen2.5-1.5B at the canonical accelerator-only cost of about \$6.9K. Based on these runs, we further formulate the \emph{Puro Cost Scaling Law} to characterize how model capability scales down with training cost under our recipe. More broadly, these results demonstrate that useful from-scratch pretraining at the billion-parameter scale is feasible with modest budgets and accessible hardware.

There remain several directions for extending this foundation. First, we plan to extend the reproducibility recipe beyond pretraining to post-training, with particular emphasis on developing and studying agentic capabilities. Second, we plan to broaden the architectural space beyond standard dense Transformers, including looped Transformers, linear-attention models, mixture-of-experts architectures, and other emerging designs. Third, we aim to expand the hardware recipes beyond RTX~5090 GPUs and cover a wider range of training budgets, making it easier to study how the optimal system and training choices change across hardware and scale. We view \fullsys not as an endpoint, but as evidence that affordable and inspectable model training is practical today, and as a baseline for future efforts to push the frontier toward lower cost, broader accessibility, and higher efficiency.

\section{Acknowledgments}

The authors would like to thank Yanfu Investments for providing computational resources.
Zhenbo Sun and Chenyi Dang contributed during the early stages of this work.
The authors also thank Kaiyue Wen, Kexian Tang, Minxing Yang, Zhan Ling, Haodong Wen, Shaowen Wang, Huaqing Zhang and Shuo Huang for their valuable feedback and insightful discussions, and Yanzheng Cai, Yuanwei Wang, Zhixuan Pan, Rui Chen, and Nuo Chen for their assistance with proofreading and manuscript revision. Multiple LLM-based tools were used to assist with drafting, proofreading, and language refinement.

\bibliographystyle{ACM-Reference-Format}
\bibliography{references, datasets}

\clearpage

\FloatBarrier

\appendix

\begin{center}
\Large
Appendices
\end{center}

\section{Limitations}
\label{app:limitations}

This section records limitations of the report's
claims that can be improved in the future.

\paragraph{Contamination and processed-data provenance.}
Most of the training mixture is assembled from already processed or filtered open datasets. Many upstream releases report some decontamination, but we do not provide a strict corpus-wide exact- or near-duplicate audit against every proxy and final benchmark. This limitation may affect the evidential strength of the curriculum comparison. 
In our post-training experiments, the advantage of curriculum over uniform data recipe persists beyond the base checkpoints, which can support its advantage beyond contamination. However, a corpus-wide contamination audit is still required to determine whether benchmark overlap contributes to that difference.

\paragraph{Chinese capability scope.}
The pretraining mixture includes Chinese data to provide a basic level of
Chinese-language coverage, but Chinese capability is not a primary target of
this release and was not the main axis used for optimization or headline model
selection.  We therefore omit Chinese benchmark scores from the release-facing
model-comparison tables, while retaining the Chinese proxy axis used to audit
data-mixture decisions.  The headline evaluation remains a mathematics, code,
reasoning, and knowledge study.  Future versions should make the intended
language targets explicit before tuning the mixture and evaluation protocol.

\paragraph{Overtrained dense regime.}
Puro-2B is a dense 2B-parameter model trained on approximately 1.4T tokens,
or about 700 tokens per parameter.  This is an overtrained, data-rich regime:
it is useful for studying data and recipe choices and for obtaining a compact
model with low inference cost, but it is not presented as compute-optimal.
The chosen point balances RTX~5090 memory and communication limits, attainable
benchmark quality, and the level of community support for a dense base model.

\paragraph{Scale-up and architecture scope.}
The Puro Cost Scaling Law is a fixed-2B, recipe-specific scale-down curve.  A
larger model or a larger world size would require a new communication and
memory design; for example, a smaller vocabulary or another embedding/LM-head
partition could be needed before scale-up.  The current report therefore does
not claim a model-size scale-up law. Especially when scaling up model size, the HBM memory capacity of RTX~5090 can be a bottleneck to deal with.

\section[Cost Assumptions]{Reproduction-Cost Assumptions}
\label{app:cost-assumptions}
\label{app:figure-one-calculation}
\begin{table}[ht]
\centering
\caption{Reference rental rates used to convert reported or estimated
accelerator usage to reproduction cost. The main figures report USD; RMB values
are retained for ledger traceability and use an exchange rate of
USD:CNY=6.8067 as of 1 July 2026.}
\label{tab:cost-reference-prices}
\begin{adjustbox}{max width=\linewidth}
\begin{tabular}{lrr}
\toprule
Reference accelerator & USD/GPU-hour & RMB/GPU-hour \\
\midrule
H200 & 4.00 & 27.23 \\
H100 & 3.25 & 22.12 \\
A100 80GB & 1.79 & 12.18 \\
A100 40GB & 1.29 & 8.78 \\
A800 80GB & 1.79 & 12.18 \\
RTX 5090 & 0.31 & 2.08 \\
\bottomrule
\end{tabular}
\end{adjustbox}
\end{table}

This section specifies how every coordinate and the comparison-model frontier
in~\Cref{fig:pareto-optimal} are constructed.  The plotted points are determined by
the benchmark scores, public source disclosures, and cost-accounting rules
described below.

\begin{table}[ht]
\centering
\caption{Cost and performance coordinates used in~\Cref{fig:pareto-optimal}.
GPU-hours and costs are rounded to the nearest unit for presentation;
USD costs use the exchange rate in~\Cref{tab:cost-reference-prices}, and the
figure uses the unrounded coordinates. Avg score is the
unweighted mean over the 15 tasks used by the figure.}

\label{tab:cost-performance-coordinates}
\setlength{\tabcolsep}{4.0pt}
\renewcommand{\arraystretch}{1.06}
\resizebox{\textwidth}{!}{%
\begin{tabular}{l l l r r r r}
\toprule
\textbf{Model} & \textbf{Evidence} & \textbf{Price basis}
& \textbf{GPU-hours} & \textbf{Cost (RMB)} & \textbf{Cost (USD)} & \textbf{Avg score} \\
\midrule
Qwen2-1.5B & 6ND, model tokens & H100 equivalent & 25,939 & 573,817 & 84,302 & 55.14 \\
Qwen2.5-1.5B & 6ND, family tokens & H100 equivalent & 66,700 & 1,475,530 & 216,776 & 60.73 \\
Qwen3-1.7B-Base & 6ND, model tokens & H100 equivalent & 147,261 & 3,257,664 & 478,597 & 65.27 \\
Gemma-2-2B & 6ND, model tokens & H100 equivalent & 12,560 & 277,860 & 40,821 & 48.60 \\
Gemma-3-1B-PT & 6ND, model tokens & H100 equivalent & 4,812 & 106,460 & 15,640 & 25.34 \\
Llama-3.2-3B & Reported GPU-hours & H100 & 460,000 & 10,176,017 & 1,495,000 & 52.17 \\
LFM2.5-1.2B-Base & 6ND, family tokens & H100 equivalent & 78,828 & 1,743,808 & 256,190 & 53.54 \\
Falcon-H1-Deep-1.5B-Base & 6ND, family tokens & H100 equivalent & 11,189 & 247,519 & 36,364 & 63.53 \\
Instella-3B & 6ND, model tokens & H100 equivalent & 30,851 & 682,470 & 100,265 & 55.20 \\
OLMoE-1B-7B-0125 & Reported count/time & H100 & 61,440 & 1,359,162 & 199,680 & 48.70 \\
Yulan-Mini-2.4B & 6ND, reported MFU & A100-80GB rate & 27,073 & 329,857 & 48,461 & 58.80 \\
SmolLM3-3B-Base & Reported count/time & H100 & 221,184 & 4,892,983 & 718,848 & 65.85 \\
MobileLLM-R1-950M-base & Reported count/time & A100-80GB rate & 36,864 & 449,151 & 65,987 & 54.23 \\
Puro-2B & Measured GPU-hours & RTX 5090 & 22,514 & 46,905 & 6,891 & 57.81 \\
\bottomrule
\end{tabular}%
}
\end{table}

\paragraph{Construction of the cost--performance figure.} The coordinates of different models are summarized in \Cref{tab:cost-performance-coordinates}. Our \sys includes a UD (\$4.4K) endpoint and the canonical CMA (\$6.9K) endpoint. The cost accounting of these two runs is summarized in \Cref{tab:pipeline-cost-summary}. More complete results of the Puro Cost Scaling Law, including different UD training variants and CD/CMA endpoints, are summarized in \Cref{fig:curriculum-endpoint-scores}. The implementation of CMA can be found in \Cref{fig:curriculum-model-average}. The checkpoint lineage and scale-down run details of Puro Cost Scaling Law are provided in \Cref{tab:puro-scaling-checkpoints,tab:puro-scaling-evaluations}, and the fitted scaling law can be found in \Cref{fig:renminbi-scaling-law}. 

\paragraph{Performance coordinate.}
For model $m$, the vertical coordinate is the unweighted arithmetic mean
\begin{equation}
    P_m = \frac{1}{15}\sum_{b\in\mathcal{B}} s_{m,b},
    \label{eq:figure-one-performance}
\end{equation}
where $s_{m,b}$ is the percentage score on benchmark $b$. The benchmark set
$\mathcal{B}$ contains GSM8K, MATH, sanitized-MBPP, HumanEval, MMLU, MMLU-Pro,
ARC-Challenge, ARC-Easy, BoolQ, CommonsenseQA, HellaSwag, PIQA, SocialIQA,
WinoGrande, and BBH. No benchmark is weighted by its number of examples. All
15 scores must be present for a plotted model. 

\paragraph{Cost evidence hierarchy.}
We assign the horizontal coordinate (reproduction cost) using the following ordered rules.
Reported monetary cost is used directly when it is available. Otherwise,
reported GPU-hours are converted using the corresponding
RMB/GPU-hour rate in~\Cref{tab:cost-reference-prices}. If a report gives only
GPU count $G$ and elapsed training time $T$, we first compute
$H=G T$ GPU-hours, with $T=24d$ for a duration of $d$ days. In all these cases,
the converted cost is
\begin{equation}
    C_m = H_m r_a,
    \label{eq:reported-gpu-hour-cost}
\end{equation}
where $r_a$ is the reference RMB/GPU-hour price for accelerator $a$.
The audit ledger retains this RMB value;~\Cref{fig:pareto-optimal} divides it by
6.8067 to report the horizontal coordinate in USD.

If neither monetary cost nor accelerator-hours are reported but a usable
training-token count is available, we estimate training compute with
$C=6ND$~\citep{kaplan2025scaling,hoffmann2022chinchilla}. For parameter count $N$, token count $D$, accelerator BF16 peak
throughput $F_a$ in TFLOP/s, and model FLOPs utilization (MFU) $\eta$, the estimate is
\begin{equation}
    H_m = \frac{6N_mD_m}{3600\times 10^{12}\times \,\eta F_a },
    \qquad C_m = H_m r_a.
    \label{eq:figure-one-6nd}
\end{equation}
For token-only entries, we use an H100-equivalent estimate with
$F_a=989.5$ TFLOP/s and $\eta=0.70$. Model-specific token counts are preferred;
when only a family-level training budget is disclosed, that use is marked as an
estimate in the coordinate table. Models lacking both accelerator usage and a
usable token count receive no cost coordinate and are omitted from the figure. As we posit an ideal MFU assumption in our accounting, we estimate the cost lower bound of reproducing our counterpart models in most cases.

\paragraph{Public sources for~\Cref{tab:cost-performance-coordinates}.}
The Qwen rows use the public training-token disclosures in the Qwen2, Qwen2.5,
and Qwen3 technical reports~\citep{qwen2,qwen2025qwen25technicalreport,qwen3}.
Gemma-2-2B and Gemma-3-1B-PT use Google's Gemma report and model-card
disclosures for the corresponding model sizes and token
budgets~\citep{morgane2024gemma2,google2024gemma2modelcard,gemma3,google2025gemma3modelcard}.
Llama-3.2-3B uses the official Llama 3.2 model card, which reports the 3B
checkpoint's H100 GPU-hours directly~\citep{meta2024llama32modelcard}.
LFM2.5-1.2B-Base uses Liquid AI's LFM2.5 release note for the disclosed
family-level training budget~\citep{liquidai2026lfm25blog}. The
Falcon-H1-Deep-1.5B-Base row uses the Falcon-H1 technical report for the token
budget and H100 training pool disclosure~\citep{zuo2025falconh1}. Instella-3B
uses the Instella report and model card for the two base-pretraining
stages~\citep{liu2025instella,amd2025instella3bcard}. OLMoE-1B-7B-0125 uses the OLMoE
paper's H100 count and elapsed-time disclosure~\citep{muennighoff2025olmoe}.
SmolLM3-3B-Base uses the Hugging Face training report for the reported H100
count and duration~\citep{smollm3}.

\paragraph{Special cases.}
Our Puro-2B point uses measured active-training time rather than $6ND$:
6,009.46 GPU-hours for Phase~1 plus 16,504.95 GPU-hours for Phase~2. Its local
RTX~5090 rate is
$12{,}000/(8\times30\times24)=2.0833$ RMB/GPU-hour, giving
$22{,}514.41\times2.0833=46{,}905.03$ RMB, or \$6,891.
Yulan-Mini-2.4B uses~\Cref{eq:figure-one-6nd} with its reported 51.57\% MFU and the
312-TFLOP/s A800 BF16 peak~\citep{yiwen-etal-2025-yulan}. Its GPU-hours are
priced with the unified A100 80GB rental rate in place of the rarely published
public A800 rental price.
MobileLLM-R1 reports 128 GPUs but not their
model~\citep{facebookresearch2025mobilellmr1}. We use the A100 80GB GPUs disclosed
by the original MobileLLM training-cost report as the MobileLLM-series
reference~\citep{facebookresearch2024mobilellm}. Two 4--5 day pretraining phases and two
1--2 day mid-training phases
imply 30,720--43,008 accelerator-hours. We price this range with the A100 80GB
rate and use the midpoint as its plotted coordinate.

\paragraph{Pareto frontier.}
To construct the dashed comparison frontier, we first remove Puro-2B. A
comparison model $m$ is Pareto-optimal if there is no other comparison model
$j$ such that
\begin{equation}
    C_j \le C_m, \qquad P_j \ge P_m,
    \label{eq:figure-one-pareto}
\end{equation}
with at least one strict inequality. Operationally, at an identical cost we
retain only the highest-scoring point, sort the remaining points by increasing
cost, and retain a point only when its score is strictly larger than every
lower-cost score seen so far. The retained comparison points are connected in
cost order.

\section[Training Details]{Production Training Details}
\label{app:training-details}

\paragraph{Production run setup.}

\begin{figure}[ht]
    \centering
    \includegraphics[width=0.92\linewidth]{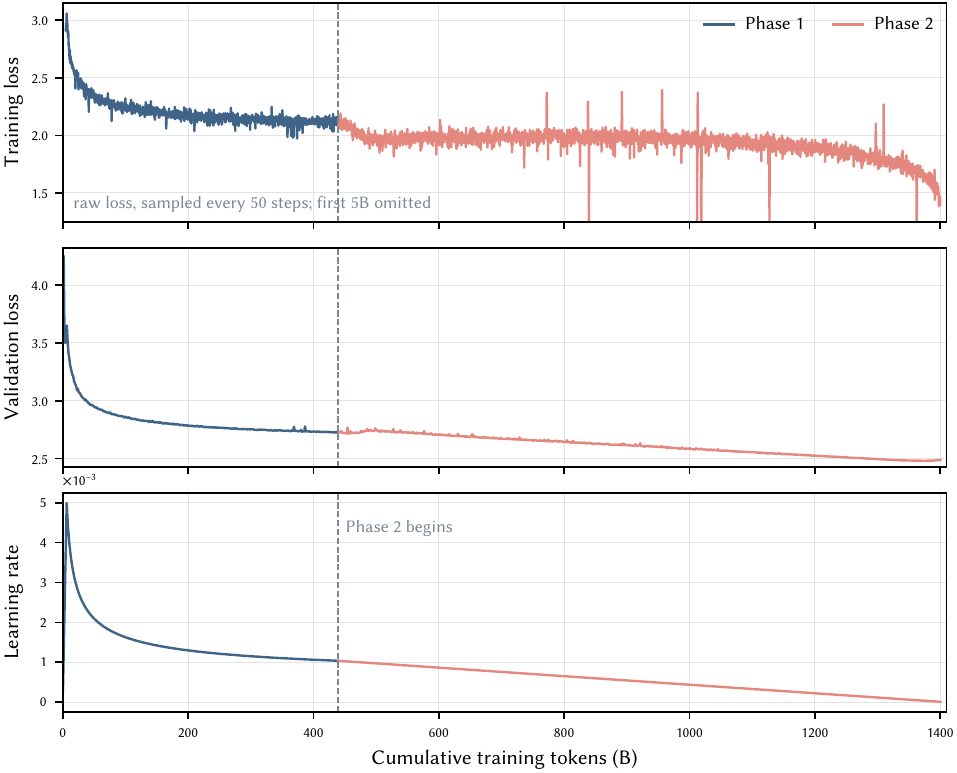}
    \caption{Concatenated production traces.  Training loss is raw (not smoothed) and sampled every 50 optimizer steps; the first 5B tokens are omitted from that panel to keep the later trajectory legible.  Every logged validation event is retained.  The vertical line marks the Phase~1/Phase~2 run boundary; the Phase~2 run begins with the distribution transition. The learning rate schedule corresponds to the base LR and the Hyperball weight LR is 10 times base LR.
    }
    \label{fig:prom-2b-training-curves}
\end{figure}

Our canonical \sys run consists of two phases. Both phases use sequence length 4,096, global batch size 1,536, and micro-batch size 2. Selected matrix weights use MuonH with zero weight decay, while the remaining parameters use AdamW with weight decay 0.1; both phases use blockwise E4M3 FP8. The LR schedule below corresponds to the base learning rate, and the Hyperball weight LR is 10 times the base learning rate. Phase~1 trains on
\PhaseOneWandBPhaseTokens with \PhaseOneWandBWorldSize GPUs and reaches
validation loss \PhaseOneWandBFinalValidationLoss; its median measured
throughput is \PhaseOneWandBMedianThroughput TFLOP/s per GPU.
From Phase~1 to Phase~2, we extend our compute resources, from \PhaseOneWandBWorldSize to \PhaseTwoWandBWorldSize GPUs.
Phase~2 resumes from this checkpoint, consumes \PhaseTwoWandBPhaseTokens and continues to
\PhaseTwoWandBCumulativeTokens cumulative tokens, and ends at validation loss
\PhaseTwoWandBFinalValidationLoss with median throughput
\PhaseTwoWandBMedianThroughput TFLOP/s per GPU.  The learning rate decays from
$5\times 10^{-3}$ to $1.04\times 10^{-3}$ in Phase~1 and from
$1.04\times 10^{-3}$ to $1\times 10^{-5}$ in Phase~2, matching the power-then-
linear schedule shown in~\Cref{fig:prom-2b-training-curves}.  The Phase~2 token
ledger is summarized in~\Cref{tab:puro-scaling-checkpoints}; the exact transition
and pool accounting is frozen in~\Cref{tab:transition-token-ledger}.  The
materialized Phase~2 pool already includes the early transition range, while Phase~1 replay is an additional consumed stream.

\begin{table}[ht]
\centering
\small
\caption{
The resulting audit table combines the GPUs cluster, configuration, base learning rate schedule, and throughput.
}
\label{tab:prom-2b-training-setup}
\begin{adjustbox}{max width=\linewidth}
\begin{tabular}{lrrrllrr}
\toprule
Phase & Phase tokens & Cumulative & GPUs & TP/PP/DP & LR schedule & TFLOP/s/GPU \\
\midrule
Phase 1 & 438.84B & 438.84B & 24 & 1/2/12 & $5.00\times 10^{-3} \rightarrow 1.04\times10^{-3}$ (power) & 238 \\
Phase 2 & 959.99B & 1.40T & 96 & 1/4/24 & $1.04\times 10^{-3} \rightarrow 1.00\times 10^{-5}$ (linear) & 192 \\
\bottomrule
\end{tabular}
\end{adjustbox}
\end{table}

\begin{table}[ht]
\centering
\small
\caption{Optimizer parameter groups used in both production pretraining
phases. Learning rate entries are relative to the base schedule.}
\begin{tabular}{@{}>{\raggedright\arraybackslash}p{0.43\linewidth}lcc@{}}
\toprule
Parameter group & Optimizer & Weight decay & Learning rate \\
\midrule
Selected attention and MLP matrices
    & MuonH & 0 & $10\times$ base schedule \\
Embeddings, normalization layers, LM head, and other parameters
    & AdamW & 0.1 & Base schedule \\
\bottomrule
\end{tabular}
\label{tab:production-optimizer-parameter-groups}
\end{table}

\paragraph{Phase 1 learning rate schedule.}

The scalar learning rate is the base rate.  The MuonH-controlled matrix
groups apply an optimizer multiplier of 10, while norm-variant parameters use
their separately configured scalar optimizer.  With optimizer step $k$, the
Phase~1 base schedule is
\begin{equation}
\eta_{\mathrm{base}}(k)=
\begin{cases}
5\times10^{-3}\, k/1000, & 0\leq k\leq1000,\\[2pt]
5\times10^{-4}+4.5\times10^{-3}
\left(1+\dfrac{k-1000}{1000}\right)^{-1/2}, & k>1000.
\end{cases}
\label{eq:production-phase1-lr}
\end{equation}
Thus $p=1/2$, $\tau=1{,}000$ steps, warmup is 1,000 steps (1,536,000 samples), and the $5\times10^{-4}$ minimum is an asymptotic floor. The power decay shape follows the power schedule~\citep{shen2024power} and supports continual training in the first phase.

\paragraph{Phase 2 learning rate schedules.}

\begin{figure*}[ht]
    \centering
    \begin{subfigure}[t]{0.45\textwidth}
        \centering
        \includegraphics[width=\linewidth]{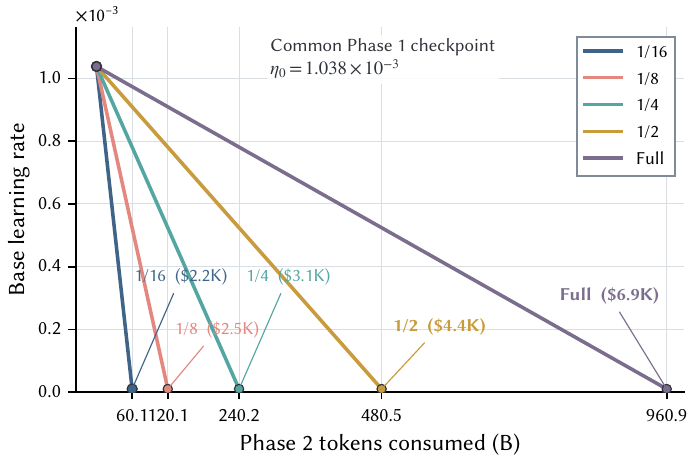}
        \caption{The five independent Phase~2 base learning rate schedules used
        in the UD points of~\Cref{fig:renminbi-scaling-law}. All runs load
        the same Phase~1 checkpoint and therefore share the same starting base
        rate. They differ only in decay horizon and all decay linearly to $10^{-5}$ for base LR.}
        \label{fig:phase2-uniform-budget-schedule}
    \end{subfigure}\hfill
    \begin{subfigure}[t]{0.54\textwidth}
        \centering
        \includegraphics[width=\linewidth]{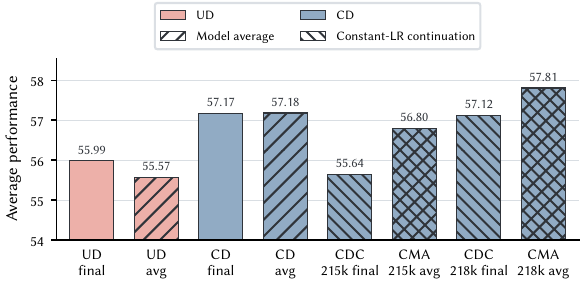}
        \caption{Endpoint evaluation scores for eight Phase~2 configurations. \emph{final} means using the final checkpoint, and \emph{avg} means computing the checkpoint average as discussed in \Cref{app:curriculum:averaging}. CDC/CMA \emph{215k} and \emph{218k} mean continual training starting from step 215k and 218k. The vertical axis is the same 15-benchmark performance defined in \Cref{tab:math_code_capabilities,tab:reasoning_knowledge}.
        The underlying endpoint scores are reported in~\Cref{tab:puro-scaling-evaluations}.}
        \label{fig:curriculum-endpoint-scores}
    \end{subfigure}
    \caption{\textbf{Phase~2 budget schedules and endpoint comparison.} The
    schedule panel gives the common Phase~1 starting point and the independent
    UD Phase~2 horizons. The endpoint panel compares the UD, CD, CDC, and CMA configurations.}
    \label{fig:phase2-schedule-endpoints}
\end{figure*}

Each Phase~2 run loads the same Phase~1 endpoint and therefore starts
from the same base learning rate, namely the terminal Phase~1 value
$\eta_0=1.04\times10^{-3}$.  We reset the Phase~2-local schedule
coordinate at that shared checkpoint, so $s=0$ always means the Phase~1
endpoint, not the run's recorded global optimizer step.  Thus, for every run
$j$, $\eta_{\mathrm{base},j}(0)=\eta_0$; the runs differ only in their
linear decay-sample budgets.  With decay-sample budget $S_j$ and terminal rate
$\eta_{\min}$, the base schedule is
\begin{equation}
\eta_{\mathrm{base},j}(s)=\eta_0+
\left(\eta_{\min}-\eta_0\right)\frac{s}{S_j},
\qquad 0\leq s\leq S_j.
\label{eq:production-phase2-lr}
\end{equation}
Here $s$ is Phase~2 samples consumed after the common checkpoint.  As in
Phase~1, MuonH-controlled matrix groups multiply each base rate by 10.  The
five schedules share the same starting rate and $\eta_{\min}=10^{-5}$, but have
different horizons of approximately 60.1,
120.1, 240.2, 480.5, and 960.9B tokens, as shown in \Cref{fig:phase2-uniform-budget-schedule}. These are the UD scaling points; the matched CD and CMA endpoints use the full Phase~2 budget. Our canonical CMA run uses the curriculum order and replaces the LR decay in the final steps (starting from step 218,000, lasting the last 29B tokens) of pretraining with constant-LR training. The details can be found in \Cref{app:curriculum-details}. 

\begin{figure}[ht]
\centering
\resizebox{\textwidth}{!}{\input{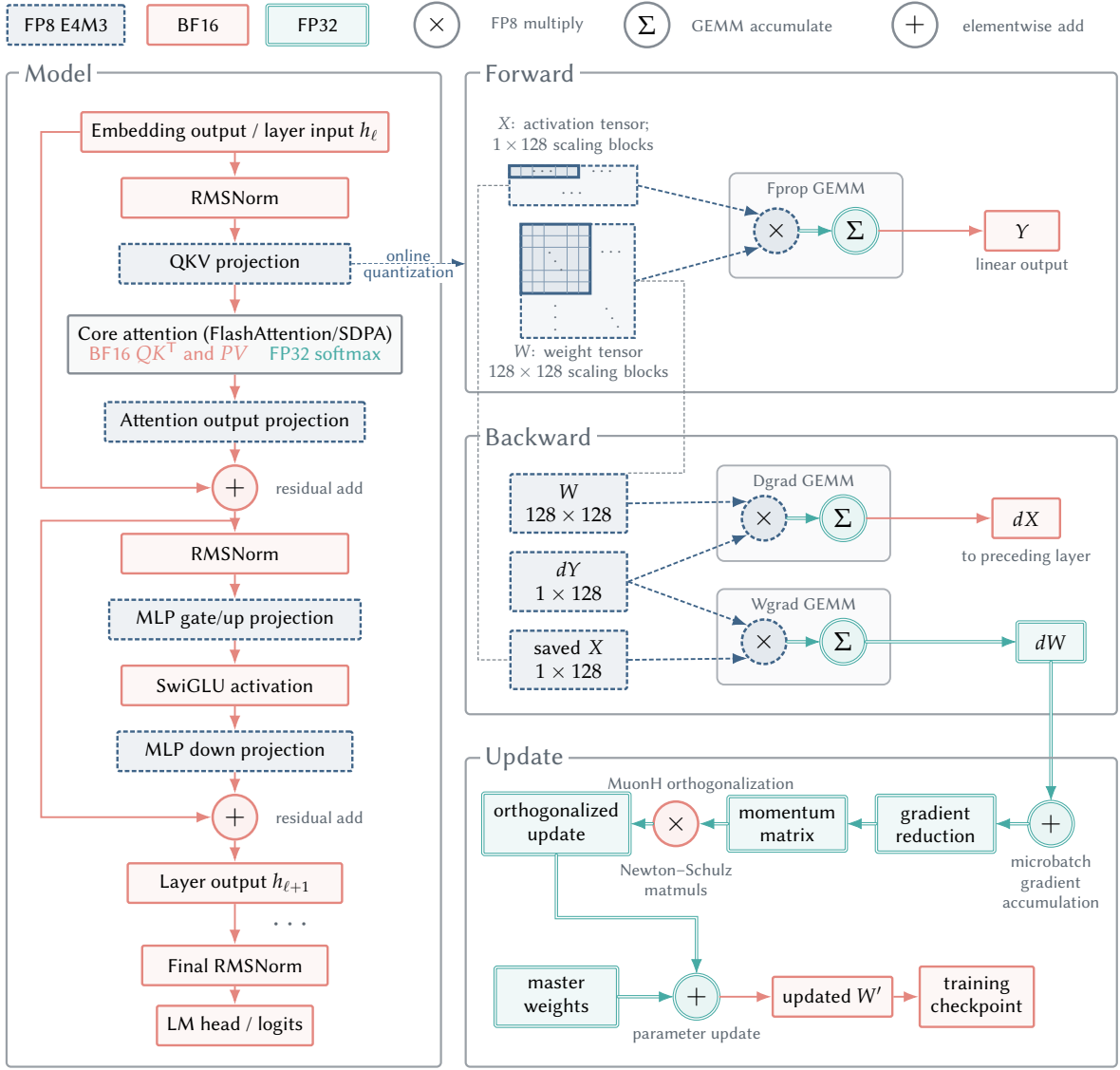}}
\caption{
Precision flow for the FP8 recipe. Linear-layer GEMMs use blockwise E4M3 for operands and saved activations. The surrounding outputs, communication tensors, and checkpoint weights remain BF16, while gradient and optimizer-state accumulation remains FP32.
}
\label{fig:fp8-precision-flow}
\end{figure}

\paragraph{FP8 implementation pipeline.} The complete pipeline of our FP8 implementation is presented in \Cref{fig:fp8-precision-flow}. The linear-layer GEMMs, including Fprop, Dgrad, Wgrad uses blockwise E4M3 for operands and activations. The hidden and residual flow, as well as LayerNorm and Embedding layers, use BF16. The gradient, optimizer-state accumulation and softmax remain FP32.

\section[Scaling Ladder]{MuonH Scaling-Ladder Analysis}
\label{app:muonh-scaling-ladder}

\paragraph{Experimental setup.}
We construct a five-model ladder spanning model scales from 0.17B to 1.7B.
Every model is trained from scratch for 20 tokens per scaling
parameter (TPP=20), following the compute-optimal token-to-parameter ratio reported in prior work~\citep{hoffmann2022chinchilla}.  All runs use the same data distribution, tokenizer,
sequence length of 4,096, global batch size of 512, validation set, and linear
learning rate decay with a 1\% warmup.
The validation set uses a subset of the Nemotron-CC corpus~\citep{su2025nemotroncctransformingcommoncrawl,nemotron_nana_v2}.
The exact architecture and token budgets
are shown in~\Cref{tab:muonh-scaling-ladder-setup}.  We compare two paired
recipes: (i) blockwise-FP8 MuonH against a tuned blockwise-FP8 Muon baseline to
evaluate the complete optimizer recipes, and (ii) blockwise-FP8 MuonH against
BF16 MuonH to isolate numerical precision.  MuonH uses a base LR of
$10^{-2}$ and a Hyperball multiplier of 2, hence a Hyperball weight LR of
$2\times10^{-2}$.  The updated ordinary-Muon baseline fixes weight decay at
0.8 and uses ordinary LRs $5\times10^{-3}$, $4\times10^{-3}$,
$4\times10^{-3}$, $3\times10^{-3}$, and $3\times10^{-3}$ from 0.17B through
1.7B.  The 0.17B and 0.33B choices are anchored by the reported LR/weight-decay
searches; the larger points are conservative size-scaled extrapolations.
The search results that motivate the first two LR/weight-decay anchors are tabulated in
\Cref{tab:muon-lr-wd-search}. The two
precision runs use identical MuonH hyperparameters.  Because the optimizer pair
uses tuned, non-identical learning rate schedules, this is a complete-recipe
comparison rather than an isolated hyperball-constraint ablation. These two experiment results are summarized in \Cref{fig:cost-factor-ladders}. 

\begin{table*}[ht]
    \centering
    \small
    \caption{Completed LR--weight-decay search results used to select the
    updated blockwise-FP8 Muon baseline anchors.  The search stages are labeled
    initial or extended.  Bold rows mark
    the best completed setting at each anchor size.
    }
    \label{tab:muon-lr-wd-search}
    \begin{tabular}{llrrr}
        \toprule
        Model & Search stage & Ordinary LR & Weight decay & Final validation loss \\
        \midrule
        0.17B & initial & $2\times10^{-3}$ & 0.1 & 3.413030 \\
        0.17B & initial & $2\times10^{-3}$ & 0.2 & 3.406015 \\
        0.17B & initial & $2\times10^{-3}$ & 0.4 & 3.391572 \\
        0.17B & initial & $2\times10^{-3}$ & 0.8 & 3.375170 \\
        0.17B & initial & $3\times10^{-3}$ & 0.1 & 3.414363 \\
        0.17B & initial & $3\times10^{-3}$ & 0.2 & 3.405061 \\
        0.17B & initial & $3\times10^{-3}$ & 0.4 & 3.383846 \\
        0.17B & initial & $3\times10^{-3}$ & 0.8 & 3.361907 \\
        0.17B & initial & $4\times10^{-3}$ & 0.1 & 3.413549 \\
        0.17B & initial & $4\times10^{-3}$ & 0.2 & 3.398726 \\
        0.17B & initial & $4\times10^{-3}$ & 0.4 & 3.381147 \\
        0.17B & initial & $4\times10^{-3}$ & 0.8 & 3.357248 \\
        0.17B & extended & $4\times10^{-3}$ & 1.6 & 3.352910 \\
        {\bfseries 0.17B} & {\bfseries extended} & $\mathbf{5\times10^{-3}}$ & $\mathbf{0.8}$ & $\mathbf{3.349810}$ \\
        0.17B & extended & $5\times10^{-3}$ & 1.6 & 3.360720 \\
        0.17B & extended & $6\times10^{-3}$ & 0.8 & 3.356287 \\
        0.17B & extended & $6\times10^{-3}$ & 1.6 & 3.370632 \\
        \midrule
        0.33B & initial & $2\times10^{-3}$ & 0.2 & 3.156636 \\
        0.33B & initial & $2\times10^{-3}$ & 0.4 & 3.142686 \\
        0.33B & initial & $2\times10^{-3}$ & 0.8 & 3.122845 \\
        0.33B & initial & $3\times10^{-3}$ & 0.2 & 3.150765 \\
        0.33B & initial & $3\times10^{-3}$ & 0.8 & 3.116963 \\
        0.33B & extended & $3\times10^{-3}$ & 1.6 & 3.117189 \\
        {\bfseries 0.33B} & {\bfseries initial} & $\mathbf{4\times10^{-3}}$ & $\mathbf{0.8}$ & $\mathbf{3.114583}$ \\
        0.33B & extended & $4\times10^{-3}$ & 1.6 & 3.130430 \\
        0.33B & initial & $5\times10^{-3}$ & 0.8 & 3.120008 \\
        0.33B & extended & $5\times10^{-3}$ & 1.6 & 3.139534 \\
        \bottomrule
    \end{tabular}
\end{table*}

\begin{table*}[t]
\centering
\small
\begin{adjustbox}{max width=\textwidth}
\begin{tabular}{lrrrrrrrrrrrr}
\toprule
Scale & $L$ & $d_{\mathrm{model}}$ & $d_{\mathrm{ff}}$ & $H_Q$ & $H_{KV}$ & $d_{\mathrm{ch}}$ & $d_{QK}$ & $d_V$ & MBS & Steps & Tokens & $C=6ND$ \\
\midrule
0.17B & 28 & 512 & 1536 & 8 & 8 & 64 & 128 & 128 & 4 & 1700 & 3.57B & $3.81\times10^{18}$ \\
0.33B & 28 & 768 & 2304 & 16 & 8 & 64 & 128 & 128 & 4 & 3100 & 6.50B & $1.27\times10^{19}$ \\
0.60B & 28 & 1024 & 3072 & 16 & 8 & 128 & 128 & 128 & 2 & 5400 & 11.32B & $3.85\times10^{19}$ \\
1.09B & 28 & 1536 & 4608 & 16 & 8 & 128 & 128 & 128 & 1 & 10200 & 21.39B & $1.37\times10^{20}$ \\
1.70B & 28 & 2048 & 6144 & 16 & 8 & 128 & 128 & 128 & 1 & 16400 & 34.39B & $3.55\times10^{20}$ \\
\bottomrule
\end{tabular}
\end{adjustbox}
\caption{Architecture and training setup of the MuonH scaling ladder. All models use a sequence length of 4,096, global batch size 512, 28 layers, and 20 training tokens per scaling parameter. $H_Q$ and $H_{KV}$ denote query and grouped-query KV heads; $d_{\mathrm{ch}}$ is the configured attention channel width (\texttt{kv\_channels}), and $d_{QK}$ and $d_V$ are the configured QK and value head dimensions.}
\label{tab:muonh-scaling-ladder-setup}
\end{table*}

\paragraph{Compute-equivalent fitting.}
For each run, we use the standard dense-transformer approximation
\begin{equation}
    C = 6ND,
\end{equation}
where $D$ is the number of training tokens and $N=D/20$ is the scaling
parameter count implied by the ladder construction.  To compare a baseline
$b$ and variant $v$, we jointly fit the ten observations with
\begin{equation}
    L_g(C) = L_{\infty} + A_g\left(\frac{C}{C_0}\right)^{-\alpha},
    \qquad g\in\{b,v\},
    \label{eq:ladder-shared-power-fit}
\end{equation}
sharing $L_{\infty}$ and $\alpha$ while allowing a group-specific amplitude.
The resulting horizontal multiplier is
\begin{equation}
    \kappa_{v\leftarrow b}
    = \left(\frac{A_b}{A_v}\right)^{1/\alpha}.
    \label{eq:ladder-compute-multiplier}
\end{equation}
Thus, the variant requires $C/\kappa_{v\leftarrow b}$ to match the fitted
baseline loss at compute $C$.  This restricted shared-shape fit estimates a
horizontal efficiency shift; it is not used to select a learning rate schedule
or to predict the absolute loss of the 2B production run.

As shown in~\Cref{fig:muonh-compute-scaling}, MuonH has a fitted
compute-equivalent multiplier of \MuonHComputeMultiplier relative to the updated
Muon baseline.
Leaving out one model size at a time gives \MuonHComputeMultiplierLOO, showing
that the estimate is not determined by one endpoint.  Applying this horizontal
factor to the \TargetTrainingCompute{}-FLOP budget of a 2B/1.4T-token run gives
\TargetMuonHCompute FLOPs at matched Muon loss, corresponding to
\MuonHComputeSaving less compute.  This transfer assumes that the
optimizer's horizontal efficiency factor persists beyond the 20-TPP ladder;
we therefore report it as an estimated compute-equivalent saving rather than an
absolute large-run loss prediction.  The leave-one-out range measures
sensitivity to individual ladder points, not uncertainty from the selected
functional form.

\paragraph{Precision and throughput decomposition.}
To compare training cost without including evaluation, checkpoint, or
experiment-tracking intervals, we convert theoretical compute to GPU-hours using
\begin{equation}
    H = \frac{C}{3600\times 10^{12}\times  \bar{p}},
    \label{eq:ladder-gpu-hours}
\end{equation}
where $\bar{p}$ is the median per-GPU throughput in TFLOP/s over training-step
history.  Because $H$ is measured in aggregate GPU-hours, the world-size factor
cancels.~\Cref{fig:fp8-ladder-summary} reports the matched validation-loss gap
across all five sizes but shows throughput only for the 1.7B run.  Smaller models
have smaller GEMMs and need not expose the throughput behavior relevant to the
2B production configuration, so their throughput ratios are omitted from the
main figure rather than treated as part of the production proxy.

At fixed theoretical compute, blockwise FP8 is consistently 0.0031--0.0039 higher in validation loss than BF16 across the ladder. A BF16-versus-FP8 fit in theoretical-compute space estimates \FPEightPrecisionRetention effective-compute retention; leave-one-size-out fits range from \FPEightPrecisionRetentionLOO. This corresponds to a \FPEightPrecisionPenalty nominal-compute penalty at matched quality. At 1.7B, median throughput increases by \FPEightThroughputSpeedup.
Using this largest-scale measured throughput ratio as a proxy for the 2B setting and accounting for the fitted precision penalty, we obtain a ladder-derived, quality-adjusted speedup estimate of \FPEightNetSpeedup, corresponding to an estimated \FPEightGPUHourSaving reduction in GPU-hours at matched quality. The FP8 speedup is a ladder-derived estimate. It combines the five-size quality fit with the measured 1.7B throughput ratio, then uses that ratio as a proxy for the 2B production setting.

\begin{figure}[ht]
    \centering
    \includegraphics[width=0.98\linewidth]{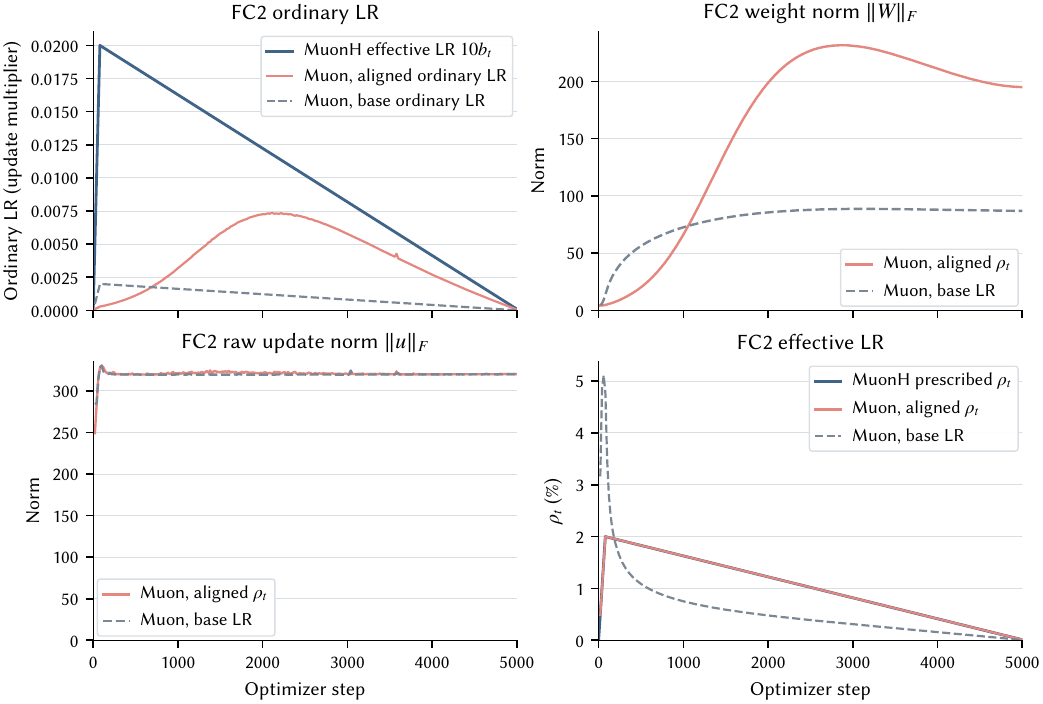}
    \caption{FC2 represents the second layer of FFN. Here we present the FC2 diagnostics for the BF16 MuonH comparison in \Cref{fig:hyperball-trace-alignment}.
    The upper-left panel shows the ordinary LR, namely the scalar coefficient that multiplies
    the ordinary-Muon optimizer update; the lower-right panel shows the induced
    effective LR. The MuonH curves in both panels are reconstructed from the shared schedule, while the norm panels show the aligned/base Muon traces.}
    \label{fig:hyperball-diagnostics-2x2}
\end{figure}

\paragraph{MuonH diagnostic traces.}
To make the optimizer comparison more concrete, this appendix plots \Cref{fig:hyperball-diagnostics-2x2}, the
learning rate and update diagnostics for the FC2\footnote{FC2 represents the second layer of FFN.} matrix in the 170M-parameter
BF16 experiment used for~\Cref{fig:hyperball-trace-alignment}.  The aligned and
base Muon runs record per-group effective learning rates, weight norms, and raw
update norms.  The MuonH run records the shared base schedule rather than these
per-group fields, so its effective LR target is reconstructed as
$10\eta_t^{\mathrm{base}}$ from the configured Hyperball multiplier.  Its weight norm is therefore
not drawn as an inferred measurement; projection keeps it at the prescribed
initial radius.

\section[Post-Training Details]{Post-Training Details}
\label{app:posttrain-details}

This appendix provides the data accounting, matched training controls, evaluation details, and per-run results for the post-training experiments in
\Cref{sec:recipe:posttrain,sec:posttrain-ablation}. It follows the same progression as the main text. The GSM8K-based SFT provides a targeted mathematics comparison, the Math\&Code SFT with replay tests whether the difference persists after a longer and broader SFT process, and the Tulu-3 mixed-domain SFT examines transfer beyond mathematics.

\subsection{Shared Construction and Evaluation Protocol}

\paragraph{Dataset construction and accounting.}

The three SFT settings use the same conversation schema, Qwen3-1.7B tokenizer, Qwen ChatML format, 4,096-token context, assistant-only loss, and document-isolated packing. A packed sequence may contain multiple conversations, but each conversation retains an independent attention context. We split every materialized dataset into 90\% training data and 10\% validation data, then train for two epochs. The tables below report assistant-supervised token counts before packing, so they exclude runtime-added EOD labels and padding.

We use \emph{UD-based} to denote an SFT model initialized from the uniform data ordering with learning-rate decay checkpoint. We use \emph{CMA-based} to denote an SFT model initialized from the curriculum model average checkpoint. Within each setting and repetition, the UD-based and CMA-based models use the same materialized examples, data order, tokenizer, optimizer, learning-rate schedule, and training budget. The pretraining initialization is the only difference within each paired comparison.

Each setting is repeated with three matched SFT data orders. These repetitions measure whether the direction of the UD--CMA difference is stable across changes in SFT data order. Because the CMA checkpoint jointly reflects curriculum ordering, late constant-LR training, and checkpoint averaging, the experiments compare the complete CMA recipe with UD rather than isolating one CMA component.

\paragraph{Training configuration.}

We continue to use the MuonH optimizer, as in pretraining~\citep{wen2026fantastic2}. The base learning rate follows a cosine schedule from $1\times 10^{-5}$ to $1\times 10^{-7}$. MuonH-managed matrix parameters use a $10\times$ multiplier for their Hyperball weight learning rate. The UD-based and CMA-based models use identical optimizer settings within every SFT setup.

The GSM8K-based experiment runs on one node with eight RTX~5090 GPUs. It uses tensor parallelism 1, pipeline parallelism 2, micro-batch size 1, global batch size 160, and sequence length 4,096. The Math\&Code and Tulu-3 settings use training durations and batch configurations adapted to their data budgets. These configurations remain matched between the UD-based and CMA-based models within each setting.

\paragraph{Evaluation protocol.}

Evaluation checkpoints are fixed before inspecting the benchmark results. The UD-based and CMA-based models are evaluated at the same training steps within each setting. The reported endpoint is step~172 for GSM8K-based SFT, step~2,431 for Math\&Code SFT with replay, and step~449 for Tulu-3 mixed-domain SFT.

For the two mathematics-oriented settings, we evaluate all 1,319 GSM8K test examples using zero-shot ChatML prompting and greedy generation. The answer extractor first truncates any continuation that begins a new user turn. It prioritizes explicitly marked final answers, then checks standard GSM8K and boxed-answer formats, and finally uses the last numeric token as a fallback. Exact numeric equivalence after normalization determines correctness.

This flexible extraction reduces score differences caused only by answer formatting. We also use an explicit-answer-only variant as a sensitivity check and compare the sets of questions solved by the two models. For correct-answer sets $C$ and $U$ from the CMA-based and UD-based models, respectively, we report
\[
\operatorname{IoU}(C,U)
=
\frac{|C\cap U|}{|C\cup U|}.
\]
We additionally report C-only and U-only counts, which measure questions solved exclusively by the CMA-based and UD-based models.

For Tulu-3 mixed-domain SFT, we evaluate the final models on the same 15-task benchmark suite used in the base-model scorecard. The suite includes MMLU-Pro and BBH, and its primary summary is an unweighted macro-average over the 15 task scores. IFEval is reported separately as an instruction-following diagnostic. All broad-evaluation numbers are averaged across the same three SFT repetitions.

\subsection{GSM8K-Based SFT}

\paragraph{Data and training budget.}

The GSM8K-based setting provides the most controlled test of whether the pretraining difference survives SFT. It emphasizes GSM-oriented mathematics data and does not include pretraining replay. The dataset contains 191,767 source conversations and 41.42M assistant-supervised tokens. After packing, it contains 15,292 training sequences and reaches its final evaluation point after 172 optimizer steps.

\begin{table}[htbp]
\centering
\small
\setlength{\tabcolsep}{4pt}
\renewcommand{\arraystretch}{1.08}
\caption{
Composition of the GSM8K-based SFT dataset. Token shares are computed from assistant-supervised tokens before packing. Small deviations from the target shares arise from indivisible records.
}
\label{tab:posttrain-focused-data}
\begin{tabular}{@{}
>{\raggedright\arraybackslash}p{0.17\linewidth}
>{\raggedright\arraybackslash}p{0.48\linewidth}
>{\raggedleft\arraybackslash}p{0.15\linewidth}
>{\raggedleft\arraybackslash}p{0.11\linewidth}
@{}}
\toprule
Component & Included source families & Supervised tokens & Token share \\
\midrule
GSM-oriented math
&
GSM8K train; Tulu grade-filtered math; Tulu OpenMath GSM; OpenMathInstruct augmented GSM8K; MetaMathQA GSM rephrased
&
37.28M
&
89.998\%
\\
Instruction
&
PersonaHub instruction following
&
2.07M
&
5.000\%
\\
General
&
WildChat; NoRobots
&
2.07M
&
5.001\%
\\
\midrule
Total & & 41.42M & 100.000\% \\
\bottomrule
\end{tabular}
\end{table}

\paragraph{Per-run results and answer analysis.}

The endpoint mean is 66.89\% for the UD-based models and 68.66\% for the CMA-based models. The resulting CMA advantage is 1.77 percentage points. All three endpoint repetitions favor CMA, with differences ranging from 1.36 to 2.12 points. The intermediate evaluations show the same overall direction.

Flexible extraction is intended to remove differences caused only by output format. Under this protocol, every endpoint repetition contains more C-only than U-only questions. The CMA-based models therefore solve more questions missed by their paired UD-based models. The explicit-answer-only sensitivity evaluation preserves the same mean difference, which supports the interpretation that the gain reflects stronger mathematical problem solving rather than answer formatting alone.

\begin{table}[htbp]
\centering
\small
\setlength{\tabcolsep}{4pt}
\renewcommand{\arraystretch}{1.08}
\caption{
Paired GSM8K results for the GSM8K-based SFT setting. UD and CMA are accuracies in percent, and $\Delta$ is CMA minus UD in percentage points. C-only and U-only count questions solved exclusively by the CMA-based and UD-based models.
}
\label{tab:focused-gsm-per-run}
\begin{tabular}{@{}rr rrr rrr@{}}
\toprule
Run & Step & UD & CMA & $\Delta$ & IoU & C-only & U-only \\
\midrule
1 & 172 & 67.32 & 69.14 & $+1.82$ & .754 & 138 & 114 \\
2 & 172 & 67.10 & 69.22 & $+2.12$ & .721 & 160 & 132 \\
3 & 172 & 66.26 & 67.63 & $+1.36$ & .730 & 147 & 129 \\
\midrule
Mean & 172 & 66.89 & 68.66 & $+1.77$ & -- & -- & -- \\
\bottomrule
\end{tabular}
\end{table}

\subsection{Math\&Code SFT with Replay}

\paragraph{Data and training budget.}

The Math\&Code setting tests whether the CMA advantage persists after a substantially longer SFT process on a broader data distribution. The mixture contains mathematics data, code data, and a small replay component from the final part of the Phase~2 curriculum. It contains 2,014,933 source conversations and 530.86M assistant-supervised tokens. After packing, it contains 194,530 sequences and reaches its final evaluation point after 2,431 optimizer steps.

The replay examples occupy 4.96\% of the source conversations but only 2.46\% of the assistant-supervised tokens. The dataset is a legacy materialized mixture whose replay slots were replaced in place. Since its original manifest did not preserve component-level token totals, we recomputed the values below with the same tokenizer and target mask used by the SFT loader. Its packing manifest also records 697 retained overlength rows.

\begin{table}[htbp]
\centering
\small
\setlength{\tabcolsep}{4pt}
\renewcommand{\arraystretch}{1.08}
\caption{
Composition of the Math\&Code SFT with replay dataset. Assistant-supervised token counts are recomputed from the materialized JSONL using the training tokenizer and target mask.
}
\label{tab:posttrain-scaled-data}
\begin{tabular}{@{}
>{\raggedright\arraybackslash}p{0.17\linewidth}
>{\raggedright\arraybackslash}p{0.48\linewidth}
>{\raggedleft\arraybackslash}p{0.15\linewidth}
>{\raggedleft\arraybackslash}p{0.11\linewidth}
@{}}
\toprule
Component & Included source families & Supervised tokens & Token share \\
\midrule
Mathematics
&
OpenMathInstruct math/GSM variants; MetaMathQA GSM/MATH variants; GSM8K train
&
430.97M
&
81.183\%
\\
Code
&
OpenCodeInstruct generic/algorithmic; Magicoder Evol-Instruct
&
86.85M
&
16.360\%
\\
Curriculum replay
&
Samples from the final Phase~2 curriculum parquet, formatted as SFT continuation data
&
13.04M
&
2.456\%
\\
\midrule
Total & & 530.86M & 100.000\% \\
\bottomrule
\end{tabular}
\end{table}

\paragraph{Per-run results and persistence after longer SFT.}

A possible explanation for the original base-model difference is that CMA receives higher-quality data near the end of pretraining. Under this explanation, its advantage could be temporary and disappear after sufficient SFT on a new distribution. The Math\&Code setting provides a stronger test of this possibility because it uses more training data, more optimizer steps, and a broader mixture than the GSM8K-based setting.

After this longer SFT process, the UD-based models reach 74.10\% mean GSM8K accuracy, while the CMA-based models reach 76.12\%. CMA is ahead in all three repetitions, and its mean advantage is 2.02 percentage points. The observed differences range from 1.21 to 3.26 points. The explicit-answer-only sensitivity evaluation gives the same qualitative and nearly identical quantitative conclusion.

These results show that the CMA advantage is not immediately overwritten by longer SFT on a broader mixture. Its mean size is also slightly larger than in the GSM8K-based setting. Since the two settings use different data mixtures and training budgets, their absolute GSM8K scores should not be compared directly. The relevant comparison is the paired difference between UD-based and CMA-based models within each setting.

\begin{table}[htbp]
\centering
\small
\setlength{\tabcolsep}{4pt}
\renewcommand{\arraystretch}{1.08}
\caption{
Paired endpoint GSM8K results for Math\&Code SFT with replay. UD and CMA are accuracies in percent, and $\Delta$ is CMA minus UD in percentage points.
}
\label{tab:scaled-gsm-per-run}
\begin{tabular}{@{}r rrr rrr@{}}
\toprule
Run & UD & CMA & $\Delta$ & IoU & C-only & U-only \\
\midrule
1 & 73.24 & 76.50 & $+3.26$ & .792 & 136 & 93 \\
2 & 75.06 & 76.27 & $+1.21$ & .798 & 120 & 104 \\
3 & 74.00 & 75.59 & $+1.59$ & .786 & 129 & 108 \\
\midrule
Mean & 74.10 & 76.12 & $+2.02$ & -- & -- & -- \\
\bottomrule
\end{tabular}
\end{table}

\subsection{Tulu-3 Mixed-Domain SFT}

\paragraph{Data and training budget.}

The first two settings mainly test mathematical capability. The Tulu-3 mixed-domain setting examines whether the UD--CMA difference remains after SFT on a broader instruction distribution. Its mixture covers general data, mathematics, code, instruction following, and safety. It contains 264,612 source conversations and 82.25M assistant-supervised tokens.

After packing, the dataset contains 39,914 sequences. Both pretraining initializations are trained for two epochs and evaluated at the predefined final step~449. This setup uses the same three matched SFT repetitions as the mathematics experiments.

\begin{table}[htbp]
\centering
\small
\setlength{\tabcolsep}{4pt}
\renewcommand{\arraystretch}{1.08}
\caption{
Composition of the Tulu-3 mixed-domain SFT dataset. Category shares are defined by assistant-supervised tokens before packing.
}
\label{tab:posttrain-broad-data}
\begin{tabular}{@{}
>{\raggedright\arraybackslash}p{0.17\linewidth}
>{\raggedright\arraybackslash}p{0.48\linewidth}
>{\raggedleft\arraybackslash}p{0.15\linewidth}
>{\raggedleft\arraybackslash}p{0.11\linewidth}
@{}}
\toprule
Component & Included source families & Supervised tokens & Token share \\
\midrule
General
&
FLAN; WildChat; NoRobots; OASST; Aya; TableGPT; SciRIFF
&
37.01M
&
45.001\%
\\
Mathematics
&
Persona Math; grade-filtered math; OpenMath GSM8K; NuminaMath; intermediate algebra
&
20.56M
&
24.999\%
\\
Code
&
Evol CodeAlpaca HumanEval; PersonaHub code
&
8.22M
&
10.000\%
\\
Instruction
&
PersonaHub instruction following
&
8.22M
&
10.000\%
\\
Safety
&
WildGuardMix; WildJailbreak; CoCoNot
&
8.22M
&
10.000\%
\\
\midrule
Total & & 82.25M & 100.000\% \\
\bottomrule
\end{tabular}
\end{table}

\paragraph{Broad capability and instruction-following results.}

Across the 15-task benchmark suite, the UD-based models reach a macro-average of 53.96\%, while the CMA-based models reach 55.13\%. The difference is 1.17 percentage points. CMA is higher on 10 of the 15 benchmarks, which shows that its aggregate advantage extends beyond mathematics. The gains are nevertheless uneven across individual tasks.

IFEval is reported separately because it directly measures instruction following rather than the broader capability average. The UD-based models reach 41.71\%, while the CMA-based models reach 43.07\%. This gives CMA an additional advantage of 1.36 percentage points on instruction following. The direction is consistent with the higher 15-task aggregate, although neither result implies that every capability improves.

\begin{table}[htbp]
\centering
\small
\setlength{\tabcolsep}{5pt}
\renewcommand{\arraystretch}{1.08}
\caption{
Tulu-3 mixed-domain SFT results averaged across three repetitions. UD and CMA are percentages, and $\Delta$ is CMA minus UD in percentage points. The 15-task macro-average excludes IFEval, which is reported separately as an instruction-following diagnostic.
}
\label{tab:posttrain-broad-details}
\begin{tabular}{@{}lrrr@{}}
\toprule
Benchmark & UD & CMA & $\Delta$ \\
\midrule
GSM8K & 59.92 & 59.23 & $-0.68$ \\
MATH & 22.39 & 26.18 & $+3.79$ \\
HumanEval & 42.27 & 40.65 & $-1.63$ \\
MBPP & 50.97 & 50.84 & $-0.13$ \\
MMLU & 53.97 & 55.80 & $+1.83$ \\
MMLU-Pro & 25.73 & 26.11 & $+0.38$ \\
ARC-C & 63.28 & 67.23 & $+3.95$ \\
ARC-E & 80.90 & 83.60 & $+2.70$ \\
BoolQ & 74.18 & 69.57 & $-4.61$ \\
CommonsenseQA & 66.09 & 68.69 & $+2.59$ \\
HellaSwag & 43.69 & 50.94 & $+7.25$ \\
PIQA & 68.79 & 69.01 & $+0.22$ \\
SIQA & 61.05 & 63.19 & $+2.13$ \\
WinoGrande & 51.62 & 52.46 & $+0.84$ \\
BBH & 44.58 & 43.52 & $-1.06$ \\
\midrule
15-task macro-average & 53.96 & 55.13 & $+1.17$ \\
\midrule
IFEval & 41.71 & 43.07 & $+1.36$ \\
\bottomrule
\end{tabular}
\end{table}

\subsection{Summary of the Three Comparisons}

The three settings answer complementary questions about post-training persistence. GSM8K-based SFT shows that the CMA-based models retain a mathematics advantage without pretraining replay. Math\&Code SFT with replay shows that this advantage remains after a longer SFT process on a broader data distribution. Tulu-3 mixed-domain SFT further shows a higher aggregate score and stronger instruction following beyond the targeted mathematics settings.

The improvement is not uniform across all tasks. Some benchmarks favor the UD-based models, even though the CMA-based models perform better on the overall mixed-domain average. The current experiments therefore support the complete CMA recipe rather than a claim of universal improvement. This task-level variation also motivates more controlled curriculum design and component-wise ablations in future work.

\section[LR Schedule Analysis]{Learning Rate Schedule Diagnostics}
\label{app:lr-schedule-analysis}
We provide more detailed diagnostics and fitting results for learning rate schedule diagnostics in this section.

\subsection{Multi-Power Law for Effective Learning Rate Schedules}\label{app:muonh:mpl}

\paragraph{Preliminary.} The Multi-Power Law (MPL) was introduced to predict loss curves across
learning rate schedules from a small number of training runs~\citep{luo2025multipowerlawlosscurve}. We adapt the
same construction to an effective learning rate signal for one selected weight
group. Let $q_t$ denote either the scalar optimizer learning rate coefficient
$\eta_t$ or the induced effective LR $\rho_t$. We define its post-warmup
trapezoidal exposure as
\[
S_q(t)
=
\sum_{k=t_{\mathrm{warm}}+1}^{t}
\frac{q_{k-1}+q_k}{2}\Delta t_k,
\]
and let $S_w$ denote the corresponding exposure accumulated during warmup under
the same choice of $q$. The resulting model is
\begin{equation}
\begin{aligned}
    L(t)
    &=
    L_0
    +A\left[S_q(t)+S_w\right]^{-\alpha}
    +B\sum_{t_{\mathrm{warm}}<k\leq t}
        \Delta q_k\,G\!\left(x_k(t)\right),
    &
    \Delta q_k
    &=
    q_k-q_{k-1},
    \\
    x_k(t)
    &=
    Cq_k^{-\gamma}
    \left[S_q(t)-S_q(k-1)\right],
    &
    G(x)
    &=
    1-(1+x)^{-\beta}.
\end{aligned}
\label{eq:effective-mpl}
\end{equation}

The intuition is to decompose loss evolution into an exposure-controlled
baseline and a schedule-shape correction. The first power-law term is
Chinchilla-like: it treats the cumulative $q$-exposure $S_q(t)+S_w$ as an
effective optimization-time coordinate and captures the training progress expected from that accumulated exposure. This term alone,
however, does not distinguish schedules that accumulate similar exposure
through different decay patterns.

The summation term accounts for the additional loss reduction induced by decreasing the learning rate signal. For a decaying schedule,
$\Delta q_k<0$. A decrease at step $k$ therefore contributes a negative correction
\[
B\Delta q_k G\!\left(x_k(t)\right)
=
-B\left(q_{k-1}-q_k\right)G\!\left(x_k(t)\right)
\]
to the loss at a later step $t$. This benefit is not assumed to appear
instantaneously. Instead, $x_k(t)$ measures the exposure accumulated after the
change, rescaled by the post-change signal level $q_k$, while
$G(x_k(t))$ represents the fraction of the eventual loss reduction that has
been realized by time $t$. As subsequent exposure accumulates, $G(x)$ approaches
one, with the unrealized fraction $1-G(x)=(1+x)^{-\beta}$ decaying according to
a power law. Finally, we approximate the total schedule-shape correction by
superposing the delayed responses induced by all preceding changes in $q$.
Thus, MPL combines one power law for cumulative training progress with a
collection of power-law responses to reductions in the effective
learning rate signal.

A useful analytical simplification assumes that every reduction is fully
realized as soon as it occurs, corresponding to $G(x)=1$. The response-weighted
sum then telescopes. With
$q_{\mathrm{warm}}=q_{t_{\mathrm{warm}}}$, we obtain
\begin{equation}
\begin{aligned}
    L_{G=1}(t)
    &=
    L_0
    +A\left[S_q(t)+S_w\right]^{-\alpha}
    +B\sum_{t_{\mathrm{warm}}<k\leq t}\Delta q_k
    \\
    &=
    L_0
    +A\left[S_q(t)+S_w\right]^{-\alpha}
    +B\left(q_t-q_{\mathrm{warm}}\right).
\end{aligned}
\label{eq:effective-mpl-g1}
\end{equation}
For a decaying schedule and $B>0$, the final term is non-positive and lowers
the predicted loss relative to the accumulated-exposure baseline. This
instantaneous-response approximation discards when the individual reductions
occurred: beyond their contribution to $S_q(t)$, the correction retains only
the net decrease from $q_{\mathrm{warm}}$ to $q_t$. We therefore use it as a
simplified interpretation of the full response model rather than as a claim
that learning rate reductions take effect instantaneously in actual training.

We test this transfer on the two BF16 ordinary-Muon runs already used in
\Cref{fig:hyperball-trace-alignment,fig:hyperball-diagnostics-2x2}.  For each
run, we fit the post-warmup validation curve using either $q_t=\eta_t$ or
$q_t=\rho_t$, with the final $20\%$ of validation targets held out.  Because
the two signals have different units, each is normalized by its own
post-warmup peak before fitting; this only rescales the fitted constants.  The
fit follows the three-stage MPL protocol: power-only, power plus a schedule-drop term, and the full response above.  As a simpler comparison, we also fit
the $G(x)=1$ model in~\Cref{eq:effective-mpl-g1}; unlike the formal response,
this simplification retains only the current change from the warmup endpoint.

\begin{figure*}[ht]
    \centering
    \includegraphics[width=0.98\linewidth]{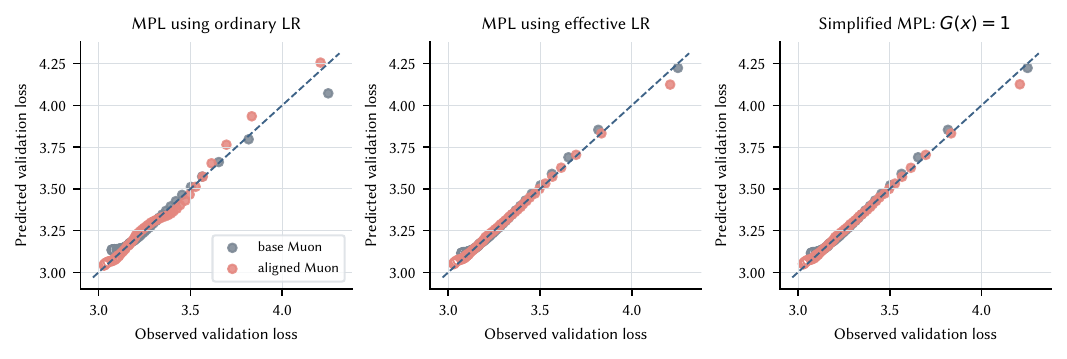}
    \caption{MPL transfer to an effective learning rate signal for the FC2 group~(MPL down-projection weight).  All panels compare predicted with observed validation loss; filled
    markers are fitting targets and hollow markers are the final $20\%$
    holdout.  \textbf{Left:} The formal MPL fit using the ordinary LR.  \textbf{Center:} The
    formal MPL fit using the induced effective LR.  \textbf{Right:} The simplified
    $G(x)=1$ model in~\Cref{eq:effective-mpl-g1}, also using the induced effective LR.  The two signals are normalized separately, and colors
    distinguish base and effective-LR-aligned Muon.}
    \label{fig:effective-lr-mpl-fc2}
\end{figure*}

The effective LR coordinate gives a lower mean held-out RMSE in this
small FC2 diagnostic ($0.0210$ versus $0.0265$ for ordinary LR), with the
largest gain for base Muon ($0.0270$ versus $0.0422$).
For aligned Muon, the
two representations are comparably predictive ($0.0149$ and $0.0108$).
The simplified
$G(x)=1$ effective LR fit has a mean held-out RMSE of $0.0207$, with
$0.0270$ for base Muon and $0.0144$ for aligned Muon, matching the vanilla MPL of effective LR on these two curves.

Then, from the perspective of \Cref{eq:effective-mpl-g1}, the loss decreasing is strongly related to the effective LR decreasing. And as shown in \Cref{fig:hyperball-trace-alignment}, the ordinary Muon run results in an LR schedule decay aggressively in the early phase, while the MuonH and aligned Muon runs use linear LR decay. The early effective LR decay in the ordinary Muon run results in the loss of potential to decay further near the end of training. Thus, the MuonH and aligned Muon runs finally surpass the ordinary run, which is not by coincidence from this diagnostic.

\subsection{WSD Sweeps and Limited-Compute Schedule Estimation}

\paragraph{The controlled sweeps isolate peak, horizon, and terminal-decay length.}
The source runs use a $0.6$B decoder-only model, BF16 training, sequence length 4,096, GBS 512, MuonH multiplier 3, linear WSD decay, and one seed per configuration.  TPP 20, 50, and 100 correspond to $11.32$, $28.31$, and $56.62$B tokens.  At TPP 20, all five positive decay ratios $\{0.2,0.4,0.6,0.8,1.0\}$ are complete for effective peaks $\{0.008,0.012,0.016,0.020,0.024\}$.  Peaks $0.008$ and $0.012$ additionally cover TPP 50 and 100.
Final loss is taken from the run summary at the nominal target step, which avoids replacing the endpoint with the last observation of a sparsely sampled history.  GBS is fixed, but the two lower-peak TPP20 families use micro-batch size 2 whereas peaks $0.016$--$0.024$ use micro-batch size 4; the peak trend is therefore a practical design diagnostic rather than a perfectly isolated micro-batch ablation.

\paragraph{Formal MPL turns a few anchor schedules into a cross-schedule trend estimate.}
For the WSD experiments below, $q_t$ is the post-warmup effective LR and the
formal response is the one in~\Cref{eq:effective-mpl}.  Only post-warmup
validation targets are fitted.  The three-stage procedure first fits
$(L_0,A,\alpha)$ on the first $25\%$ of each anchor curve, then adds the
instantaneous LR-drop coefficient $B$ on the complete curves, and finally
refines all formal-MPL parameters.  Six initializations are compared; each
curve contributes at most 16 fitting targets, while the final $10\%$ receives
weight 4.  All reported WSD fits converged in all three stages.

\begin{figure*}[ht]
\centering
\includegraphics[width=0.98\linewidth]{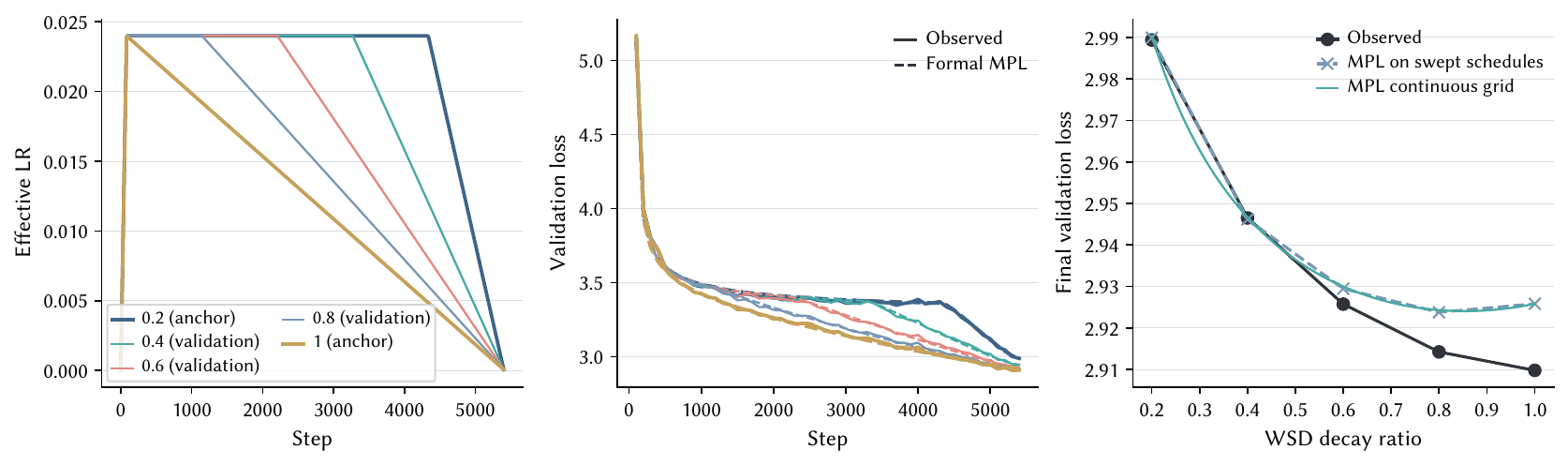}
\caption{\textbf{A two-anchor MPL example at effective peak $0.024$ and TPP 20.}  Only decay ratios $0.2$ and $1.0$ are fitted; ratios $0.4$, $0.6$, and $0.8$ are unseen validation schedules.  \textbf{Left:} WSD schedules.  \textbf{Center:} Held-out loss curves, with observations shown as solid lines and MPL predictions as dashed lines.  \textbf{Right:} Final-loss ranking over the swept and continuous decay grids.  The two anchors place the estimated minimum near ratio $0.85$, consistent with the long-decay end of the direct sweep.}
\label{fig:app-wsd-two-anchor-example}
\end{figure*}

\paragraph{A two-anchor protocol recovers the peak-to-decay trend at fixed horizon.}
Each fixed-peak TPP20 fit sees only ratios $0.2$ and $1.0$, leaving $0.4$, $0.6$, and $0.8$ unseen.  Across these 15 unseen schedules, mean curve RMSE is $0.0157$ and endpoint MAE is $0.0067$.  The predicted continuous-grid ratio rises as $0.33$, $0.57$, $0.67$, $0.75$, and $0.85$ over peaks $0.008$ through $0.024$.  This agrees with the monotone rise in the lower edge of the directly observed competitive region, although the individual observed grid minima remain noisy.

\begin{figure*}[ht]
\centering
\includegraphics[width=0.98\linewidth]{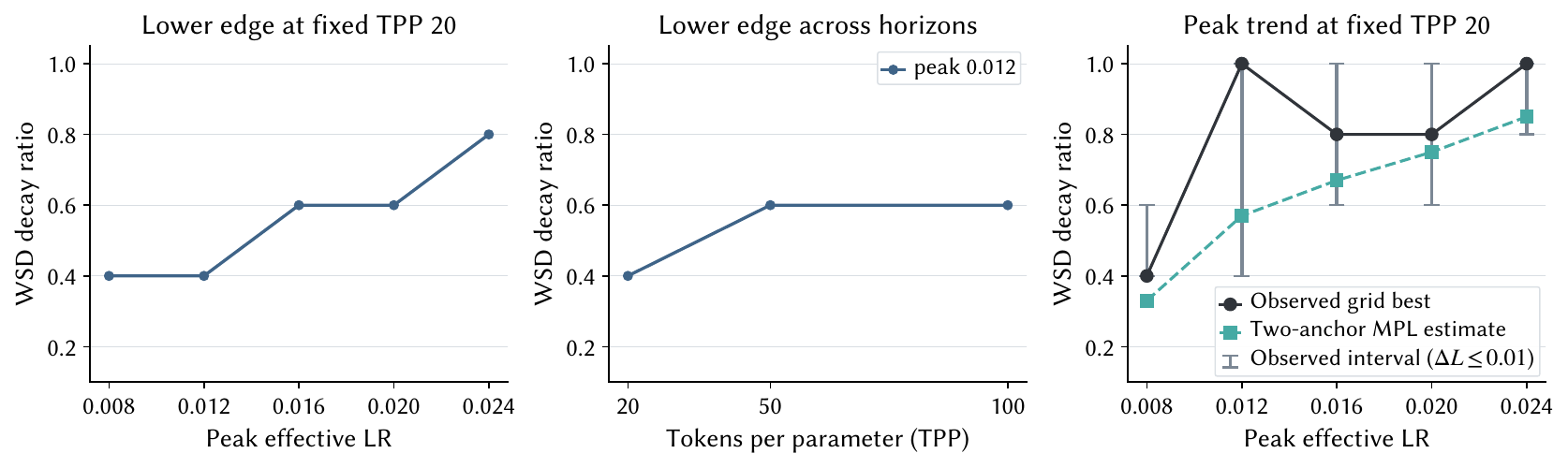}
\caption{WSD decay trends and the limited-compute diagnostic.  \textbf{Left:} At fixed TPP 20, the lower edge of the region within $0.01$ validation loss of the minimum rises nondecreasingly with effective peak LR.  \textbf{Center:} The same lower edge across horizons for peak $0.012$.  \textbf{Right:} Black points are the best ratios on the observed TPP20 grid, gray capped intervals contain every ratio within $0.01$ loss of the minimum, and green squares are continuous-ratio estimates from MPL fitted only to ratios $0.2$ and $1.0$.  The intervals describe competitive configurations, not uncertainty across seeds.}
\label{fig:app-wsd-limited-compute}
\end{figure*}

\paragraph{The estimator is a screening tool rather than an optimum certificate.}
The diagnostic helps choose which peak-rate and decay settings to test next. Broad plateaus, one seed per configuration, and noisy minima limit the precision of any inferred optimum.
The continual-training role of the Phase~1 power schedule is likewise a design property of its open-ended tail; these WSD sweeps support the long-decay features, not the exact power exponent.

\begin{figure}[ht]
    \centering
    \begin{subfigure}[t]{0.31\linewidth}
        \centering
        \includegraphics[width=\linewidth]{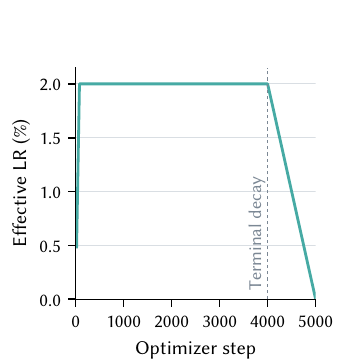}
        \caption{Predefined effective LR}
    \end{subfigure}\hfill
    \begin{subfigure}[t]{0.34\linewidth}
        \centering
        \includegraphics[width=\linewidth]{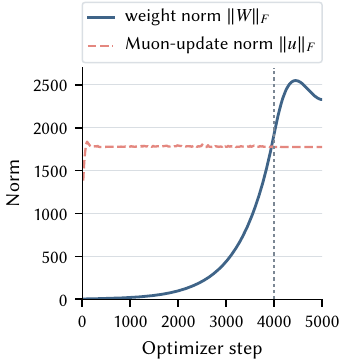}
        \caption{Weight and Muon-update norms}
    \end{subfigure}\hfill
    \begin{subfigure}[t]{0.31\linewidth}
        \centering
        \includegraphics[width=\linewidth]{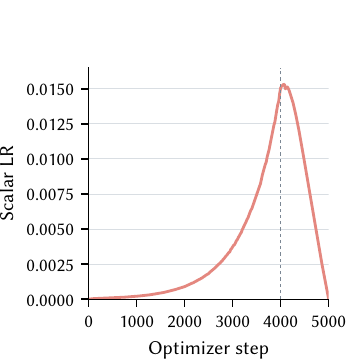}
        \caption{Scalar LR for ordinary Muon}
    \end{subfigure}
    \caption{Indirect effective LR control in an ordinary-Muon WSD diagnostic, shown on linear axes.  All quantities refer to MLP down-projection matrices.  \textbf{(a)} The predefined effective LR holds near $2\%$ after warmup.  \textbf{(b)} The weight norm grows while the orthogonalized Muon-update norm stays roughly constant.  \textbf{(c)} The scalar LR must therefore rise to maintain the effective LR before falling during terminal decay.}
    \label{fig:hyperball-hill}
\end{figure}

\subsection{A Prescribed Effective LR Can Induce a Hill-Like Scalar LR in Ordinary Muon}
\label{app:hill-like}
The warmup--stable--decay (WSD) diagnostic in~\Cref{fig:hyperball-hill} ramps the effective LR to $0.02$, holds it there, and begins terminal decay at step 4,000.  During the stable interval, the weight norm of the MLP down-projection matrices grows while the orthogonalized Muon-update norm remains roughly constant.  Ordinary Muon must therefore increase its scalar LR to maintain the predefined effective LR, before reducing it during terminal decay.  The run remains numerically stable and reaches validation loss $3.0678$ after 5,000 steps.  This constructed example shows why a scalar LR curve cannot be interpreted without its induced effective LR schedule; it demonstrates feasibility, not optimality or a universal requirement for ordinary Muon.

\section[Scaling Checkpoint Ledger]{Puro-2B Scaling Checkpoint Ledger}
\label{app:puro-scaling-ledger}

The following ledger freezes the identifiers and accounting coordinates used by
Figure~\ref{fig:cost-saving-factors}(b).  Phase~1 is shared by all rows.  Phase~2
tokens are schedule consumption, so the production Phase~2 pool is not added a
second time when the early transition range is listed.
The endpoint comparison is visualized in~\Cref{fig:curriculum-endpoint-scores,fig:curriculum-model-average}.

\begin{landscape}
\pagestyle{plain}
\begin{table}[ht]
\centering
\scriptsize
\caption{Checkpoint lineage for the five UD points in the Puro Cost
Scaling Law, the CD endpoint, and the canonical CMA endpoint.
GPU-hours are total active-training GPU-hours and costs use the normalized RTX
5090 rate and are reported in USD.  Here, \emph{SMA6} (simple moving average)
denotes the equal-weight average of model parameters from six late checkpoints
on the step-218,000 constant-LR continuation; optimizer states are not averaged
(details in~\Cref{app:curriculum:averaging}).}
\label{tab:puro-scaling-checkpoints}
\resizebox{\linewidth}{!}{%
\begin{tabular}{l r r l c r r}
\toprule
Released checkpoint ID & Phase~2 tokens & Total tokens & Phase~2 ordering & Model averaging & Total GPU-h & Cost (USD) \\
\midrule
\texttt{Puro-2B-UD-Phase2-1of16} & 60.00B & 498.83B & UD global reshuffle & No & 7,041 & \$2.16K \\
\texttt{Puro-2B-UD-Phase2-1of8} & 120.00B & 558.83B & UD global reshuffle & No & 8,073 & \$2.47K \\
\texttt{Puro-2B-UD-Phase2-1of4} & 240.00B & 678.83B & UD global reshuffle & No & 10,136 & \$3.10K \\
\texttt{Puro-2B-UD-Phase2-1of2} & 480.00B & 918.83B & UD global reshuffle & No & 14,262 & \$4.37K \\
\texttt{Puro-2B-UD} & 959.99B & 1.40T & UD global reshuffle & No & 22,514 & \$6.89K \\
\texttt{Puro-2B-CD} & 959.99B & 1.40T & CD component-local curriculum; decay final & No & 22,514 & \$6.89K \\
\texttt{Puro-2B-CMA} & 959.99B & 1.40T & CMA curriculum + late constant LR & Yes (SMA6) & 22,514 & \$6.89K \\
\bottomrule
\end{tabular}%
}
\end{table}

\vfill

\begin{table}[ht]
\centering
\scriptsize
\setlength{\tabcolsep}{2.4pt}
\caption{15-benchmark score sheet for the UD, CD, CDC, and CMA branches(detailed definitions in~\Cref{sec:recipe:curriculum}).
Scores are percentages, and Avg15 is the unweighted arithmetic mean over the
15 displayed benchmarks. The canonical Puro-2B row(CMA 218k avg) is the released
six-checkpoint SMA endpoint.
}
\label{tab:puro-scaling-evaluations}
\resizebox{\linewidth}{!}{%
\begin{tabular}{l*{16}{r}}
\toprule
Model & GSM8K & MATH & MBPP & HumanEval & MMLU & MMLU-Pro & ARC-C & ARC-E & BoolQ & CSQA & HSwag & PIQA & SIQA & WinoGrande & BBH & Avg15 \\
\midrule
UD 1/16 & 47.23 & 21.78 & 45.53 & 18.90 & 49.71 & 20.67 & 61.02 & 79.19 & 75.11 & 65.68 & 62.28 & 75.30 & 60.18 & 59.51 & 37.66 & 51.98 \\
UD 1/8 & 48.22 & 24.64 & 46.69 & 25.00 & 51.37 & 22.29 & 63.39 & 79.54 & 77.16 & 64.78 & 61.25 & 75.90 & 60.24 & 59.98 & 38.09 & 53.24 \\
UD 1/4 & 42.30 & 26.02 & 48.64 & 23.78 & 53.96 & 23.83 & 66.44 & 81.83 & 74.83 & 65.27 & 62.59 & 76.17 & 63.10 & 60.22 & 41.20 & 54.01 \\
UD 1/2 (\$4.4K) & 56.03 & 27.86 & 49.81 & 20.73 & 54.82 & 24.02 & 66.78 & 84.66 & 77.34 & 67.65 & 63.62 & 75.84 & 62.90 & 59.91 & 41.14 & 55.54 \\
UD decay avg & 55.72 & 26.70 & 50.58 & 18.90 & 56.23 & 24.87 & 68.81 & 84.13 & 79.27 & 67.16 & 63.32 & 77.69 & 62.64 & 60.30 & 37.29 & 55.57 \\
UD full & 55.72 & 26.94 & 50.19 & 20.73 & 56.22 & 25.55 & 70.17 & 84.83 & 79.48 & 67.49 & 63.30 & 77.58 & 62.64 & 60.38 & 38.75 & 56.00 \\
CD decay final & 57.77 & 29.68 & 52.14 & 25.00 & 57.11 & 25.56 & 73.56 & 85.89 & 76.24 & 68.88 & 63.38 & 75.46 & 64.43 & 60.93 & 41.45 & 57.17 \\
CD decay avg & 57.32 & 29.30 & 51.75 & 26.22 & 57.12 & 26.50 & 73.56 & 86.07 & 76.18 & 69.29 & 62.96 & 75.63 & 64.33 & 60.69 & 40.80 & 57.18 \\
CDC 215k final & 54.74 & 27.60 & 46.30 & 22.56 & 56.28 & 25.44 & 73.90 & 84.83 & 73.52 & 67.49 & 62.46 & 74.92 & 62.85 & 61.33 & 40.39 & 55.64 \\
CMA 215k avg & 55.95 & 28.18 & 50.19 & 25.00 & 57.05 & 26.09 & 75.93 & 86.07 & 74.80 & 67.73 & 62.72 & 75.41 & 62.69 & 61.48 & 42.78 & 56.80 \\
CDC 218k final & 57.47 & 29.22 & 50.19 & 30.49 & 56.45 & 25.12 & 73.90 & 84.13 & 78.10 & 67.49 & 63.02 & 75.08 & 63.51 & 61.48 & 41.19 & 57.12 \\
CMA 218k avg (\$6.9K) & 59.67 & 30.30 & 52.92 & 31.10 & 57.44 & 25.81 & 74.24 & 86.07 & 76.85 & 67.81 & 63.21 & 75.52 & 63.61 & 60.62 & 42.05 & 57.81 \\
\bottomrule
\end{tabular}%
}
\end{table}
\end{landscape}
\pagestyle{fancy}

\section[Curriculum and CMA]{Curriculum Construction and Model Averaging}
\label{app:curriculum-details}

This appendix specifies our concrete Phase~2 curriculum and model-averaging
configuration. The curriculum first assigns each example a component-local
token-percentile coordinate under the configured preference order of its own
component. Scores are therefore used only to order examples within a component;
we do not assume that scores are comparable across components. Equal percentile
ranges from all components are then aligned into production curriculum buckets,
which approximately preserves the configured component mixture while allowing
each scored component to progress through its own preference order.

The Phase~2 schedule begins with a transition that combines deterministic
Phase~1 replay with the earliest ranges of the Phase~2 curriculum. The released
endpoint subsequently uses a constant learning rate continuation and an
equal-weight parameter average over six late checkpoints. We first describe the
scalable construction of the curriculum coordinates and buckets, then specify
the transition and UD control, and finally document the continuation,
averaging rule, and scope of the available comparisons.

Throughout this appendix, UD denotes the uniform-data ordering with learning rate decay, CD denotes the component-local curriculum ordering with the same decay, CDC denotes curriculum followed by constant-LR continuation without averaging, and CMA denotes the production continuation with six-checkpoint averaging.

\subsection{Scalable Construction of Component-Local Curriculum Buckets}
\label{app:curriculum:buckets}

For each component, the curriculum construction must determine where an example
lies within that component's ordered token range. An exact implementation could
sort every document by
\[
    (\text{configured score},\text{stable hash}),
\]
compute cumulative token counts over the sorted rows, and divide the resulting
token range into 376 equal-percentile intervals. Applying document-level
distributed sorts and cumulative windows to the full Phase~2 pool would,
however, incur substantial Spark shuffle and serialization overhead. We
therefore approximate the same component-local token coordinate using a smaller
collection of aggregated internal bins.

For a component with a usable score, preprocessing first applies the configured
score direction so that the resulting sequence runs from the less-preferred to
the more-preferred score range. Spark's approximate-quantile operation then
estimates 256 score intervals with relative error \(0.01\). Each interval is
subdivided into 64 deterministic hash shards. The score intervals determine the
coarse preference order, while the hash shards provide deterministic ordering
and finer token-accounting granularity among examples with equal or nearly
equal scores. Thus, a scored component contains up to 256 score-ordered strata,
each with 64 deterministic subdivisions, although repeated score values can
produce fewer distinct strata. Null scores are placed at the configured
least-preferred end.

A component without a usable score instead uses 4096 deterministic random
shards. These shards define a reproducible component-local order, but we do not
interpret that order as a quality or difficulty ranking.

Preprocessing next aggregates the token count of every internal bin. For
internal bin \(j\) in component \(c\), let \(T_{c,j}\) be its token count and
let
\[
    T_c = \sum_j T_{c,j}
\]
be the total number of tokens in the component. The token fraction preceding
bin \(j\) is
\begin{equation}
    p_{c,j}
    =
    \frac{\sum_{i<j} T_{c,i}}{T_c}.
    \label{eq:curriculum-bin-prefix}
\end{equation}

The implementation additionally uses a deterministic within-bin position to
spread examples over the token interval occupied by that bin. For exposition,
let \(u_{c,j}(x)\in[0,1)\) denote the resulting fractional position of example
\(x\) within bin \(j\). Its component-local token coordinate can then be written
as
\begin{equation}
    p(x)
    =
    \frac{
        \sum_{i<j(x)}T_{c,i}
        +
        u_{c,j(x)}(x)T_{c,j(x)}
    }{T_c}.
    \label{eq:curriculum-example-position}
\end{equation}
Thus, \(p(x)=0.25\) means that \(x\) appears after approximately \(25\%\) of
the token mass of its own component, irrespective of how many documents
produce those tokens. Using token mass rather than document count prevents a
short document and a very long document from contributing equally to the
curriculum coordinate.

Each example is mapped to one of 376 production curriculum buckets:
\begin{equation}
    b(x)
    =
    \min\left\{
        375,\,
        \left\lfloor 376\,p(x)\right\rfloor
    \right\}.
    \label{eq:curriculum-production-bucket}
\end{equation}
Equivalently, bucket \(k\) receives from every component the examples whose
component-local token coordinates lie approximately in
\[
    \left[
        \frac{k}{376},
        \frac{k+1}{376}
    \right),
    \qquad
    k=0,\ldots,375.
\]

Consequently, each production bucket draws approximately the same token
fraction from every component. This alignment approximately preserves the
configured component mixture across the curriculum while allowing different
components to follow their own source-local preference orders. In particular,
the construction does not impose a single global quality ranking over examples
from different sources.

\subsection{Phase Transition and UD Control}
\label{app:curriculum:transition}

The Phase~2 schedule begins with a transition from the Phase~1 mixture to the
ordered Phase~2 pool. The materialized transition contains approximately
\(43.9\)B tokens: \(21.9\)B tokens of deterministic Phase~1 replay and
\(21.9\)B tokens from the earliest \(2.34\%\) of the ordered Phase~2 range.
The latter is a subset of the frozen Phase~2 component pool rather than an
additional copy of Phase~2 data. Therefore, only the Phase~1 replay is added
to the component-pool total when computing the complete production schedule.

\begin{table}[ht]
    \centering
    \small
    \caption{
        Token ledger for the Phase~2 transition and production schedule.
        The Phase~2 early range is already included in the component pool;
        the production total adds only the Phase~1 replay.
    }
    \label{tab:transition-token-ledger}
    \begin{tabular}{l r p{0.42\linewidth}}
        \toprule
        Stream or quantity & Tokens & Definition \\
        \midrule
        Phase~1 replay
        & 21.93B
        & Deterministic stable-hash sample from the Phase~1 mixture. \\

        Phase~2 early range
        & 21.93B
        & Earliest approximately \(2.34\%\) of the ordered Phase~2 pool. \\

        Realized transition
        & 43.87B
        & Phase~1 replay combined with the Phase~2 early range. \\

        Remaining Phase~2 range
        & 916.13B
        & Portion of the Phase~2 pool following the early transition range. \\

        Phase~2 component pool
        & 938.06B
        & Early range plus the remaining Phase~2 range. \\

        Production Phase~2 schedule
        & 959.99B
        & Phase~2 component pool plus Phase~1 replay. \\
        \bottomrule
    \end{tabular}
\end{table}

The aligned bucket construction approximately preserves the groups recorded in
the frozen preprocessing manifest throughout the non-transition portion of
Phase~2. These scheduler diagnostics use the manifest taxonomy rather than the
semantic-domain taxonomy used in the main data tables. The two taxonomies
differ for the code-only MegaMath-Code partition: the frozen preprocessing
manifest retains its original Mathematics label, whereas
\Cref{tab:data-recipe-summary,tab:data-components-detail} report that partition
under Code.

\begin{table}[ht]
    \centering
    \caption{
        Observed manifest-group token-share ranges across non-transition
        Phase~2 curriculum buckets.
    }
    \label{tab:curriculum-bucket-domain-ranges}
    \begin{tabular}{lc}
        \toprule
        Manifest group & Observed token-share range \\
        \midrule
        Chinese               & \(9.25\% \text{--} 9.54\%\)   \\
        Code                  & \(6.37\% \text{--} 8.73\%\)   \\
        English               & \(58.32\% \text{--} 59.79\%\) \\
        Mathematics           & \(22.38\% \text{--} 22.98\%\) \\
        Instruction-formatted & \(1.30\% \text{--} 1.40\%\)   \\
        \bottomrule
    \end{tabular}
\end{table}

Examples in the transition portion are deterministically shuffled with seed
\texttt{20260601}, and the subsequent curriculum buckets use seed
\texttt{20260602}. The UD stream is constructed from the same
materialized Phase~2 component pool, but removes the aligned curriculum order
through a deterministic global reshuffle with seed \texttt{20260609}. Thus, the
CD and UD streams match the Phase~2 token multiset at the pool
level while differing in its presentation order. The resulting streams are the
UD and CD inputs used in the endpoint comparison.

\subsection{Constant-LR Continuation and Checkpoint Averaging}
\label{app:curriculum:averaging}

The available late-stage configurations include averaging without launching a
new continuation, a constant-LR continuation resumed from step \(215{,}000\),
and a constant-LR continuation resumed from step \(218{,}000\). The production
export uses the branch resumed from step \(218{,}000\). These configurations
were not constructed as a matched ablation over continuation entry points.
Their comparison therefore motivates a production configuration choice but
does not establish that step \(218{,}000\) is an optimal resume point.

The CDC controls use the same curriculum order and these constant-LR continuations but report the unaveraged final checkpoints. CMA denotes the step-218,000 continuation with equal-weight averaging of six late checkpoints.

At step \(218{,}000\), the logged base learning rate coefficient is
\(4.08\times10^{-5}\). The MuonH matrix-parameter group applies a multiplier
of 10, yielding a group-level coefficient of \(4.08\times10^{-4}\). The
continuation holds these coefficients fixed instead of following the remaining
terminal decay. For a MuonH-wrapped matrix, the group-level coefficient is the
prescribed effective LR defined in \Cref{sec:recipe:hyperball}.

The production model equally averages the model parameters stored at the
following six optimizer steps:
\[
\begin{gathered}
    222{,}100,\quad
    222{,}200,\quad
    222{,}300,\\
    222{,}400,\quad
    222{,}500,\quad
    222{,}569.
\end{gathered}
\]
The averaging window spans 469 optimizer steps, corresponding to approximately
\(2.95\)B training tokens under the production batch configuration. Most
neighboring checkpoints are separated by 100 optimizer steps, or approximately
\(0.63\)B tokens, while the final interval contains 69 steps.

Only model parameters are averaged; optimizer
states are not. The released model therefore contains the equal-weight
arithmetic mean of the six checkpoints above rather than the parameters of the
unaveraged final iterate. Our implementation uses this simple model average,
whereas the reference CMA method evaluates several averaging rules and uses an
exponential moving average over the final six checkpoints as its default
configuration~\citep{luo2025learningratedecaywastes}.

\begin{landscape}
\pagestyle{plain}
\section[Data Recipe]{Data Recipe Details}
\label{app:data-recipe}
\begin{longtblr}[
  caption = {Dataset-family accounting for the stationary Phase 1 and Phase 2 component pools.},
  label = {tab:data-components-detail},
  note{a} = {Token counts are tokenizer-dependent materialized exposure, not necessarily unique upstream content; a family can aggregate overlapping configurations. They exclude the Phase 2 transition and replay. The code-only MegaMath-Code partition is reported under Code although its raw scheduling manifest used the broader math label. Terms describe current public metadata unless otherwise noted.},
]{
  colspec = {X[1.35,l] c r r X[1.25,l] X[1.7,l] X[2.25,l]},
  width = \linewidth,
  rowhead = 1,
  rows = {font=\fontsize{8.2}{9.4}\selectfont, valign=m, rowsep=0.5pt},
  row{1} = {font=\bfseries\fontsize{8.2}{9.4}\selectfont, bg=gray!10},
  hline{1,Z} = {1pt},
  hline{2} = {0.5pt},
}
Dataset / family & Domain & P1 tokens & P2 tokens & Materialization & Upstream source & Declared upstream terms\TblrNote{a} \\
FineWeb-Edu-CN & Chinese & 19.73B & 34.79B & Both: Score-filtered & opencsg/Fineweb-Edu-Chinese-V2.1~\cite{yu2025opencsgchinesecorpusseries} & OpenCSG Community License and Apache 2.0 \\
FineWeb-Edu-Chinese V2.2 & Chinese & 16.50B & 28.11B & P1: Score-filtered; P2: No additional sampling & opencsg/Fineweb-Edu-Chinese-V2.2~\cite{yu2025opencsgchinesecorpusseries} & Apache 2.0 metadata; web-source terms also apply \\
ChineseWebText2.0 (quality-filtered subset) & Chinese & 11.05B & 17.50B & Both: Score-filtered & CASIA-LM/ChineseWebText2.0~\cite{zhang2024chinesewebtext20largescalehighquality} & Apache 2.0 metadata \\
Deduplicated merged Chinese web & Chinese & -- & 5.77B & P2: Score-filtered & CCI\{,2,3\}, SkyPile-150B, WanJuan1.0, IndustryCorpus\{,2\}, and WuDaoCorpus2.0~\cite{wang2024cci30hqlargescalechinesedataset,wei2023skywork,he2023wanjuan,he2024opendatalabempoweringgeneralartificial,industrycorpus2,ZHANG2021216,ZHANG202193} & Mixed upstream terms; see appendix license note \\
Baidu-Baike & Chinese & 1.19B & 1.19B & Both: No additional sampling & mohamedah/baidu\_baike & MIT metadata; upstream content rights are not clarified \\
FineWiki-CN & Chinese & 1.10B & 1.10B & Both: No additional sampling & HuggingFaceFW/finewiki~\cite{penedo2025finewiki} & CC BY-SA 4.0; older Wikipedia content may also be GFDL \\
UNDL ZH-EN Aligned & Chinese & 1.75B & -- & P1: No additional sampling & bot-yaya/undl\_zh2en\_aligned & MIT metadata; underlying UN document terms also apply \\
Alpaca-Zh & Chinese & 0.01B & 0.01B & Both: No additional sampling & hfl/alpaca\_zh\_51k~\cite{chinese-llama-alpaca} & Apache 2.0 \\
MegaMath-Code & Code & -- & 42.77B & P2: No additional sampling & LLM360/MegaMath~\cite{zhou2025megamath} & ODC-By 1.0 \\
Nemotron Synthetic Code & Code & 17.09B & 25.64B & Both: Random subsample & nvidia/Nemotron-Pretraining-Code-v1 (Synthetic-Code)~\cite{nemotron_nana_v2} & NVIDIA Data Agreement for Model Training; gated, no raw-data redistribution \\
Swallow-Code-v2 & Code & 14.12B & 9.30B & Both: Score-filtered & tokyotech-llm/swallow-code-v2~\cite{fujii2025rewritingpretrainingdataboosts} & Apache 2.0 and original code licenses \\
CoderForge Trajectories & Code & -- & 15.78B & P2: No additional sampling & togethercomputer/CoderForge-Preview~\cite{CoderForge2026} & No dataset-level license declared; per-row licenses are provided \\
StackExchange & Code & -- & 5.54B & P2: Random subsample & togethercomputer/RedPajama-Data-1T~\cite{together2023redpajama} & CC BY-SA 2.5/3.0/4.0 \\
Python Code Large & Code & -- & 4.37B & P2: No additional sampling & ajibawa-2023/Python-Code-Large & MIT metadata; upstream code provenance undocumented \\
Python-Edu & Code & 3.41B & -- & P1: No additional sampling & HuggingFaceTB/smollm-corpus~\cite{lozhkov2024starcoder,benallal2024smollmcorpus} & ODC-By 1.0 and original code licenses \\
Codeforces-CoTs & Code & -- & 2.93B & P2: No additional sampling & open-r1/codeforces-cots~\cite{penedo2025codeforces} & CC BY 4.0 metadata; upstream Codeforces terms also apply \\
GitHub Top Code & Code & -- & 1.55B & P2: No additional sampling & ronantakizawa/github-top-code & MIT metadata; original repository licenses apply \\
Jupyter Agent & Code & -- & 0.23B & P2: No additional sampling & jupyter-agent/jupyter-agent-dataset~\cite{jupyteragentdataset} & Apache 2.0; referenced source terms also apply \\
LongCodeU CU-DFA & Code & -- & 0.02B & P2: No additional sampling & longcodeu/longcodeu-dataset~\cite{ye2025longproc} & Apache 2.0 \\
Nemotron HQ & English & 107.51B & 447.97B & Both: Score-filtered & nvidia/Nemotron-CC-v2~\cite{su2025nemotroncctransformingcommoncrawl,nemotron_nana_v2} & NVIDIA Data Agreement for Model Training; gated, no raw-data redistribution \\
Nemotron HQ Synthetic & English & 97.67B & 39.64B & Both: Score-filtered & nvidia/Nemotron-CC-v2~\cite{su2025nemotroncctransformingcommoncrawl,nemotron_nana_v2} & NVIDIA Data Agreement for Model Training; gated, no raw-data redistribution \\
FineWeb-Edu-EN & English & 59.79B & -- & P1: Score-filtered & HuggingFaceTB/smollm-corpus~\cite{benallal2024smollmcorpus} & ODC-By 1.0; Common Crawl terms also apply \\
Cosmopedia-v2 & English & 27.41B & 27.41B & Both: No additional sampling & HuggingFaceTB/smollm-corpus~\cite{benallal2024smollmcorpus} & ODC-By 1.0 \\
DCLM-Dedup & English & -- & 33.77B & P2: Score-filtered & mlfoundations/dclm-baseline-1.0~\cite{DCLM} & CC BY 4.0; Common Crawl terms also apply \\
ArXiv & English & 28.93B & -- & P1: No additional sampling & togethercomputer/RedPajama-Data-1T~\cite{together2023redpajama} & Metadata CC0 1.0; article content has author-selected licenses \\
FineWiki-EN & English & -- & 8.74B & P2: No additional sampling & HuggingFaceFW/finewiki~\cite{penedo2025finewiki} & CC BY-SA 4.0; older Wikipedia content may also be GFDL \\
UltraData-Math & Math & -- & 118.33B & P2: No additional sampling & openbmb/UltraData-Math~\cite{ultradata-math} & Apache 2.0 on the current dataset card \\
SwallowMath-v2 & Math & 9.99B & 13.32B & Both: Random subsample & tokyotech-llm/swallow-math-v2~\cite{fujii2025rewritingpretrainingdataboosts} & Apache 2.0 \\
Nemotron-CC-Math & Math & 11.48B & 2.47B & Both: Score-filtered & nvidia/Nemotron-CC-Math-v1~\cite{nemotron_nana_v2} & NVIDIA Data Agreement/Open Data terms; production revision to pin \\
MegaMath-Web-Pro & Math & -- & 13.45B & P2: No additional sampling & LLM360/MegaMath~\cite{zhou2025megamath} & ODC-By 1.0 \\
OpenWebMath & Math & -- & 13.23B & P2: No additional sampling & open-web-math/open-web-math~\cite{paster2023openwebmath} & ODC-By 1.0; Common Crawl terms also apply \\
FineMath & Math & 10.10B & -- & P1: No additional sampling & HuggingFaceTB/finemath~\cite{smol2} & ODC-By 1.0 \\
AutoMathText & Math & -- & 8.71B & P2: No additional sampling & math-ai/AutoMathText~\cite{zhang-etal-2025-autonomous} & CC BY-SA 4.0 \\
STEM Reasoning Complex & Math & -- & 1.21B & P2: No additional sampling & galaxyMindAiLabs/stem-reasoning-complex & Apache 2.0 \\
NuminaMath-CoT & Math & -- & 0.44B & P2: No additional sampling & AI-MO/NuminaMath-CoT~\cite{numina_math_datasets} & Apache 2.0 \\
FineProofs SFT & Math & -- & 0.13B & P2: No additional sampling & lm-provers/FineProofs-SFT & Apache 2.0 \\
Nemotron Terminal Corpus & SFT / instruction & -- & 6.27B & P2: No additional sampling & nvidia/Nemotron-Terminal-Corpus~\cite{pi2026dataengineeringscalingllm} & CC BY 4.0 on the current dataset card; production revision to pin \\
JiuZhang3.0 PT-CoT & SFT / instruction & -- & 3.58B & P2: No additional sampling & ToheartZhang/JiuZhang3.0-Corpus-PT-CoT & Not specified on the upstream dataset card \\
Tulu-3 SFT & SFT / instruction & -- & 0.64B & P2: No additional sampling & allenai/tulu-3-sft-mixture~\cite{lambert2024tulu3} & ODC-By 1.0 mixture; stricter subset terms may apply \\
ToolMind & SFT / instruction & -- & 0.53B & P2: No additional sampling & Nanbeige/ToolMind~\cite{yang2025toolmindtechnicalreportlargescale} & Apache 2.0 \\
ToolBench & SFT / instruction & -- & 0.47B & P2: No additional sampling & tuandunghcmut/toolbench-v1~\cite{qin2023toolllm} & Apache 2.0 metadata \\
ToolMind-Web-QA & SFT / instruction & -- & 0.43B & P2: No additional sampling & Nanbeige/ToolMind-Web-QA~\cite{yang2026nanbeige413bsmallgeneralmodel} & Apache 2.0 \\
Multi-Turn Long Context & SFT / instruction & -- & 0.22B & P2: No additional sampling & TreeAILab/Multi-turn\_Long-context\_Benchmark\_for\_LLMs~\cite{li2025loopserveadaptivedualphasellm} & CC BY 4.0 \\
SlimOrca & SFT / instruction & -- & 0.20B & P2: No additional sampling & Open-Orca/SlimOrca~\cite{mukherjee2023orca,longpre2023flan} & MIT \\
LongAlpaca & SFT / instruction & -- & 0.12B & P2: No additional sampling & Yukang/LongAlpaca-12k~\cite{long-alpaca} & No license declared on the current dataset card \\
Organic CoT & SFT / instruction & -- & 0.11B & P2: No additional sampling & CodonProject/Organic-Reasoning-195k (ingested as A03HCY/Organic-CoT-Reasoning-SFT) & Apache 2.0 \\
OpenThoughts Agent & SFT / instruction & -- & 0.11B & P2: No additional sampling & open-thoughts/OpenThoughts-Agent-v1-SFT~\cite{openthoughts-agent} & Apache 2.0 \\
\end{longtblr}

\end{landscape}
\pagestyle{fancy}
\markboth{\thesection\quad DATA RECIPE DETAILS}{\thesection\quad DATA RECIPE DETAILS}

\paragraph{Component accounting and license scope.}
\Cref{tab:data-components-detail} reports the materialized data components used in the two pretraining phases. Its token counts come from the component-level materialization records and therefore reflect the sampled, tokenized components actually used by the training recipe. The table identifies the upstream dataset or source family and states the current public license or terms that we could verify. A missing license is reported as ``to verify'' or ``not specified''. Nemotron-CC-v2 and the Synthetic-Code partition of Nemotron-Pretraining-Code-v1 use NVIDIA's data agreement, which permits model training but prohibits redistribution of the raw dataset~\cite{nvidiamodellicense}. The merged Chinese web component uses the same CCI, SkyPile-150B, WanJuan1.0, IndustryCorpus, and WuDaoCorpus2.0 source families documented by the Kaiyuan recipe and inherits their respective terms~\cite{wang2024cci30hqlargescalechinesedataset,wei2023skywork,he2023wanjuan,he2024opendatalabempoweringgeneralartificial,industrycorpus2,ZHANG2021216,ZHANG202193,cciagreement,skyworklicense}. Due to these terms, we also release our data preprocessing framework for data construction.

\paragraph{Proxy measurement protocol.}
\label{app:data-recipe:proxy-protocol}

We use a Qwen3-0.6B model for proxy experiments. The proxy experiment starts from a checkpoint
trained on 86B tokens on the same broad base mixture. Then it receives approximately 8.4B continuation tokens. It corresponds to 2,000 steps with a sequence length of 4,096 and a global batch size of 1,024.
The continuation differs only in its candidate source or within-source slice.
We evaluate the final checkpoint on a suite of benchmarks and aggregate evaluation scores into Math, Code, Chinese, and General axes. %
Math combines GSM8K and MATH, Code combines MBPP and HumanEval, Chinese combines C-Eval and CMMLU, and General summarizes the remaining knowledge and
commonsense tasks.

We adopt a continuous data schedule in the proxy experiments. Let $r(t)$ denote the candidate fraction at continuation step $t$. The candidate share follows
\[
  r(t)=\min\!\left(0.8,\;0.8\times \frac{t}{1600}\right),
  \qquad 0\leq t\leq 2000,
\]
so the candidate ramps from 0\% to 80\% during the first 1,600 steps and remains
at 80\% for the final 400 steps.
This schedule controls the distribution shift to avoid abrupt changes in distribution for measurement.
The resulting feature vector depends on the experiment setup, including proxy scale, continuation schedule, and evaluation suite.

\begin{figure}[!ht]
\centering
\begin{subfigure}[t]{0.40\linewidth}
\centering
\includegraphics[width=\linewidth]{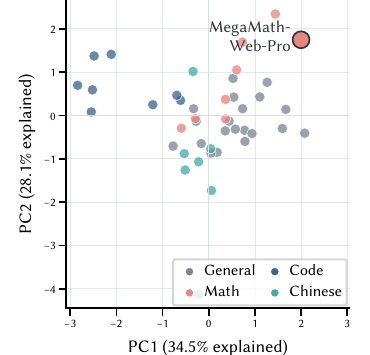}
\caption{Candidate profiles }
\end{subfigure}
\begin{subfigure}[t]{0.40\linewidth}
\centering
\includegraphics[width=\linewidth]{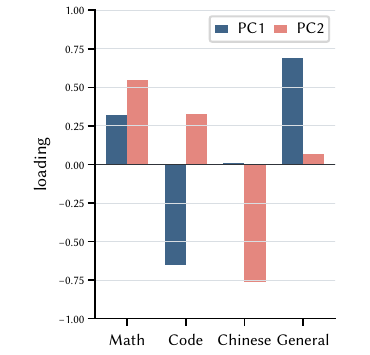}
\caption{PCA loadings }
\end{subfigure}
\caption{Proxy evidence used in recipe design: \textbf{(a)} Candidate profiles on the first two principal components and \textbf{(b)} the capability contrasts exposed by the PCA loadings.  PCA is descriptive; the comparisons do not establish causal gains.
}
\label{fig:proxy-feature-pca}
\end{figure}

\paragraph{Proxy Capability Dimensions}
\label{app:data-recipe:proxy-pca}

The raw proxy export contains 39 source or score-quantile slices evaluated on 15 benchmarks. We use four capability dimensions to make the benchmark panel interpretable: Math averages GSM8K and MATH; Code averages MBPP and HumanEval; Chinese averages C-Eval and CMMLU; and General averages the remaining nine English knowledge and commonsense tasks. The raw percentage averages and their within-panel ranks over the 39 data slices are reported in \Cref{tab:proxy-benchmark-aggregates}. For the PCA visualization, each
benchmark $b$ is first converted to a population
z-score across the 39 candidates,
\[
  z_{i,b}=\frac{x_{i,b}-\mu_b}{\sigma_b}.
\]
The PCA visualization averages the corresponding z-score columns within each dimension and then standardizes the four aggregate axes once more before
fitting PCA.  The first two components explain 34.5\% and 28.1\% of the
standardized four-axis variance.  Panel~(b) shows the resulting loading
contrasts: PC1 primarily contrasts General with Code, while PC2 primarily
contrasts Math with Chinese.  These directions summarize co-variation in this
panel.

\begin{landscape}
\pagestyle{plain}
\begingroup
\fontsize{7.5}{9}\selectfont
\setlength{\tabcolsep}{2.8pt}
\begin{longtable}{@{}llrrrrrrrrrrrrrrr@{}}
\caption{Raw proxy benchmark scores (percentage points) before per-task standardization. Dataset labels follow the canonical family names in~\Cref{tab:data-components-detail}. Source qualifiers remain in the Slice column. The redundant DCLM-Dedup-50B probes are omitted because they duplicate DCLM-Dedup in the production recipe.}\label{tab:proxy-raw-benchmarks}\\
\toprule
Source & Slice & GSM8K & MATH & MBPP & HE & CEval & CMMLU & MMLU & ARC-C & ARC-E & BoolQ & CSQA & HSwag & PIQA & SIQA & WinoG \\
\midrule
\endfirsthead
\multicolumn{17}{c}{\tablename\ \thetable\ -- continued}\\
\toprule
Source & Slice & GSM8K & MATH & MBPP & HE & CEval & CMMLU & MMLU & ARC-C & ARC-E & BoolQ & CSQA & HSwag & PIQA & SIQA & WinoG \\
\midrule
\endhead
\bottomrule
\endfoot
Algebraic Stack Train & random & 0.38 & 0.96 & 3.11 & 5.49 & 29.78 & 30.37 & 30.82 & 28.14 & 61.20 & 62.32 & 47.91 & 38.71 & 65.61 & 43.14 & 52.41 \\
ArXiv & random & 0.68 & 1.04 & 1.95 & 5.49 & 29.06 & 30.53 & 30.85 & 26.44 & 58.91 & 62.26 & 47.50 & 38.39 & 66.92 & 43.86 & 52.49 \\
AutoMathText & random & 0.61 & 1.44 & 6.61 & 4.88 & 29.58 & 30.47 & 31.03 & 29.15 & 65.08 & 57.40 & 49.63 & 38.70 & 67.30 & 44.98 & 52.57 \\
Cosmopedia-v2 & random & 0.38 & 1.54 & 5.06 & 5.49 & 30.97 & 30.35 & 31.24 & 27.12 & 63.14 & 53.09 & 47.83 & 40.09 & 67.68 & 44.11 & 52.80 \\
DCLM-Dedup & q00 & 0.91 & 0.96 & 1.95 & 4.88 & 29.39 & 31.04 & 32.03 & 29.83 & 63.67 & 62.32 & 51.02 & 40.69 & 68.72 & 44.42 & 53.20 \\
DCLM-Dedup & q25 & 0.91 & 1.34 & 3.50 & 7.32 & 28.73 & 30.71 & 30.95 & 26.78 & 61.02 & 60.49 & 49.30 & 40.63 & 68.66 & 44.68 & 53.59 \\
DCLM-Dedup & q50 & 0.38 & 0.88 & 1.56 & 4.88 & 29.75 & 30.42 & 30.59 & 28.47 & 61.20 & 60.34 & 49.47 & 40.70 & 67.85 & 44.32 & 54.30 \\
DCLM-Dedup & q75 & 1.14 & 0.92 & 2.72 & 5.49 & 30.79 & 30.68 & 30.11 & 28.14 & 58.02 & 61.77 & 48.65 & 40.10 & 68.55 & 43.45 & 52.88 \\
FLAN & random & 0.61 & 0.84 & 1.17 & 3.05 & 29.28 & 30.49 & 30.37 & 28.14 & 60.14 & 60.15 & 47.91 & 39.46 & 67.19 & 43.76 & 52.17 \\
PES2O & random & 0.45 & 0.66 & 2.72 & 5.49 & 29.37 & 30.84 & 31.48 & 26.78 & 60.32 & 56.67 & 47.26 & 38.12 & 66.27 & 43.04 & 53.20 \\
FineMath & random & 1.59 & 1.74 & 3.89 & 3.05 & 28.70 & 30.67 & 30.84 & 26.78 & 61.55 & 62.72 & 47.83 & 38.64 & 67.36 & 44.42 & 50.43 \\
FineWeb-Edu-CN & q00 & 0.83 & 0.56 & 1.17 & 6.10 & 31.83 & 32.44 & 30.76 & 27.46 & 60.49 & 55.69 & 48.24 & 38.36 & 67.79 & 43.65 & 52.72 \\
FineWeb-Edu-CN & q25 & 0.76 & 0.60 & 4.28 & 4.88 & 29.61 & 31.24 & 30.59 & 29.49 & 60.49 & 56.12 & 48.48 & 38.53 & 67.25 & 43.96 & 53.28 \\
FineWeb-Edu-CN & q50 & 0.38 & 0.76 & 3.11 & 5.49 & 29.05 & 31.16 & 30.92 & 28.47 & 61.38 & 52.51 & 49.30 & 39.27 & 67.41 & 44.32 & 54.14 \\
FineWeb-Edu-CN & q75 & 0.99 & 0.90 & 2.33 & 6.71 & 28.51 & 30.39 & 30.44 & 29.15 & 61.02 & 55.38 & 47.91 & 38.48 & 67.85 & 43.76 & 52.64 \\
FineWeb-Edu-EN & q00 & 0.61 & 1.06 & 2.72 & 4.88 & 29.85 & 30.58 & 32.48 & 30.85 & 66.84 & 59.88 & 49.63 & 39.90 & 68.99 & 43.04 & 52.33 \\
FineWeb-Edu-EN & q25 & 0.38 & 0.68 & 2.72 & 5.49 & 30.00 & 30.24 & 31.51 & 25.76 & 61.55 & 58.69 & 49.47 & 40.29 & 68.12 & 45.39 & 52.88 \\
FineWeb-Edu-EN & q50 & 0.68 & 1.56 & 2.72 & 4.88 & 29.34 & 30.16 & 30.77 & 29.83 & 60.32 & 61.25 & 50.12 & 41.32 & 68.06 & 45.14 & 52.49 \\
FineWeb-Edu-EN & q75 & 0.45 & 1.14 & 2.33 & 5.49 & 30.16 & 30.56 & 31.54 & 26.78 & 61.20 & 55.11 & 49.63 & 40.92 & 69.15 & 44.83 & 52.49 \\
FineWiki-EN & random & 0.99 & 0.70 & 2.72 & 6.71 & 30.17 & 30.37 & 29.98 & 28.14 & 63.49 & 61.68 & 48.32 & 38.77 & 66.59 & 44.63 & 51.62 \\
MegaMath-Code & random & 1.14 & 1.30 & 13.23 & 10.37 & 29.64 & 30.60 & 31.14 & 28.81 & 61.90 & 62.05 & 46.76 & 38.53 & 66.49 & 43.19 & 51.70 \\
MegaMath-Web & random & 0.99 & 0.62 & 3.11 & 2.44 & 29.68 & 30.18 & 30.73 & 28.47 & 62.96 & 50.61 & 47.67 & 38.30 & 66.49 & 43.30 & 51.30 \\
MegaMath-Web-Pro & random & 3.03 & 1.90 & 3.50 & 5.49 & 29.93 & 30.53 & 31.86 & 30.51 & 63.14 & 62.78 & 49.80 & 39.64 & 67.03 & 44.47 & 51.54 \\
Nemotron-CC-Math & random & 1.82 & 3.74 & 2.72 & 6.71 & 29.13 & 30.48 & 31.36 & 29.83 & 62.08 & 61.99 & 48.98 & 38.96 & 66.70 & 43.91 & 53.28 \\
Nemotron HQ & random (medium-high-quality) & 0.61 & 1.32 & 1.56 & 6.10 & 29.33 & 30.55 & 30.82 & 27.46 & 59.79 & 58.35 & 50.04 & 40.72 & 68.34 & 44.83 & 53.04 \\
Nemotron HQ & random & 0.53 & 1.24 & 2.33 & 6.71 & 29.17 & 30.45 & 30.80 & 31.53 & 62.26 & 60.21 & 48.08 & 41.31 & 68.17 & 43.30 & 52.80 \\
Nemotron HQ Synthetic & random & 0.38 & 1.12 & 5.06 & 6.71 & 30.60 & 30.74 & 31.22 & 27.80 & 63.67 & 60.40 & 47.50 & 40.94 & 68.93 & 43.81 & 52.25 \\
Nemotron Synthetic Code & random & 0.53 & 0.70 & 7.39 & 10.37 & 29.22 & 30.45 & 30.32 & 25.42 & 60.85 & 57.25 & 46.36 & 38.18 & 66.54 & 44.22 & 51.07 \\
OpenWebMath & random & 0.53 & 1.18 & 3.11 & 4.88 & 29.27 & 30.65 & 31.09 & 29.83 & 61.90 & 53.55 & 47.99 & 38.48 & 67.95 & 44.98 & 52.96 \\
StackExchange & random & 0.45 & 0.54 & 5.84 & 4.88 & 29.24 & 30.17 & 30.64 & 29.83 & 60.85 & 55.41 & 47.75 & 39.14 & 67.30 & 43.91 & 52.96 \\
Stack-v2-Smol & random & 0.91 & 1.10 & 5.84 & 4.88 & 30.23 & 30.07 & 30.13 & 29.49 & 59.08 & 62.42 & 45.78 & 37.87 & 67.14 & 43.24 & 50.51 \\
StarCoder Tokens & random & 0.68 & 0.50 & 6.61 & 4.27 & 29.68 & 29.92 & 30.89 & 27.80 & 61.55 & 62.23 & 45.86 & 38.11 & 67.30 & 42.94 & 53.75 \\
Swallow-Code-v2 & q00 & 0.61 & 0.90 & 8.95 & 10.37 & 28.20 & 30.71 & 30.55 & 28.14 & 60.49 & 60.40 & 47.99 & 38.83 & 67.30 & 43.40 & 52.64 \\
Swallow-Code-v2 & q25 & 0.30 & 0.82 & 9.34 & 9.15 & 28.95 & 30.65 & 30.50 & 28.14 & 58.91 & 59.66 & 48.48 & 38.57 & 66.59 & 42.84 & 52.17 \\
Swallow-Code-v2 & q75 & 0.68 & 0.42 & 8.95 & 9.76 & 29.71 & 30.74 & 31.13 & 26.10 & 61.55 & 60.89 & 46.11 & 38.63 & 66.70 & 42.99 & 51.70 \\
SwallowMath-v2 & random & 1.21 & 4.60 & 0.39 & 7.32 & 30.26 & 30.18 & 30.80 & 27.12 & 61.38 & 62.11 & 47.34 & 38.60 & 67.57 & 43.76 & 51.07 \\
Deduplicated merged Chinese web & q00 & 0.53 & 0.76 & 2.33 & 4.27 & 30.65 & 30.96 & 30.77 & 27.46 & 59.61 & 59.05 & 47.91 & 38.33 & 67.79 & 44.37 & 52.72 \\
Deduplicated merged Chinese web & q50 & 0.83 & 0.68 & 3.89 & 4.88 & 30.40 & 30.65 & 30.74 & 26.10 & 59.79 & 57.43 & 47.91 & 38.73 & 67.19 & 44.47 & 51.85 \\
Deduplicated merged Chinese web & q75 & 0.45 & 0.60 & 2.72 & 4.88 & 29.74 & 31.03 & 30.82 & 27.12 & 60.14 & 57.74 & 47.26 & 38.97 & 67.14 & 44.06 & 51.93 \\
\end{longtable}

\begin{longtable}{@{}llrrrrrrrrrr@{}}
\caption{Raw proxy benchmark aggregates and descending ranks by data slice. Dataset labels follow the canonical family names in~\Cref{tab:data-components-detail}; source qualifiers remain in the Slice column. Math, Code, and CN average 2 benchmarks each; General (the English knowledge and commonsense axis) averages 9 benchmarks; Overall averages all 15 raw benchmark scores. The ranks are computed over the 39 displayed non-duplicate slices; rank 1 is highest.}\label{tab:proxy-benchmark-aggregates}\\
\toprule
Source & Slice & Math avg & Math rank & Code avg & Code rank & CN avg & CN rank & General avg & General rank & Overall avg & Overall rank \\
\midrule
\endfirsthead
\multicolumn{12}{c}{\tablename\ \thetable\ -- continued}\\
\toprule
Source & Slice & Math avg & Math rank & Code avg & Code rank & CN avg & CN rank & General avg & General rank & Overall avg & Overall rank \\
\midrule
\endhead
\bottomrule
\endfoot
MegaMath-Code & random & 1.22 & 5 & 11.80 & 1 & 30.12 & 17 & 47.84 & 17 & 34.46 & 1 \\
MegaMath-Web-Pro & random & 2.46 & 3 & 4.50 & 18 & 30.23 & 11 & 48.97 & 3 & 34.34 & 2 \\
DCLM-Dedup & q00 & 0.94 & 14 & 3.42 & 35 & 30.21 & 14 & 49.54 & 1 & 34.34 & 3 \\
FineWeb-Edu-EN & q00 & 0.83 & 19 & 3.80 & 29 & 30.21 & 14 & 49.33 & 2 & 34.24 & 4 \\
Nemotron-CC-Math & random & 2.78 & 2 & 4.71 & 13 & 29.80 & 30 & 48.57 & 7 & 34.11 & 5 \\
Nemotron HQ Synthetic & random & 0.75 & 24 & 5.88 & 6 & 30.67 & 4 & 48.50 & 8 & 34.08 & 6 \\
Swallow-Code-v2 & q00 & 0.76 & 22 & 9.66 & 2 & 29.45 & 38 & 47.75 & 21 & 33.97 & 7 \\
AutoMathText & random & 1.02 & 9 & 5.75 & 7 & 30.02 & 23 & 48.43 & 10 & 33.96 & 8 \\
Nemotron HQ & random & 0.89 & 15 & 4.52 & 16 & 29.81 & 29 & 48.72 & 5 & 33.93 & 9 \\
FineWeb-Edu-EN & q50 & 1.12 & 7 & 3.80 & 29 & 29.75 & 34 & 48.81 & 4 & 33.91 & 10 \\
DCLM-Dedup & q25 & 1.12 & 6 & 5.41 & 9 & 29.72 & 35 & 48.46 & 9 & 33.91 & 11 \\
Swallow-Code-v2 & q75 & 0.55 & 36 & 9.36 & 3 & 30.23 & 12 & 47.31 & 32 & 33.74 & 12 \\
DCLM-Dedup & q50 & 0.63 & 30 & 3.22 & 37 & 30.09 & 21 & 48.58 & 6 & 33.67 & 13 \\
Swallow-Code-v2 & q25 & 0.56 & 34 & 9.24 & 4 & 29.80 & 32 & 47.32 & 31 & 33.67 & 14 \\
FineWiki-EN & random & 0.84 & 18 & 4.71 & 13 & 30.27 & 10 & 48.14 & 13 & 33.66 & 15 \\
SwallowMath-v2 & random & 2.90 & 1 & 3.85 & 27 & 30.22 & 13 & 47.75 & 20 & 33.58 & 16 \\
DCLM-Dedup & q75 & 1.03 & 8 & 4.11 & 22 & 30.73 & 3 & 47.96 & 14 & 33.56 & 17 \\
FineWeb-Edu-EN & q25 & 0.53 & 37 & 4.11 & 22 & 30.12 & 18 & 48.18 & 11 & 33.54 & 18 \\
Nemotron HQ & random (medium-high-quality) & 0.96 & 11 & 3.83 & 28 & 29.94 & 25 & 48.15 & 12 & 33.52 & 19 \\
StarCoder Tokens & random & 0.59 & 32 & 5.44 & 8 & 29.80 & 31 & 47.83 & 18 & 33.47 & 20 \\
FineWeb-Edu-EN & q75 & 0.80 & 21 & 3.91 & 26 & 30.36 & 9 & 47.96 & 15 & 33.45 & 21 \\
Cosmopedia-v2 & random & 0.96 & 12 & 5.28 & 12 & 30.66 & 5 & 47.46 & 29 & 33.39 & 22 \\
Algebraic Stack Train & random & 0.67 & 28 & 4.30 & 20 & 30.07 & 22 & 47.81 & 19 & 33.36 & 23 \\
FineMath & random & 1.67 & 4 & 3.47 & 34 & 29.68 & 37 & 47.84 & 16 & 33.35 & 24 \\
FineWeb-Edu-CN & q25 & 0.68 & 27 & 4.58 & 15 & 30.43 & 7 & 47.58 & 24 & 33.30 & 25 \\
StackExchange & random & 0.49 & 39 & 5.36 & 10 & 29.70 & 36 & 47.53 & 26 & 33.26 & 26 \\
Nemotron Synthetic Code & random & 0.61 & 31 & 8.88 & 5 & 29.84 & 28 & 46.69 & 38 & 33.26 & 27 \\
Stack-v2-Smol & random & 1.00 & 10 & 5.36 & 10 & 30.15 & 16 & 47.30 & 33 & 33.25 & 28 \\
OpenWebMath & random & 0.85 & 17 & 4.00 & 25 & 29.96 & 24 & 47.64 & 23 & 33.22 & 29 \\
FineWeb-Edu-CN & q00 & 0.69 & 26 & 3.63 & 33 & 32.13 & 1 & 47.24 & 35 & 33.21 & 30 \\
FineWeb-Edu-CN & q50 & 0.57 & 33 & 4.30 & 20 & 30.11 & 19 & 47.52 & 27 & 33.18 & 31 \\
Deduplicated merged Chinese web & q00 & 0.65 & 29 & 3.30 & 36 & 30.80 & 2 & 47.56 & 25 & 33.17 & 32 \\
FineWeb-Edu-CN & q75 & 0.94 & 13 & 4.52 & 16 & 29.45 & 39 & 47.40 & 30 & 33.10 & 33 \\
ArXiv & random & 0.86 & 16 & 3.72 & 32 & 29.80 & 33 & 47.51 & 28 & 33.09 & 34 \\
Deduplicated merged Chinese web & q50 & 0.76 & 22 & 4.38 & 19 & 30.52 & 6 & 47.13 & 36 & 33.04 & 35 \\
FLAN & random & 0.72 & 25 & 2.11 & 39 & 29.89 & 27 & 47.70 & 22 & 32.98 & 36 \\
Deduplicated merged Chinese web & q75 & 0.53 & 38 & 3.80 & 29 & 30.39 & 8 & 47.24 & 34 & 32.97 & 37 \\
PES2O & random & 0.56 & 35 & 4.11 & 22 & 30.11 & 19 & 47.02 & 37 & 32.84 & 38 \\
MegaMath-Web & random & 0.81 & 20 & 2.77 & 38 & 29.93 & 26 & 46.65 & 39 & 32.46 & 39 \\
\end{longtable}

\endgroup
\end{landscape}
\pagestyle{fancy}

\end{document}